\documentclass{article}

\usepackage[final, nonatbib]{neurips_workshop_2021}

\usepackage[utf8]{inputenc} 
\usepackage[T1]{fontenc}    
\usepackage{hyperref}       
\usepackage{url}            
\usepackage{booktabs}       
\usepackage{amsfonts}       
\usepackage{nicefrac}       
\usepackage{microtype}      
\usepackage{xcolor}         
\usepackage{subfig}
\usepackage{graphicx}
\usepackage{amsmath}
\usepackage{amsthm}
\usepackage{stmaryrd}
\usepackage{comment}
\usepackage{multirow}
\usepackage{wrapfig}
\usepackage{tabularx}

\title{Neural NID Rules}
\author{%
  Luca Viano\\
  EPFL \\
  \texttt{luca.viano@epfl.ch} \\
  \And
  Johanni Brea\\
  EPFL \\
  \texttt{johanni.brea@epfl.ch} \\
}

\begin{document}

\maketitle

\begin{abstract} 
Abstract object properties and their relations are deeply rooted in human common sense, allowing people to predict the dynamics of the world even in situations that are novel but governed by familiar laws of physics. Standard machine learning models in model-based reinforcement learning are inadequate to generalize in this way.
Inspired by the classic framework of noisy indeterministic deictic (NID) rules, we introduce here Neural NID, a method that learns abstract object properties and relations between objects with a suitably regularized graph neural network. We validate the greater generalization capability of Neural NID on simple benchmarks specifically designed to assess the transition dynamics learned by the model.
\end{abstract}

\section{Introduction}

Human cognition relies on core knowledge about \emph{space}, \emph{actions} and \emph{objects} \cite{spelke2007core}.
Whereas \emph{actions} naturally occur in traditional Reinforcement Learning \cite{sutton2018reinforcement} and inductive biases related e.g. to translation invariance in \emph{space} are straightforward to specify, it is less clear how to implement core knowledge about \emph{objects}.
Whereas abstract (symbolic) representations of objects were already popular in classical Artificial Intelligence approaches \cite{McDermott98} or Relational Reinforcement Learning \cite{saso2001relational}, recent works focus on learning object-centric representations from raw sensory input \cite{Konidaris18,Ugur15,battaglia2018relational,battaglia2016interaction,Kulkarni19,Locatello20,kipf2020contrastive,cranmer2020discovering,veerapaneni2020entity,zambaldi2018relational,zhengyao2019neural,carvalho2020ROMA,greff2017neural, vansteenkiste2018relational}. 
Advantages of object-centric approaches are exemplified in the classic framework of \emph{Noisy Indeterministic Deictic (NID)} rules \cite{Pasula_2007, lang2012exploration}.
This framework allows to describe the agent environment by means of only few object properties and relations, neglecting all the irrelevant ones.
In particular, the idea of representing a scene only by means of the relevant object properties and useful relations among them allows for (i) immediate generalization across objects sharing the same relevant properties and (ii) generalization across tasks that can be described by the same relational predicates.

Consider, for example, the situation depicted in Figure \ref{fig:inclined_plane}.
The yellow and the green objects share the same spherical shape. It is the only property a human would use to predict if an object can roll down from the plane if left unconstrained.
For example, the frame sequence shown in the right column of Figure \ref{fig:inclined_plane} can be easily predicted by a human after having seen the frame sequence on the left.
With NID rules a machine can also learn to make correct predictions when given the example on the left in \autoref{fig:inclined_plane} and appropriate object properties (shape and whether the object is on the right or the left slope). 

Classic methods based on NID rules rely on human expertise because the relevant properties and relations are required as input.
In contrast, the recent neural network approaches rely on a ``representation'' feedforward neural network to extract relevant properties from raw observations and an ``interaction'' graph neural network to model relations between objects \cite{battaglia2018relational,kipf2020contrastive}. 

Although the neural network approaches alleviate the need for human expertise, we argue here that they bear undesirable symmetries.
The basic argument is the following: in a setting where we want to predict the transformation $f(X, C)$ of an object $X$ in a context $C$ with a representation network $r(X)$ followed by prediction network $g(r(X), C)$, i.e. $f(X, C) \approx g(r(X), C))$, two objects A and B that behave the same $g(r(\mathrm{A}), \mathrm C_1) = g(r(\mathrm{B}), \mathrm C_1)$ in some context C$_1$ may have different representations $r(\mathrm{A})\neq r(\mathrm{B})$ and therefore potentially differing predictions $g(r(\mathrm{A}), \mathrm{C}_2)\neq g(r(\mathrm{B}), \mathrm{C}_2)$ in another context C$_2$.
In other words, these systems have a symmetry characterised by the invariant set of representation networks $\{r | g(r(\mathrm{A}), \mathrm C_1) = g(r(\mathrm{B}), \mathrm C_1)\}$.
For efficient generalization, it is desirable to break this symmetry with a prior that reflects the common sense reasoning ``if it looks like a duck and walks like a duck, it is a duck''.
Here we propose a prior to break this symmetry and we investigate its effectiveness empirically.
Additionally we show how our neural network approach relates to classical NID rules.

\section{Neural NID Rules}
\label{sec: Neural NID}
NID rules \cite{Pasula_2007, lang2012exploration} consist of a formalism to describe the transition dynamics in relational domains. In these domains, one assumes access to an action set $\mathcal{A}$, to an object set $\mathcal{O}$, to a property set $\mathcal{P} = \left\{p_{j}: \mathcal{O} \rightarrow \{\mathrm{True}, \mathrm{False}\}\right\}_{j}$, to a function (or relation) set $\mathcal{F} = \left\{f_{j}: \mathcal{O}^{k_j} \rightarrow \{\mathrm{True}, \mathrm{False}\}\right\}_{j}$ where $k_j$ is an integer denoting how many objects are required as input of the $j^{th}$ function in $\mathcal{F}$. A rule $r$ is defined as
\begin{equation}
    a_r(\mathcal{X}): \phi_r(\mathcal{X}) \rightarrow \begin{cases}
        p_{r1} &: \Omega_{r1}(\mathcal{X}) \\
        \vdots & \vdots \\
        p_{rm} &: \Omega_{rm}(\mathcal{X}) \\
    \end{cases}
    \label{rule}
\end{equation}
where $\mathcal{X}$ is a subset of the object set $\mathcal{O}$, $a_r(\mathcal X)$ indicates that action $a_r$ is applied, $\phi_r(\mathcal{X})$ is an abstract boolean state context described in terms of properties and functions applied to the objects $\mathcal X$ (e.g. is\_round(object\_1) $\wedge$ on(object\_1, left\_plane)), and $\Omega_{rz}(\mathcal{X})$  is an outcome occurring with probability $p_{rz}$  for all $z \in \{1, \dots, m\}$.
The set $\mathcal{P}$ is designed such that all the irrelevant object properties are ignored. Thus, NID rules validly apply to contexts of unseen objects but with known properties and relations.
The main drawback of NID rules is the requirement of handcrafting the properties and functions sets $\mathcal{F}$ and $\mathcal{P}$.

With our Neural NID we bypass this requirement using an encoder network $f^{\mathrm{enc}}$ to learn properties and a graph neural network $f^\mathrm{edge}$ to learn relations.
We assume the sensory state $x_t$ at time $t$ of an agent consists of a set of low-level object representations $x_t = (o_1, \ldots, o_{N_t})$ where $N_t$ is the currently available number of objects.
The object representations $o_i$ could be images or features like shape, color or position.
The goal is to learn with as few observations as possible an accurate transition model $T$ such that $x_{t+1} \approx T(a_t, x_t)$, where $a_t$ are the agent's actions.
To get efficient generalization we want to equip the model with an inductive bias that favors grouped abstract representations for objects that behave the same under all training observations, even if they have different low-level representations.

We split the transition model into two parts: a \emph{transition map} $\Omega_z(x_t, i)$ that predicts the next low-level representation of object $i$ in state $x_t$ under transition $z$ and a \emph{transition selector} $P(z|x_t, i)$.
This split into transition selector and transition map is inspired by the NID rules (see Eq.~\ref{rule}).
In contrast to the NID rules, however, both parts are learned from experience.

\subsection{The Inclined Plane Domain}
As a proof of concept we study a simple domain with rollable and non-rollable objects on inclined planes (see Fig.~1).
For most experiments we do not include any actions in this domain (but see Appendix B).
The low-level object representations consist of the color and the x-coordinate of the objects, i.e. features that are uninformative about rollability and the descending direction of the slope.
Rollable objects move one step to the left when they are on the left plane and one step to the right otherwise, unless there is a non-rollable object that blocks their way.
The transition model needs to discover these rules from experience.

In the following, $o\in\mathcal{O}$ and $p\in\mathcal P$ are integers encoding the color and the x-coordinate of an object, respectively; $\mathbf{e}_{o}, \mathbf{e}_p$ denotes  the corresponding one hot encoded
vectors.
We consider as input at time step $t$ a tensor $\mathbf{x}_t \in \mathbb{R}^{|\mathcal{O}| \times |\mathcal P|}$.
For a fixed color index $o$, the tensor slice $\mathbf{x}_t[o]$ is a probability density function over the spatial domain that represents how likely it is to find an object in a particular position.
We denote as $\mathbf{x}_t[o, p]$ the probability of finding an object of color $o$ in position $p$.
Bold quantities, e.g. $\mathbf{v}$ are multidimensional tensors.
We denote access to tensor entries with squared brackets, e.g $\mathbf{v}[1]$.
We assume that all the other dimensions that are not explicitly indexed are kept.

For the Inclined Plane Domain the transition map $ \{ \Omega_z: \mathbb{R}^{|\mathcal{O}| \times |\mathcal{P}|} \rightarrow \mathbb{R}^{|\mathcal{O}| \times |\mathcal{P}|} \}_z$ is a collection of functions implemented via convolutional filters (see Appendix~\ref{sec:conv_filters}). 
Below we focus mostly on the transition selector which we write here as $P(z|\mathbf{x}, o, p)$ for an object with color $o$ at x-coordinate $p$ in context $\mathbf{x}$.
The network's prediction for the next x-coordinate $p^\prime$ of an object of color $o$ in position $p$ is given by a distribution $\widehat{\mathbf{x}}_{t+1}[o]$, where
\begin{equation}
    \widetilde{\mathbf{x}}_{t+1}[o, p^\prime] = \sum^{m}_{z_t=1} P( z_t | \mathbf{x}_t, o, p) \Omega_{z_t}(\mathbf{x}_t)[o,p^\prime] \quad \widehat{\mathbf{x}}_{t+1}[o] = \mathrm{softmax}(\widetilde{\mathbf{x}}_{t+1}[o])\, .
    \label{eq:output}
\end{equation}
Note that averaging over all possible transitions is expected to work in deterministic domains like the Inclined Plane Domain (an alternative would be to take the arg-max of $P(z_t|\mathbf{x}_t, o, p)$ or sampling).
In general stochastic domains it may be needed during training to include information from $\mathbf{x}_{t+1}$ while performing inference on $z_t$.

\subsection{The Transition Selector}
\label{sec:transition_selector}
The transition selector can be seen as a Graph Neural Network.
We define an encoding function such that $\boldsymbol{\theta}_t[o,p] = f^{\mathrm{enc}}(o, p)$.
The interactions between objects are modeled using an edge function: $\boldsymbol{\Phi}_t[o_1, o_2, p] = f^{\mathrm{edge}}(\mathbf{x}_t[o_1], \mathbf{x}_t[o_2], p)$.
The tensor is used to update the encoded object state with a node function $f^{\mathrm{node}}$:
$\tilde{\boldsymbol{\theta}}_t[o,p] = f^{\mathrm{node}}(\boldsymbol{\theta}_t[o,p], \sum_{\tilde{o} \neq o} \boldsymbol{\Phi}_t[o, \tilde{o}, p])$.
Finally, we get the output probability with a decoding function $f^{\mathrm{dec}}$: $P( z_t | \mathbf{x}_t, o, p) = f^{\mathrm{dec}}(\tilde{\boldsymbol{\theta}}_t[o,p])$.

A crucial requirement for efficient generalization is the design and training of $f^{\mathrm{enc}}$ such that low-level object representations are mapped to abstract representations that are relevant to predict the transitions.
For example in the Inclined Plane Domain a useful encoding would map the low-level ``color and x-position'' representation to an abstract ``shape and on left or on right plane'' representation.
In addition, efficient generalization relies also on the relational dynamics between rollable and blocking objects that should be learned from experience by $f^{\mathrm{node}}$.

We learn object properties with $f^{\mathrm{enc}}$ implemented as a particular fully connected architecture of the form $\boldsymbol{\theta}_t[o,p] = \sigma( \mathrm{softmax}(\mathrm{Concat}(\mathbf{e}_o, \mathbf{e}_p)^T\boldsymbol{Q}) \boldsymbol{V} ) \boldsymbol{W}$ where $\sigma$ denotes a sigmoid function, $\boldsymbol{W} \in \mathbb{R}^{d^1 \times d^P}$, $\boldsymbol{Q} \in \mathbb{R}^{(|\mathcal{O}| + |\mathcal{P}|) \times K }$ and $\boldsymbol{V} \in \mathbb{R}^{K \times d^1}$, where $\mathrm{softmax}$ acts row-wise.
The idea behind the design of $f^{\mathrm{enc}}$ is to extract abstract object properties.
We can think of the $K$ rows of $V$ as vectors representing learned abstract properties.
The $\mathrm{softmax}$ layer outputs an object-dependent probability distribution over those property vectors.

Note that the first layer of this architecture defines a symmetry.
Suppose we have an object of color $o_1$ at x-coordinate $p_1$ and a second object of color $o_2$ at $p_2$. Then, defining $\mathrm{softmax}(\mathrm{Concat}(\mathbf{e}_{o_i}, \mathbf{e}_{p_i})^T\boldsymbol{Q}) := \mathbf{P}_i$ for $i = 1,2$, it is possible that $\mathbf{P}_1 V = \mathbf{P}_2 V$ even if $\mathbf{P}_1 \neq \mathbf{P}_2$.
To avoid these cases we break the induced symmetry with entropy regularization terms that favor solutions with $\mathbf{P}_1 = \mathbf{P}_2$.
Thus, the training loss is given by the binary cross entropy (BCE) with entropy regularizers for $f^{\mathrm{enc}}$,
\begin{equation}
    L = \mathrm{BCE}( \mathbf{x}_{t+1}, \widehat{\mathbf{x}}_{t+1}) - \lambda_1 \sum^{|\mathcal{O}| + |\mathcal{P}|}_{i=1} \sum^K_{k=1} P(k, i) \log P(k|i) - \lambda_2 \sum^K_{k=1} P(k) \log P(k)\ .
    \label{eq:loss}
\end{equation}
where $P(k|i)$ is the $k^{th}$ entry of $\boldsymbol{P}_i$, $P(k,i) = P(k|i)\frac{1}{|\mathcal O| + |\mathcal P|}$, $P(k) = \sum_i P(k,i) $.
\section{Experiments}
\label{sec:evaluating_compactness}
\begin{figure}
\centering
\begin{tabular}{c} 
\subfloat[Schematics of the task \emph{Inclined Plane}
\label{fig:inclined_plane}]{  
\includegraphics[width=0.4\textwidth]{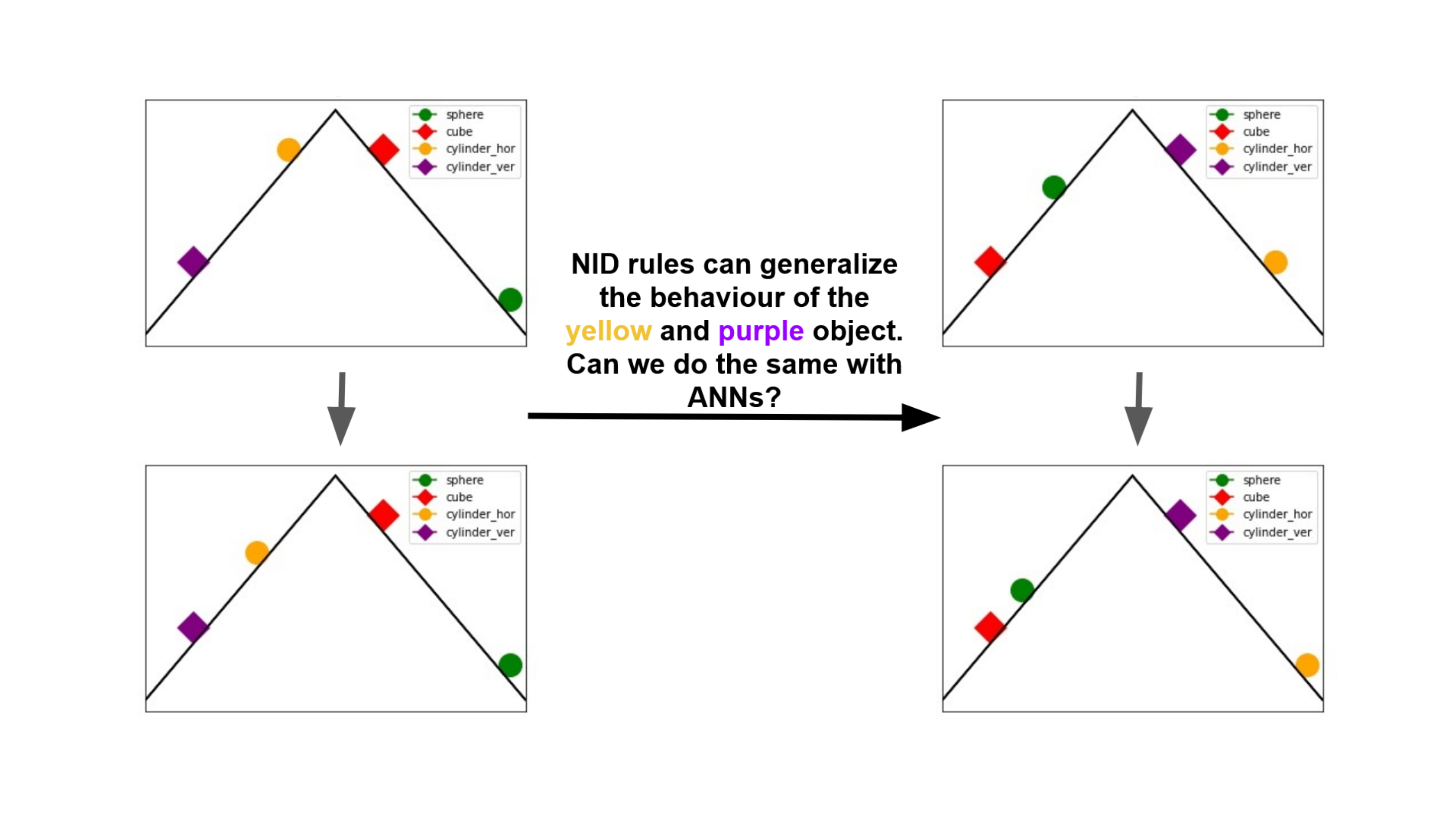}
 } \hfill
\subfloat[ Compound Training Errors \label{fig:main_compound_train_valley}]{%
       \includegraphics[width=0.3\linewidth]{ 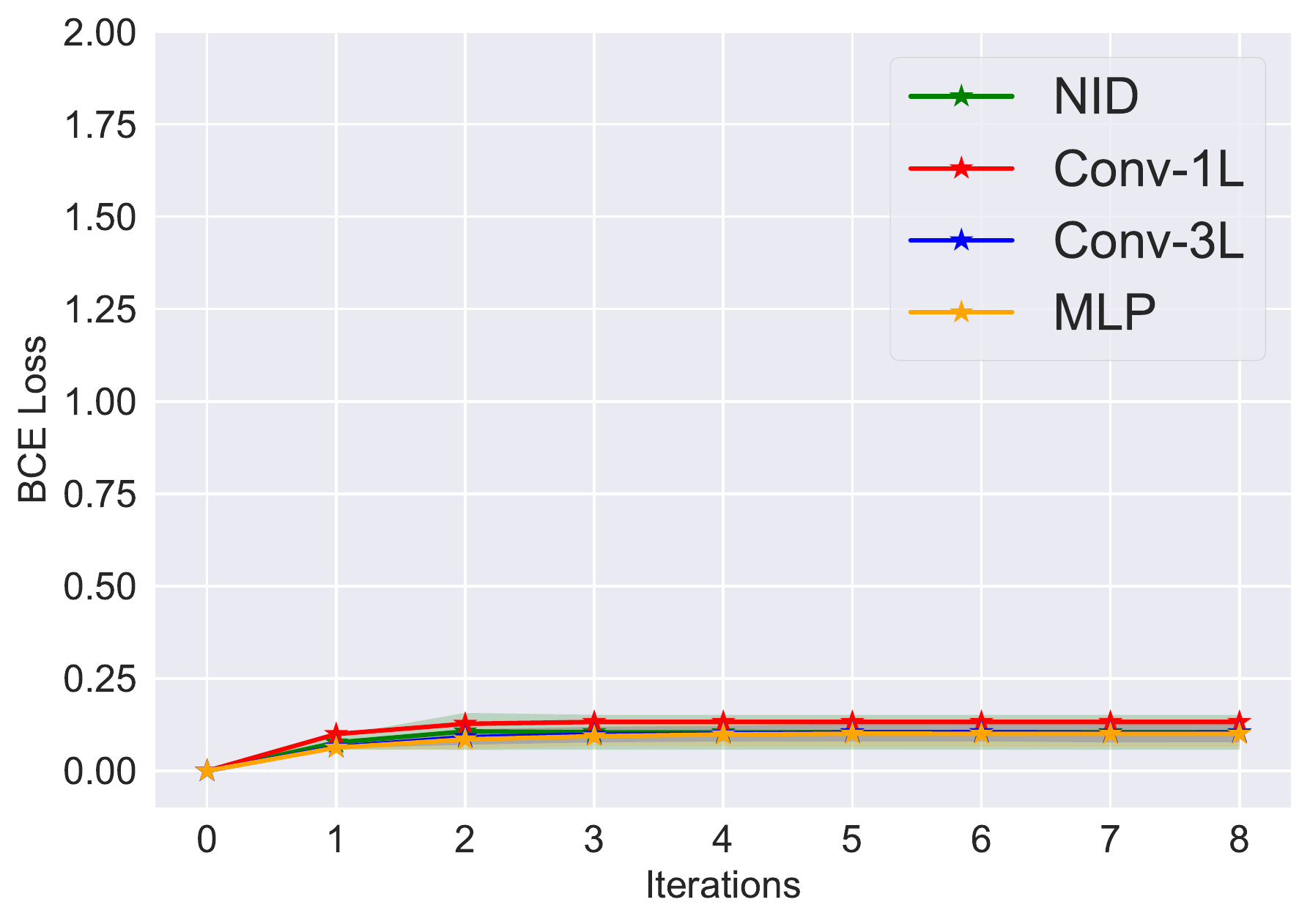}
     } \hfill
     \subfloat[ Compound Test Errors\label{fig:main_compound_test_valley}]{
       \includegraphics[width=0.3\linewidth]{ 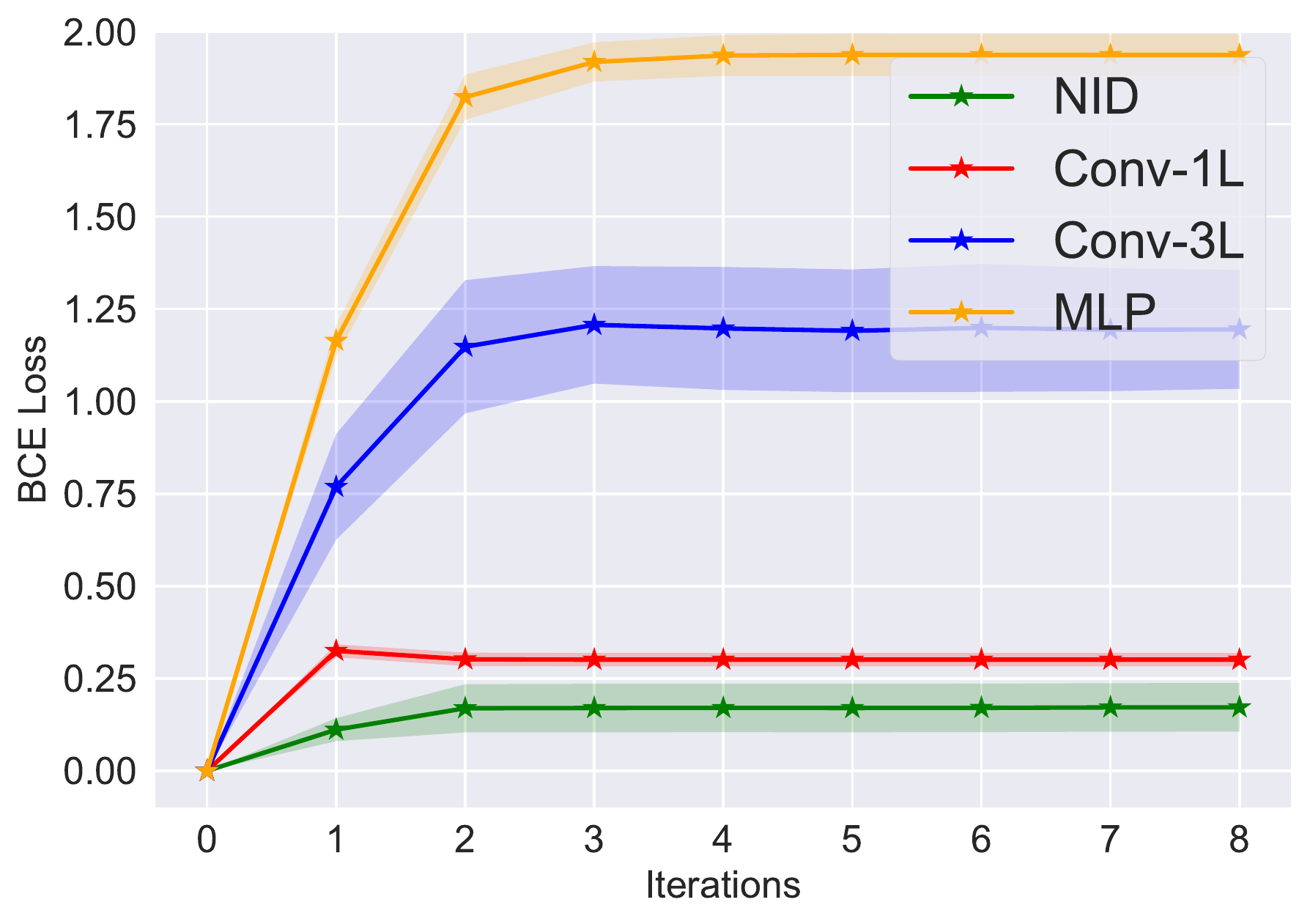}
       }  \\
      \subfloat[$\lambda_2$ coloring \label{fig:lambda_2_silohuette_main}]{%
       \includegraphics[width=0.33\linewidth]{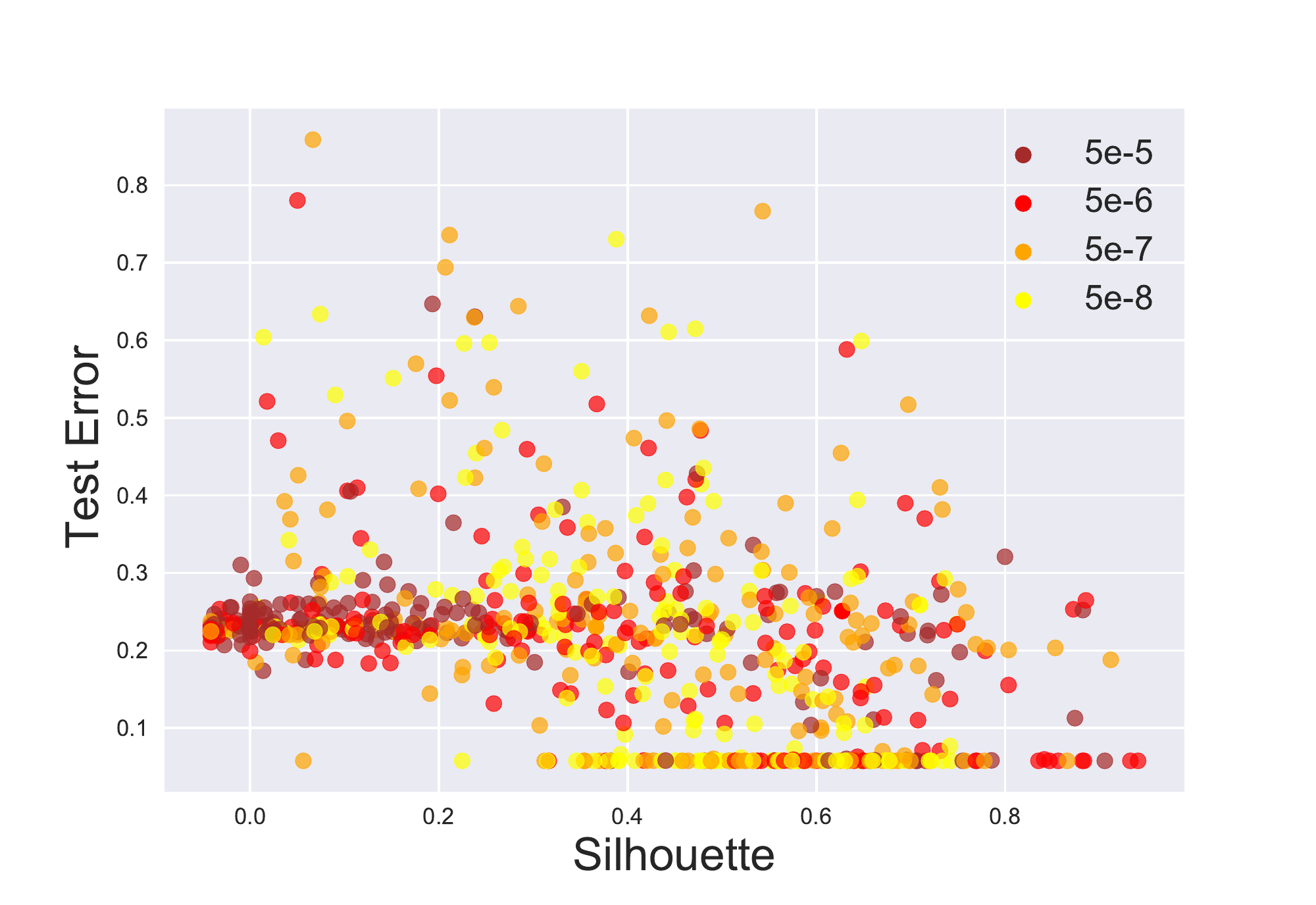}
       } \hfill
       \subfloat[$\lambda_1 = \lambda_2 = 0$ \label{f_enc_1}]{%
       \includegraphics[width=0.33\linewidth]{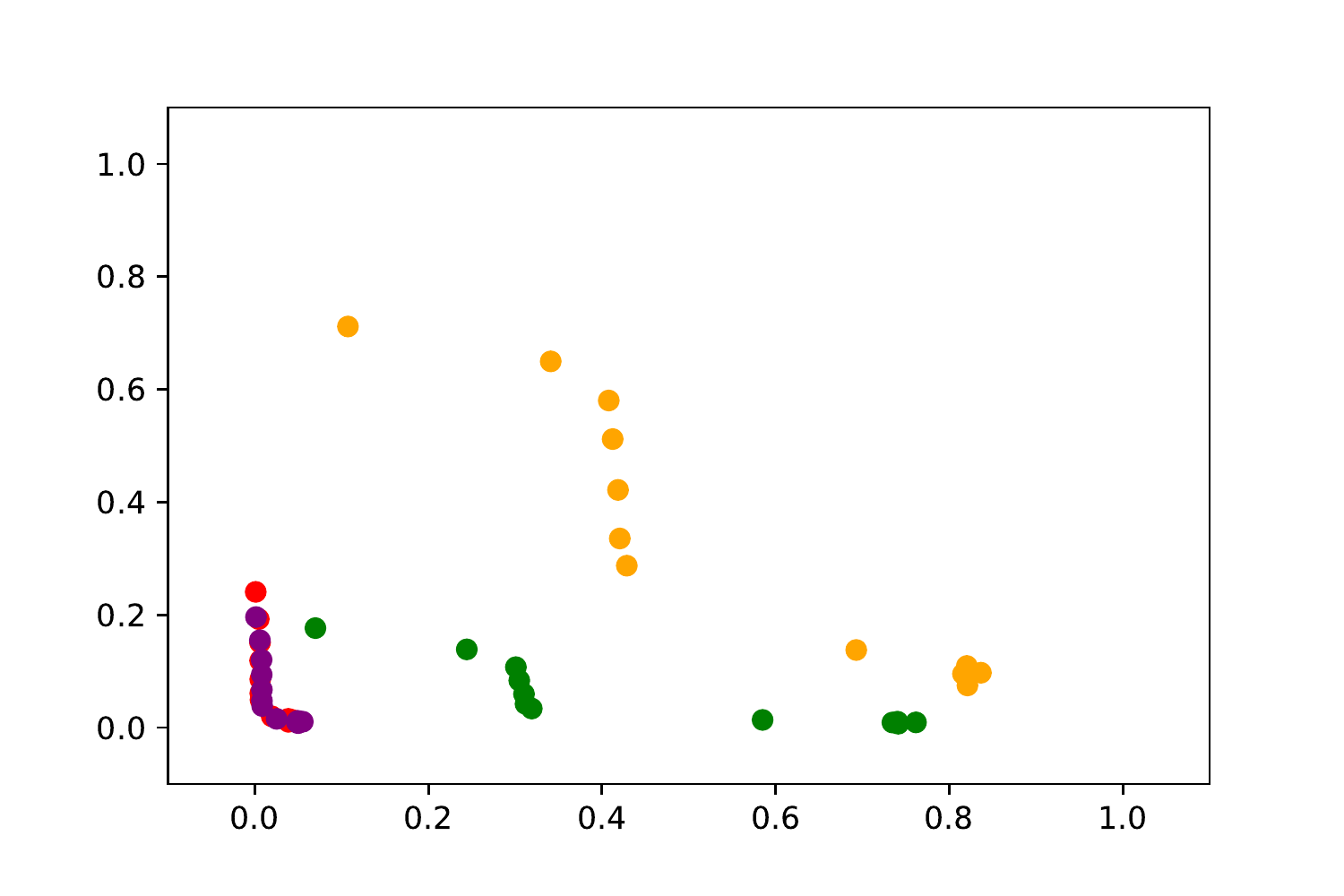}
       }  \hfill
       \subfloat[$\lambda_1 = 5e{-}8$, $\lambda_2 = 5e{-}6$ \label{f_enc_4}]{%
       \includegraphics[width=0.33\linewidth]{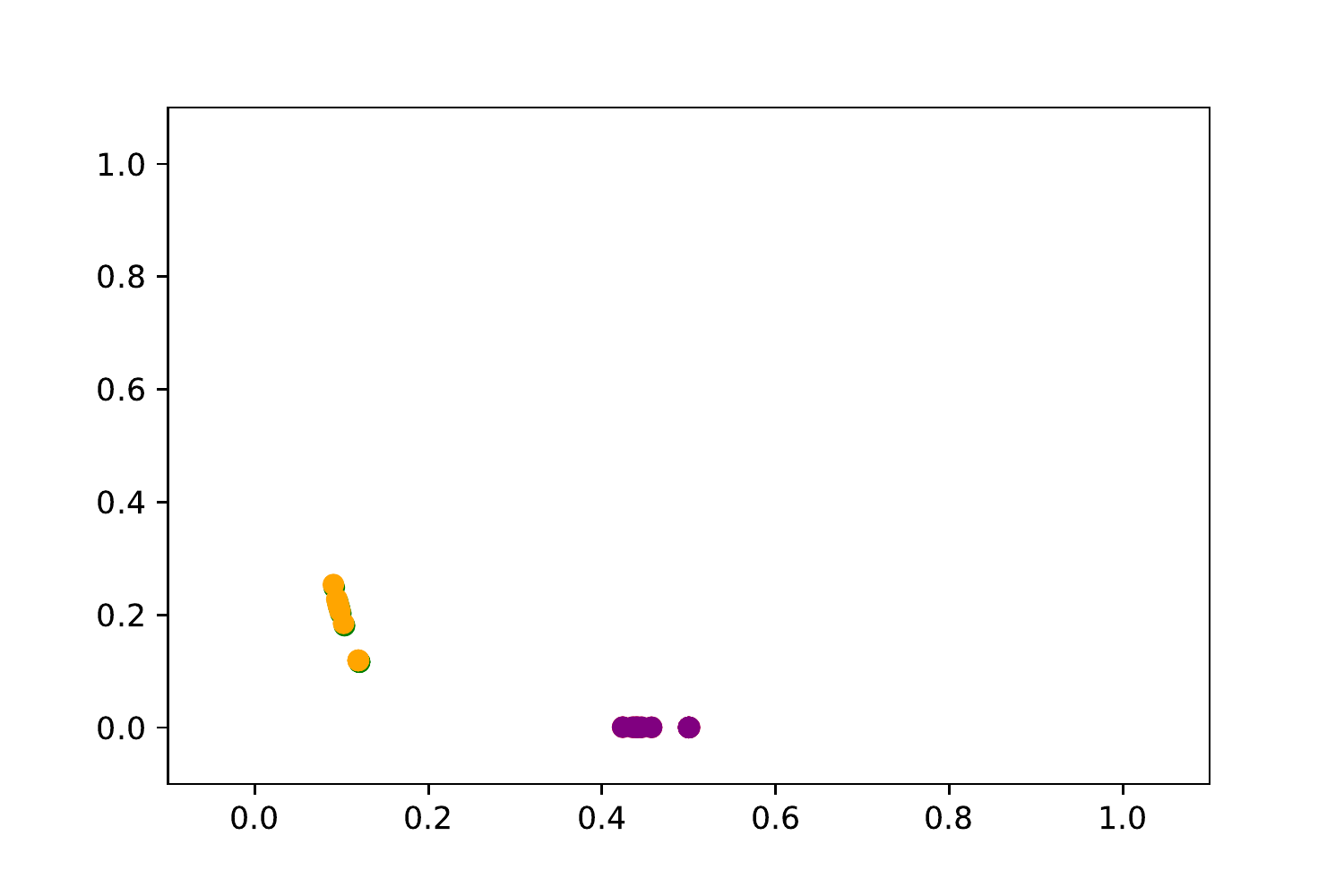}
       }
\end{tabular}
\caption{ \ref{fig:inclined_plane} During training in the Inclined Plane Domain the yellow and the purple object are seen only on the left slope, whereas the red and the green object are seen on both slopes. Colors and x-coordinates are given as input. Neural NID learns an abstract representation where the green and the yellow object are grouped together and the purple and the red one, therefore generalizing well to tests where the yellow and purple objects are seen on the right slope. \ref{fig:main_compound_train_valley} All methods achieve low training error. \ref{fig:main_compound_test_valley} Only Neural NID achieves low test error.
\ref{fig:lambda_2_silohuette_main} Configurations achieving low test error tend to have high Silhouette scores, suggesting that successful generalization depends on clear clustering. Without entropy regularization (\ref{f_enc_1}) the abstract representations of the green and the yellow object at different x-coordinates are usually not grouped, whereas they are with entropy regularization (\ref{f_enc_4}; yellow and purple dots are almost entirely covering green and red dots, respectively).}
\label{fig:main_curves}
\end{figure}
In our experiments with the Inclined Plane Domain,
we seek an empirical answer to the two following questions: (i) Can Neural NID effectively generalize in the setting described in Figure \ref{fig:inclined_plane}?
(ii) Is the generalization of Neural NID connected to learning abstract properties as conjectured in the introduction?
To address the first question we train Neural NID with different trajectories of objects of different color starting at different initial x-coordinates.
Some of the objects appear in the training set only on one of the two slopes, whereas others appear on both sides of the slope.
After that, we test the model on rollouts sampled from all possible initial conditions, i.e. all objects on all slopes.
The cumulative error of the predictions are shown in Figure \ref{fig:main_compound_train_valley}.
Neural NID attains the same performance as standard baselines on the training set (MLP, CNN with 1 and 3 layers).
Generalization to the test set is by far best for Neural NID (Fig.~\ref{fig:main_compound_test_valley}; see also Appendix).

To answer our second question we look at the correlation between the cumulative error on the test set and the Silhouette score \cite{ROUSSEEUW198753} attained by the clustering algorithm that assigns the labels corresponding to the three logic categories needed to explain the next state of the system, i.e, $C_1$ is the cluster of learned representations of objects that do not roll, $C_2$ the cluster of representations of rollable objects on the left slope, and, $C_3$ the cluster of representations of rollable objects on the right slope.
The Silhouette score is computed on the points generated by $f^{\mathrm{enc}}$ for different $o$ and $p$.
These points are plotted in Figures \ref{f_enc_1}, \ref{f_enc_4}.
It can be seen how in presence of regularization (Figure \ref{f_enc_4}) objects with similar properties are grouped in dense clusters.
This allows better generalization as can be seen e.g. in Fig.~\ref{fig:lambda_2_silohuette_main} that shows how the cumulative test error at the end of the rollout tends to be low for Silhouette values that approach $1$, i.e. when clustering works well. We find for regularization constants $\lambda_1\approx 5e{-}8$ and $\lambda_2\approx 5e{-}6$ that many simulations reach a high Silhouette score above 0.8. For these high Silhouette scores the test error is usually very low. The choice of the number $K$ of abstract feature vectors  seems less crucial. 

\section{Conclusions}
We have empirically shown that Neural ND achieves successful out-of-distribution generalization in a toy setting. Remarkably, for out-of-distribution generalization Neural NID does not need features  like the object shape because the relevant abstract features are learned by experience. The Neural NID framework may be general enough to be applied on top of more complex architectures for model based Reinforcement Learning~\cite{Kulkarni19,zambaldi2018relational,zhengyao2019neural,kipf2020contrastive}. Furthermore, Neural NID ideas may be useful to learn abstract representations of PDDL domains~\cite{silver2020pddlgym} from raw observations, without the need of specifying predicates.
{\small
\bibliography{neurips_workshop_2021}
\bibliographystyle{plain}
}
\appendix
\newpage
\section{Architecture of the transition map}
\label{sec:conv_filters}
As mentioned, an essential ingredient of Neural NID is a collection of possible outcome functions $\{ \Omega_z: \mathbb{R}^{|\mathcal{O}|\times D} \rightarrow \mathbb{R}^{|\mathcal{O}|\times D} \}$. In modeling the dynamics of a relational environment is reasonable to assume that each object is subject only to a local transformation across consecutive steps at a fine enough time scale. Therefore, under the assumption of fine enough time scale, the use of convolution for modeling the possible outcomes is a well motivated inductive bias.
In formulas, we have:
\begin{equation}
    \Omega_z(\mathbf{x}_t) [o, p] = \sum_{l =-S_2}^{S_2} \mathbf{\omega}_z[l] \mathbf{x}_t[o, p - l]
\end{equation}
where the convolutional kernels $\mathbf{\omega}_z$ are shared across the object dimension and have size $(2S_2 + 1)$. Furthermore, it is implicitly assumed that $\mathbf{x}_t[o, p - l]$ is properly zero padded so that the dimension of $\Omega_z(\mathbf{x}_t) $ is equal to the dimension of the input $\mathbf{x}_t$.
\section{Additional experiments on Inclined Plane}
\begin{figure}[] 
\centering
\begin{tabular}{cc}
\subfloat[\textbf{Inclined Plane: }Compound Errors on Train Set (Avg. 10 seeds) \label{fig:compound_train_inclined_plane}]{%
       \includegraphics[width=0.5\linewidth]{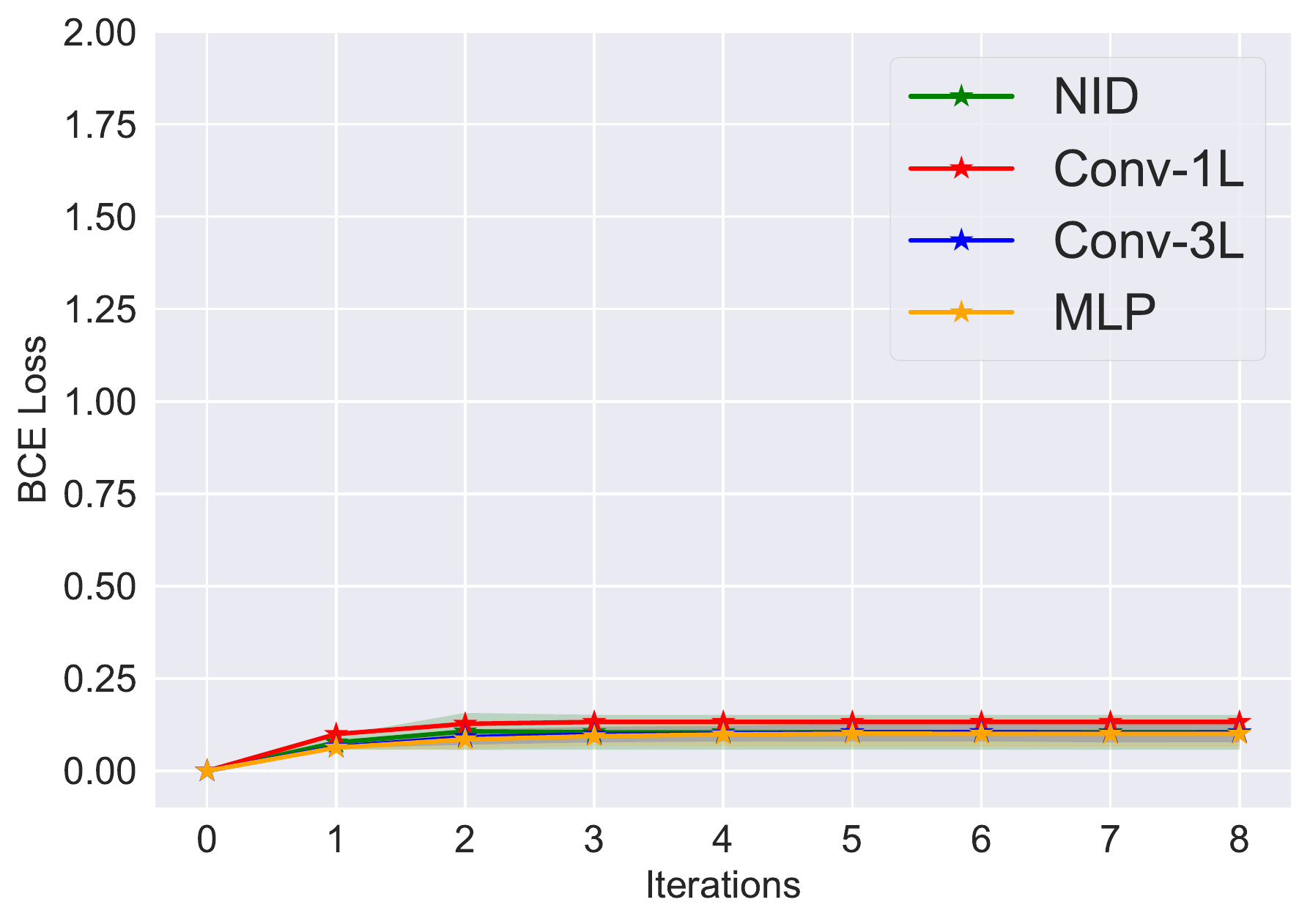}
     } &
\subfloat[\textbf{Inclined Plane: }Compound Errors on Test Set (Avg. 10 seeds) \label{fig:compound_test_inclined_plane}]{%
       \includegraphics[width=0.5\linewidth]{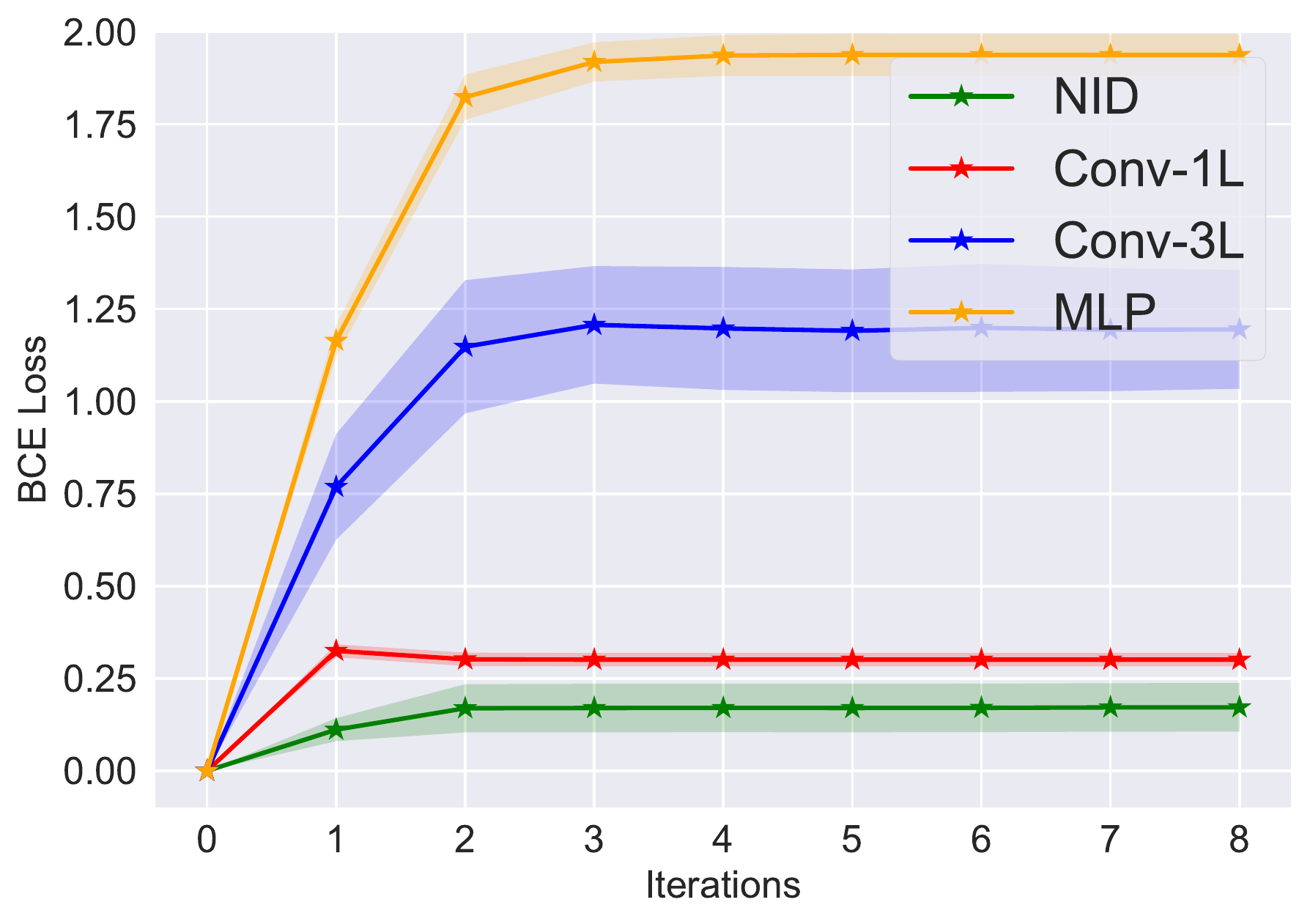}
       } \\
\end{tabular}
\caption{Rollouts and corresponding cumulative errors for the four models on the training set and test set respectively. It can be noticed that while all the models perform similarly on the train set, only our NID generalizes acceptably on the test set. For each seed, we compute the mean of the BCE Loss across $100$ rollouts of $8$ steps. Then we compute the average and the standard deviation across $10$ different seeds to obtain the mean values and the standard deviations reported in the plot.}
\label{fig:curves}
\end{figure}

\begin{figure}[t]
    \centering
\begin{tabular}{cccc}
\subfloat[$t=0$]{%
       \includegraphics[width=0.225\linewidth]{ 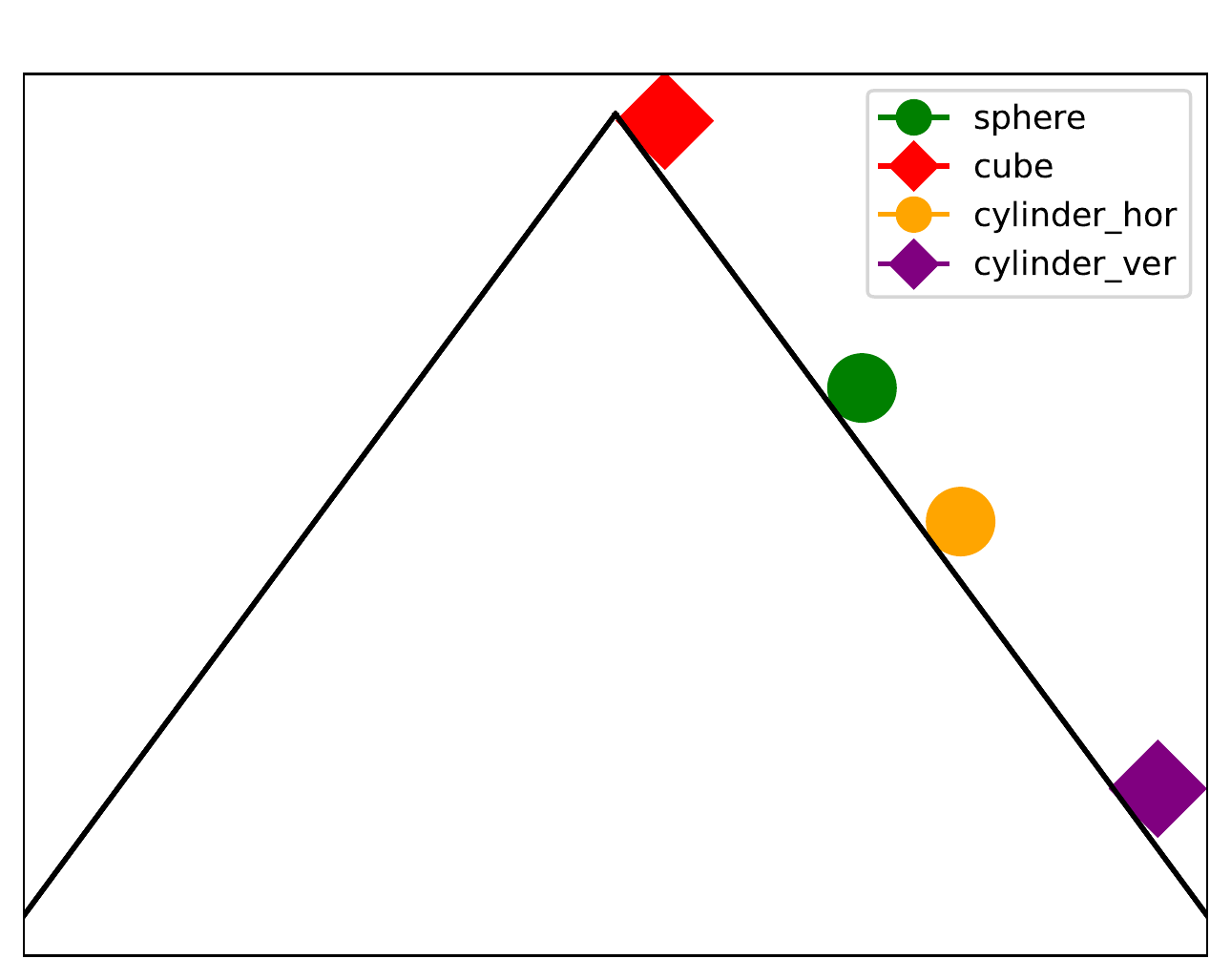}
       }&
\subfloat[NID $t=1$ \label{fig:NID_roll_0}]{%
       \includegraphics[width=0.225\linewidth]{ 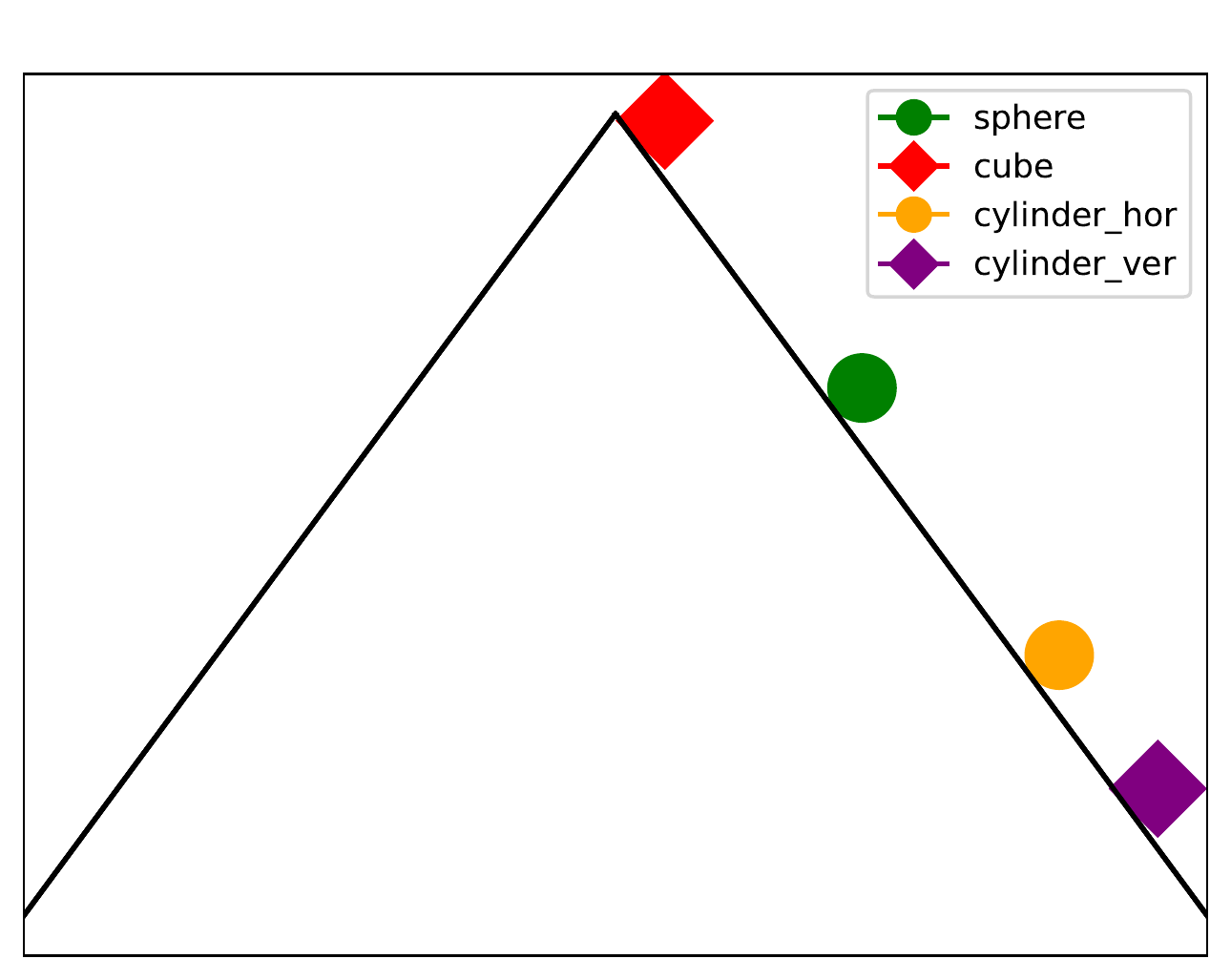}
     } &
\subfloat[NID $t=2$ \label{fig:NID_roll_1}]{%
       \includegraphics[width=0.225\linewidth]{ 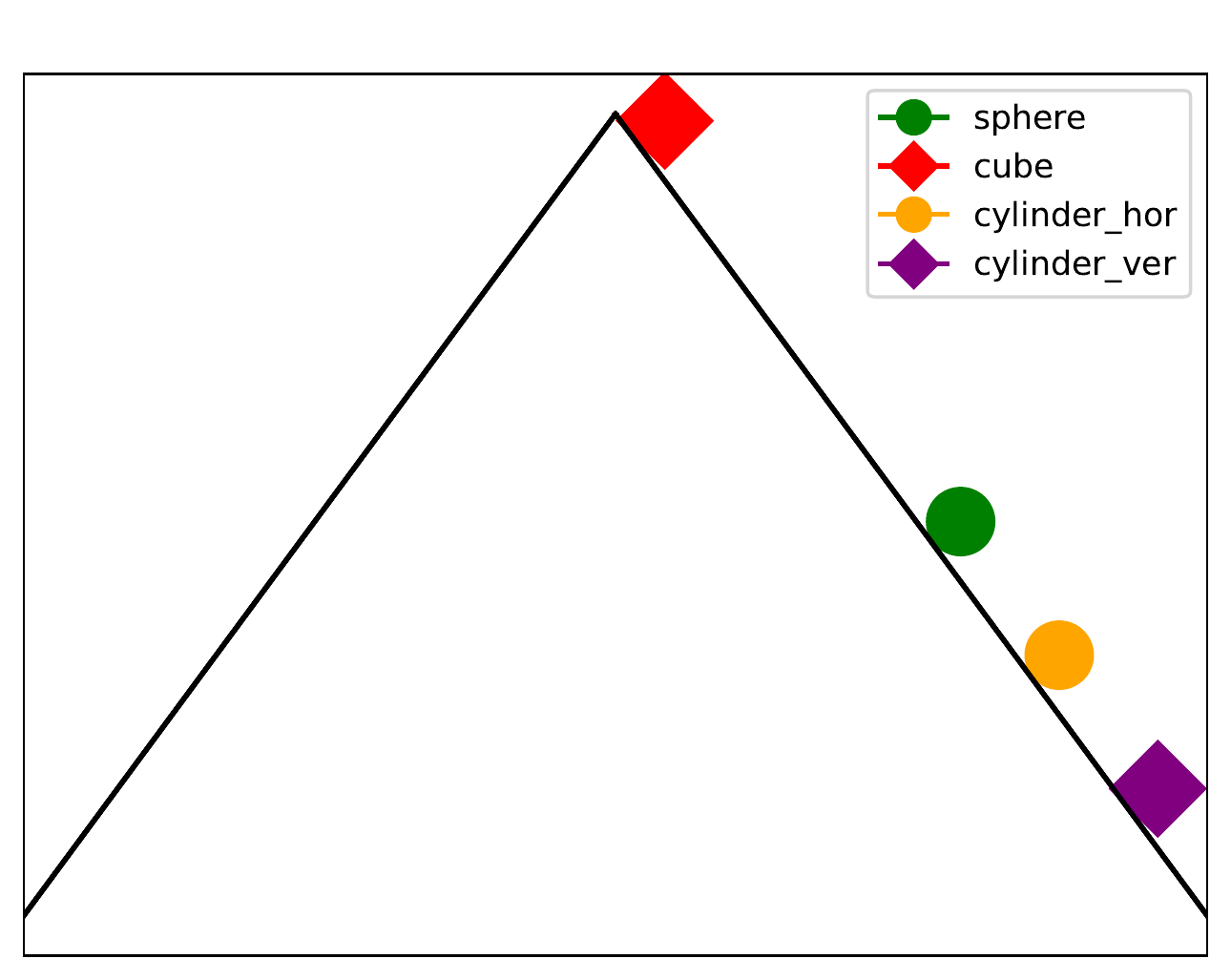}
       } &
\subfloat[NID $t=3$ \label{fig:NID_roll_2}]{%
       \includegraphics[width=0.225\linewidth]{ 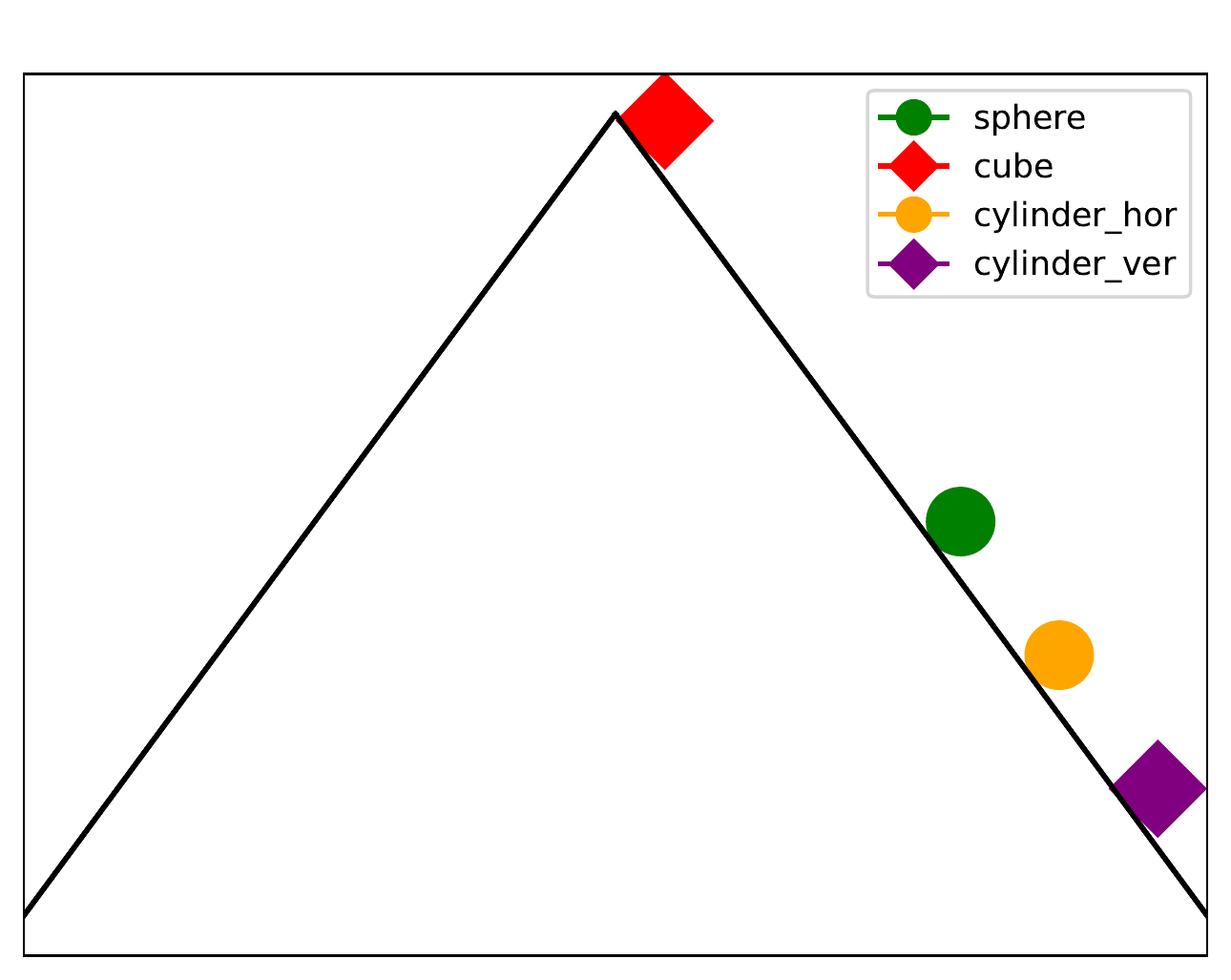}
       } \\
       \subfloat[$t=0$]{%
       \includegraphics[width=0.225\linewidth]{ frames/InclinedPlane/start.pdf}
       }&
       \subfloat[MLP $t=1$ \label{fig:MLP3L_roll_0}]{%
       \includegraphics[width=0.225\linewidth]{ 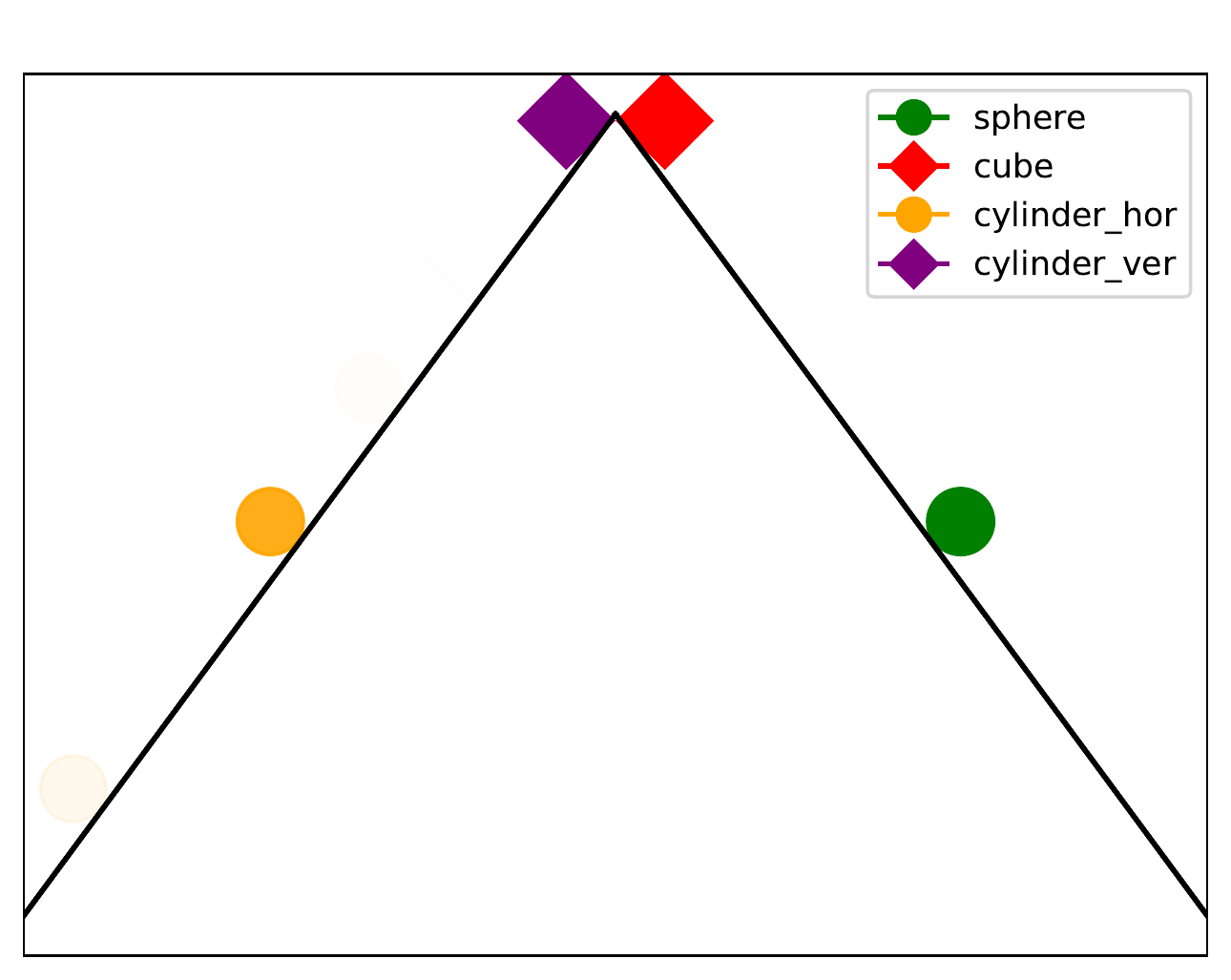}
     } &
\subfloat[MLP $t=2$ \label{fig:MLP3L_roll_1}]{%
       \includegraphics[width=0.225\linewidth]{ 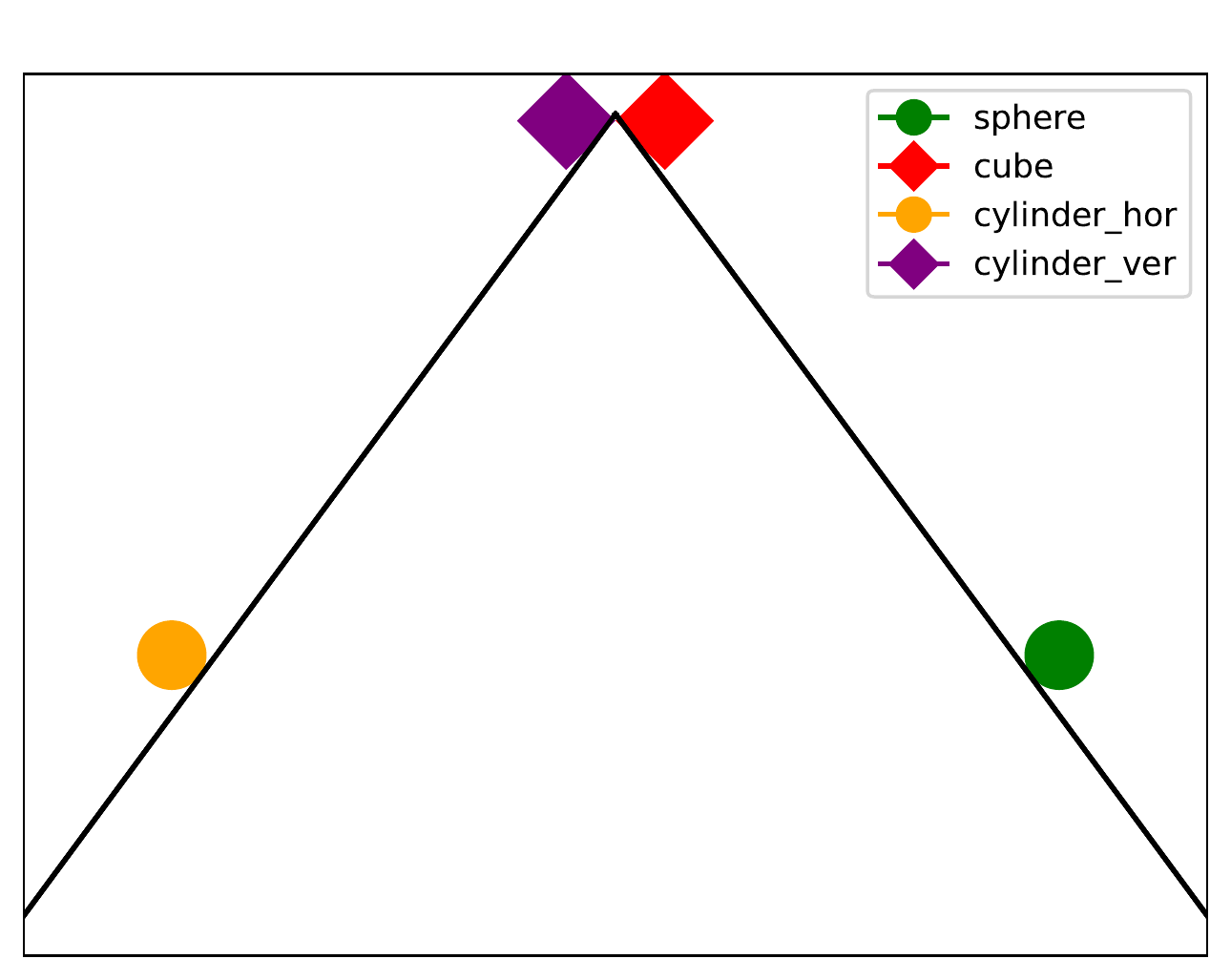}
       } &
\subfloat[MLP $t=3$ \label{fig:MLP3L_roll_2}]{%
       \includegraphics[width=0.225\linewidth]{ 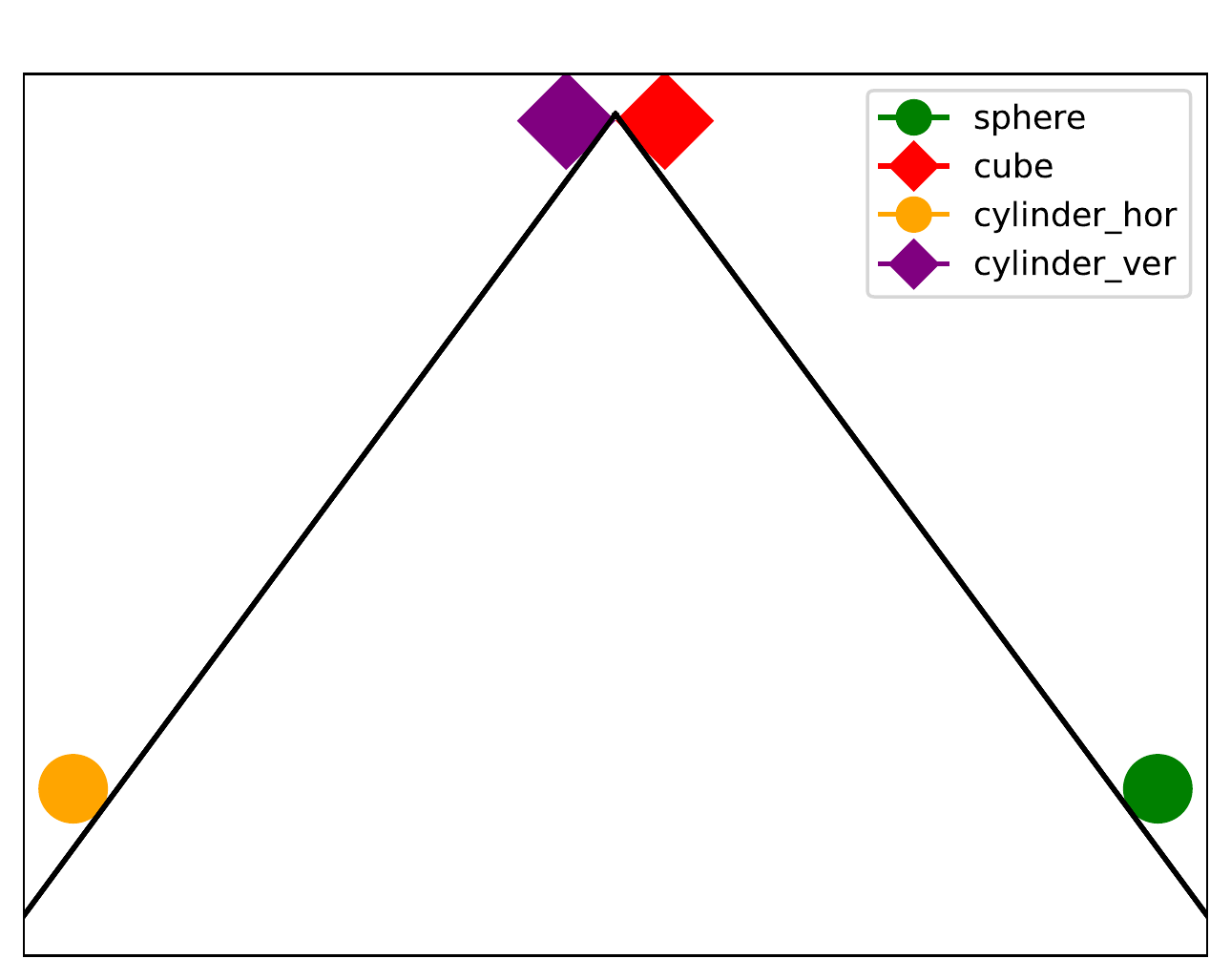}
       } \\
       \subfloat[$t=0$]{%
       \includegraphics[width=0.225\linewidth]{ frames/InclinedPlane/start.pdf}
       }&
       \subfloat[Conv-1L $t=1$ \label{fig:CONV1L_roll_0}]{%
       \includegraphics[width=0.225\linewidth]{ 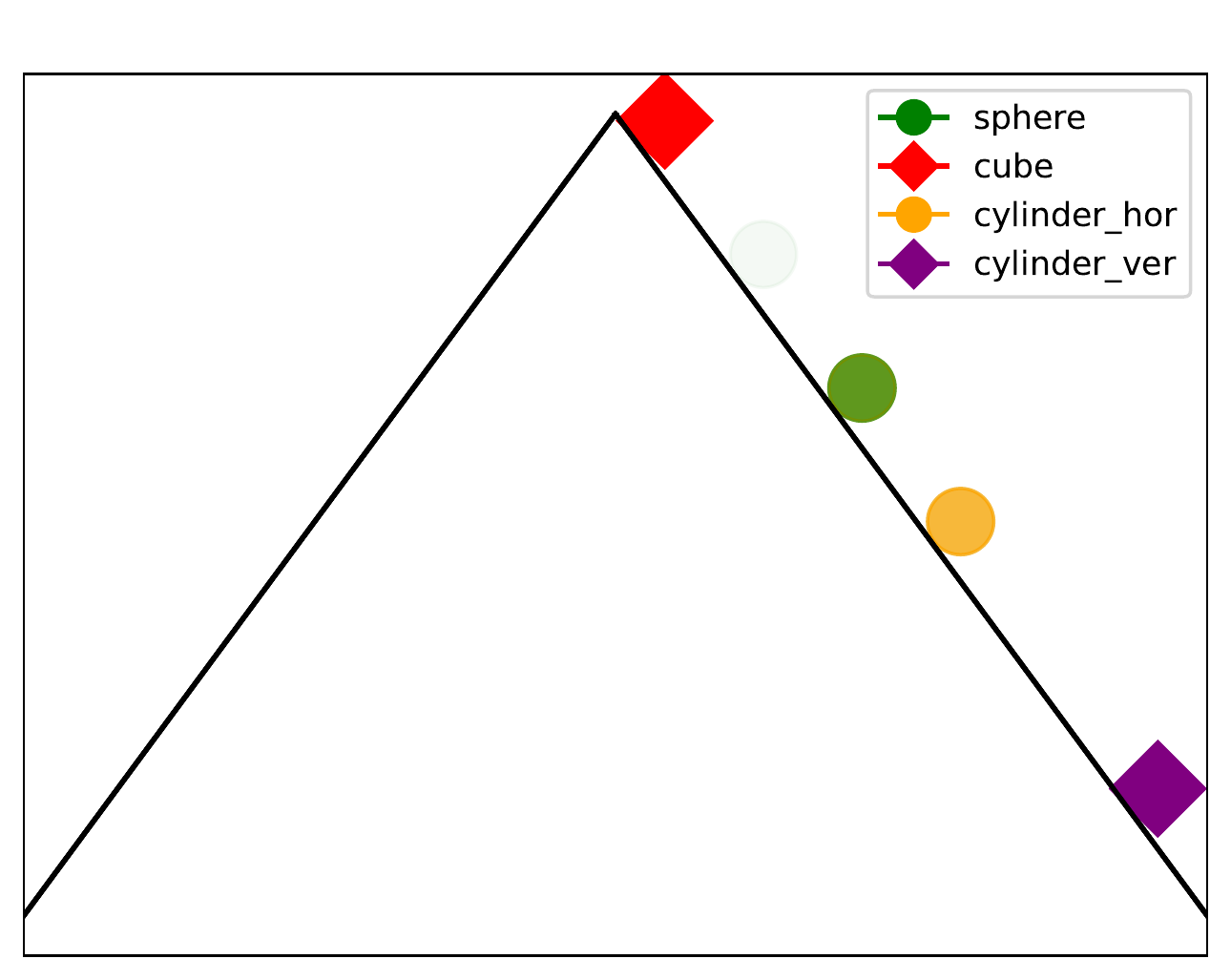}
     } &
\subfloat[Conv-1L $t=2$ \label{fig:CONV1L_roll_1}]{%
       \includegraphics[width=0.225\linewidth]{ 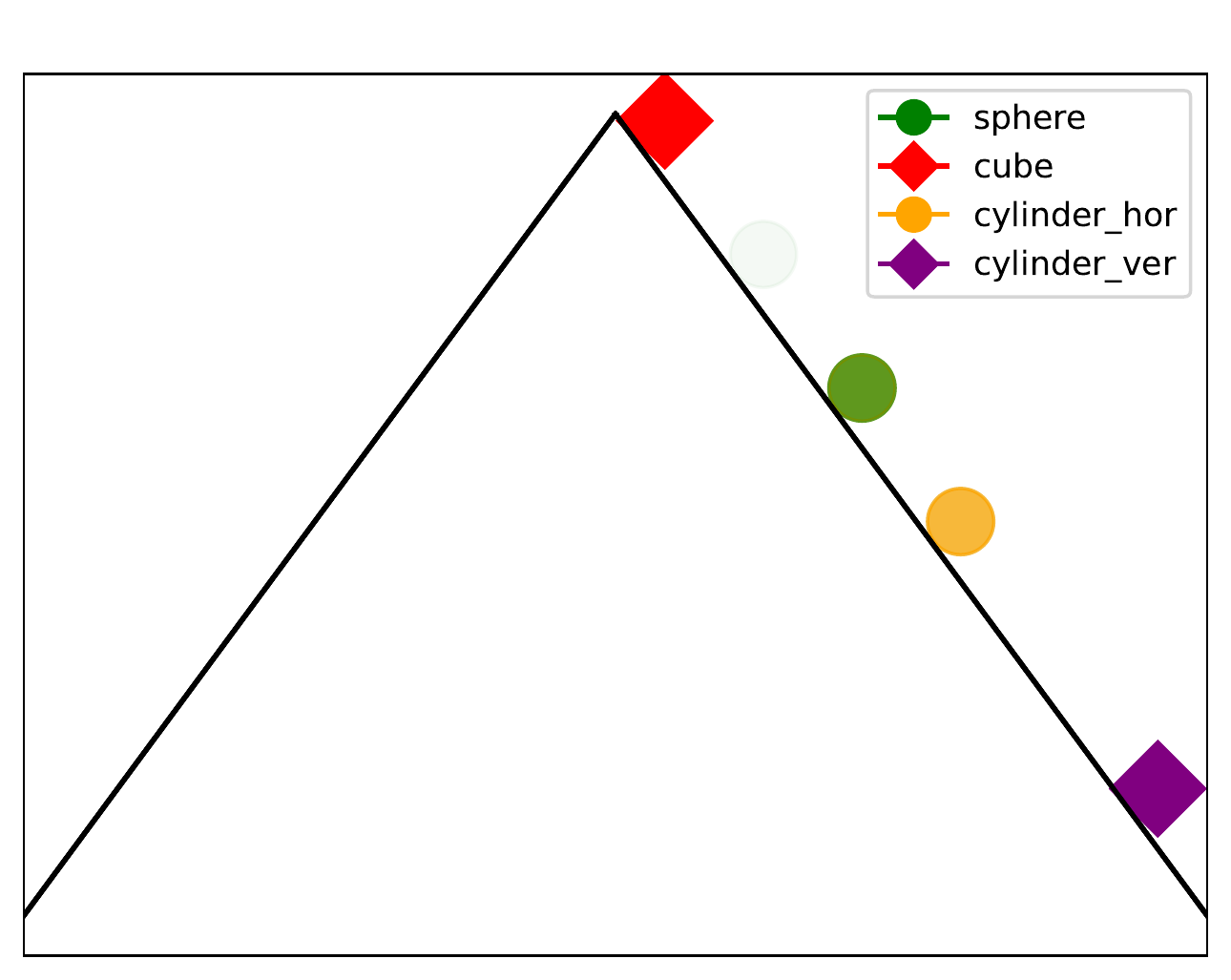}
       } &
\subfloat[Conv-1L $t=3$ \label{fig:CONV1L_roll_2}]{%
       \includegraphics[width=0.225\linewidth]{ 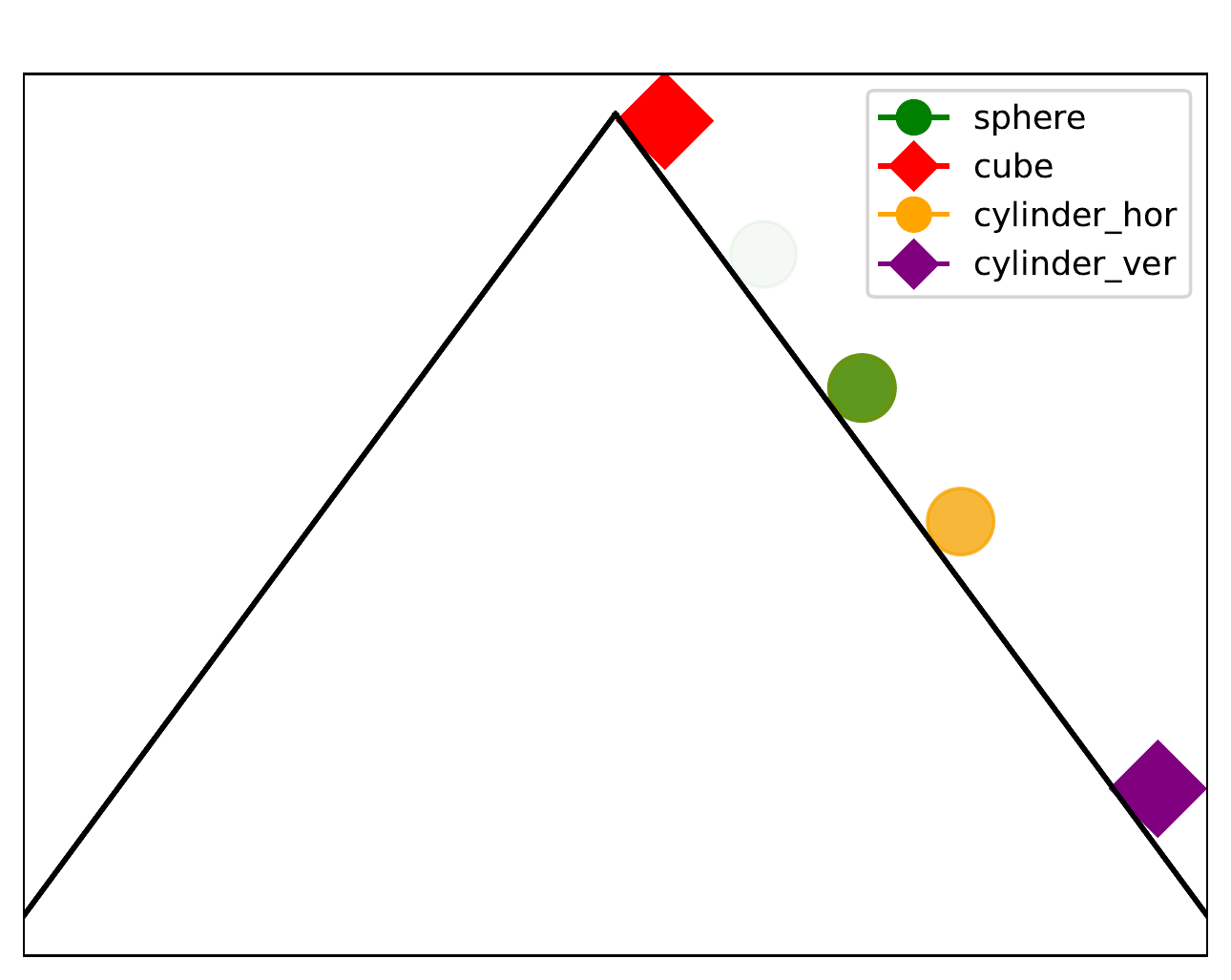}
       } \\
       \subfloat[$t=0$]{%
       \includegraphics[width=0.225\linewidth]{ frames/InclinedPlane/start.pdf}
       }&
       \subfloat[Conv-3L $t=1$ \label{fig:CONV3L_roll_0}]{%
       \includegraphics[width=0.225\linewidth]{ 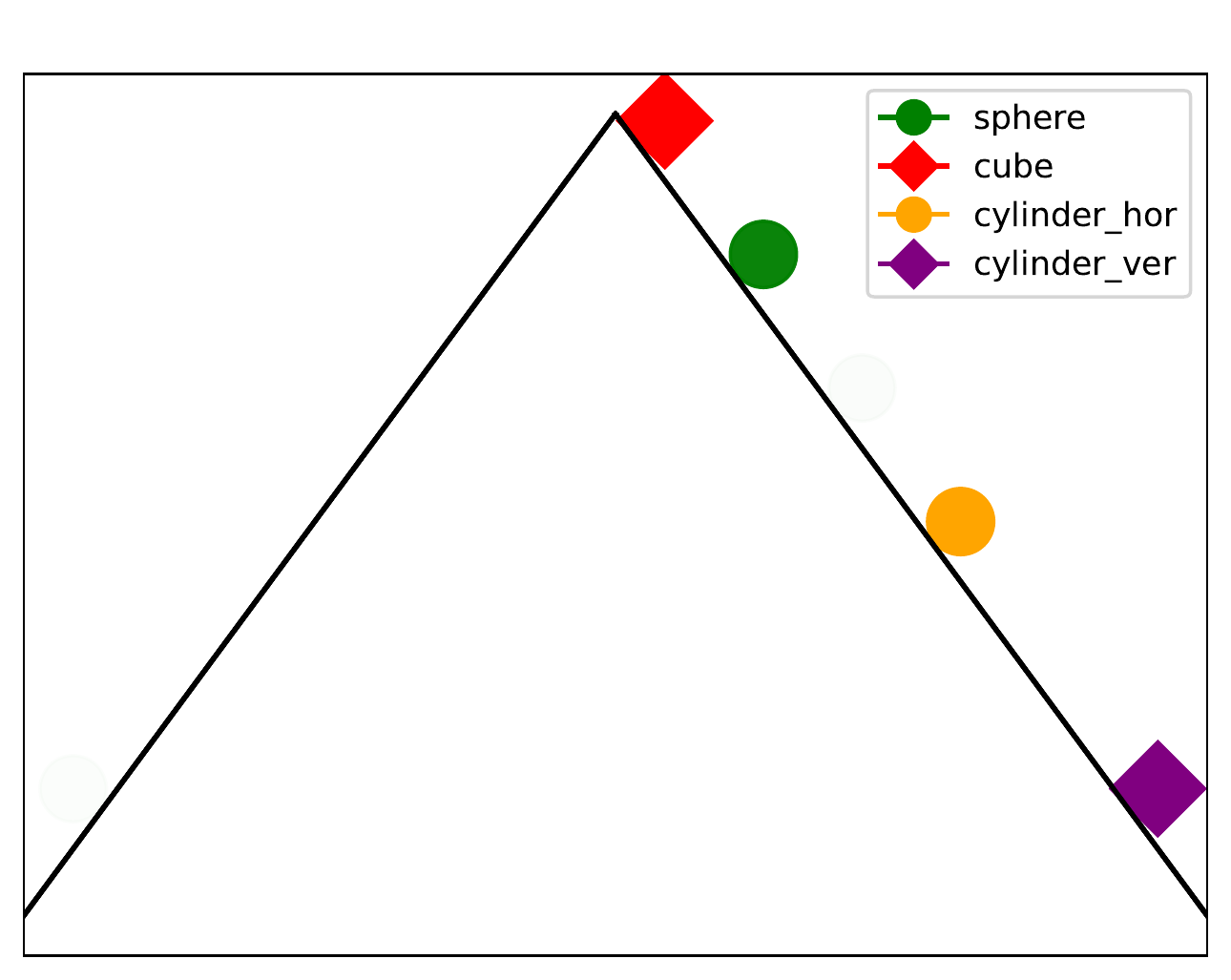}
     } &
\subfloat[Conv-3L $t=2$ \label{fig:CONV3L_roll_1}]{%
       \includegraphics[width=0.225\linewidth]{ 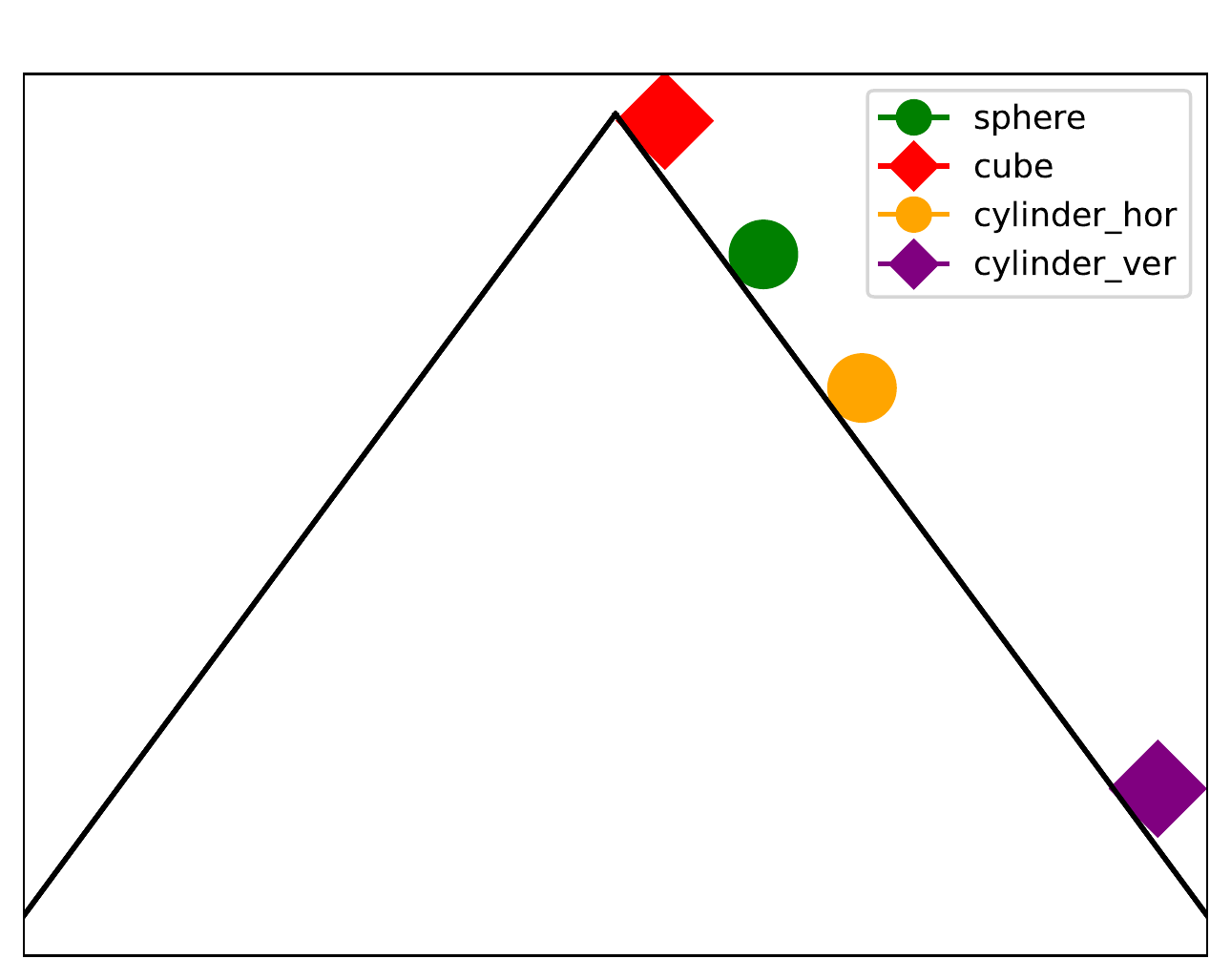}
       } &
\subfloat[Conv-3L $t=3$ \label{fig:CONV3L_roll_2}]{%
       \includegraphics[width=0.225\linewidth]{ 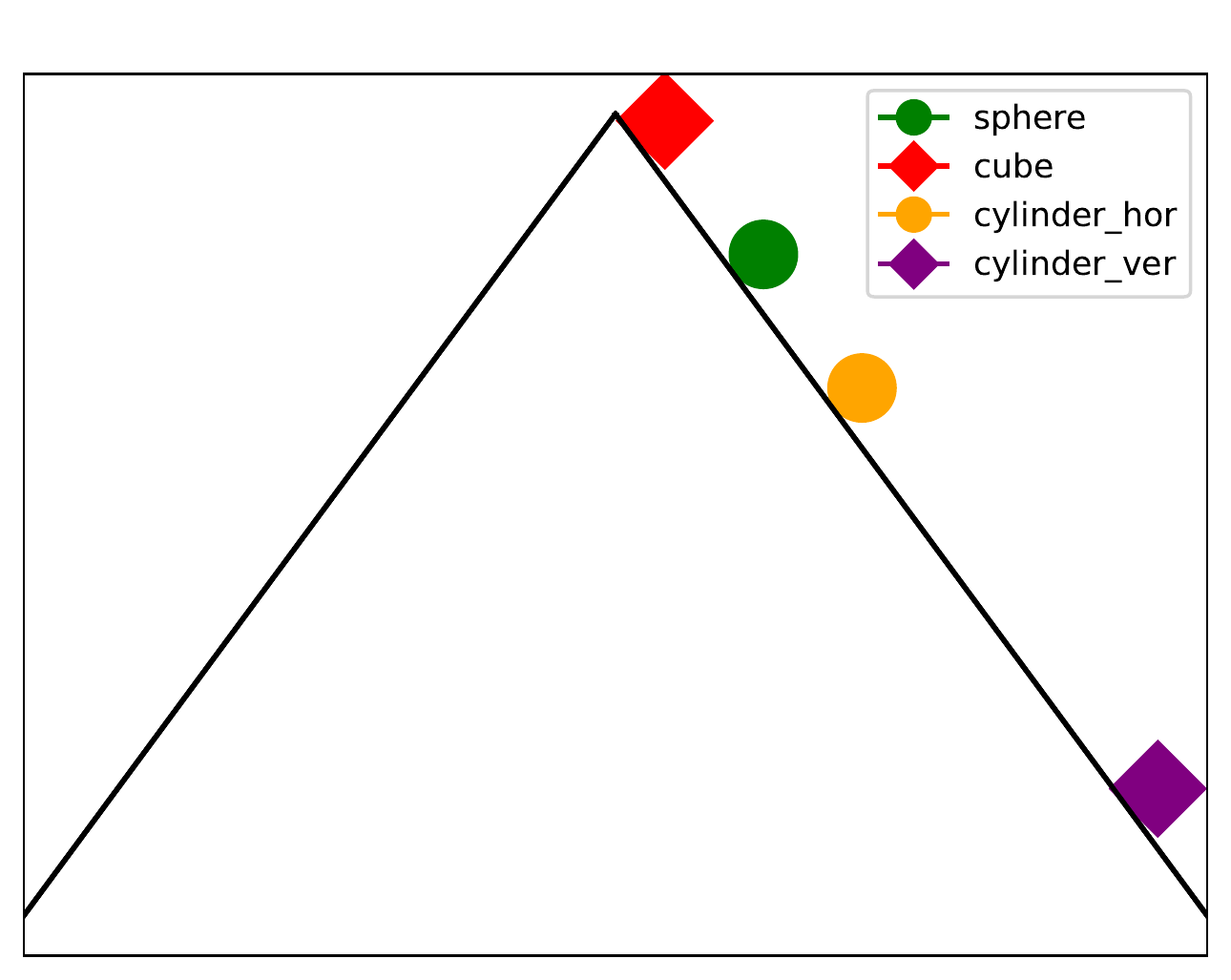}
       } \\
\end{tabular}
\caption{Visual inspection of the models predictions on the test set for \emph{Inclined Plane}. The first row presents the rollout simulated by our model. It can be seen that correctly capture the downhill motion of the yellow object that was not seen on that slope during training. The other rows report the evolution predicted by the baselines that are wrong for various reasons. For example, the convolutional network with only one layer assumes full invariance so it is not expressive enough to model the two possible directions of motion. However, if we add some flexibility like $2$ extra convolutional layers with all to all connections across the channels, we get the result in the last row that predict the yellow and green object to move upwards. Finally, the MLP predictions in the second rows move the purple and yellow objects on the left slope where they were located during the training phase.}
\label{fig:rollouts_inclined_plane}
\end{figure}
Our first experiment is on a task that we introduce with this work called \emph{Inclined Plane}. It has been designed to specifically test the generalization across objects capability of Neural NID.
In particular, we have some objects on an inclined plane. Some of them are spheres and consequently can roll down while others stay in place. In addition, the environment presents two slopes that cause the spheres to roll on different directions according to their positions. One of these objects, the yellow one in Figure \ref{fig:inclined_plane}, is presented only on the left slope during the training while it is located only on the right during testing. Figure \ref{fig:inclined_plane} depicts a sample from the training set in the left column and a sample from the test set on the right column.

Generalizing the behaviour of the yellow objects across slopes would be natural for humans. Pointing to the fact that the yellow object rolls on the left slope, they would associate the property \emph{can roll} to it. Leveraging on this property they would predict that the yellow object moves in the opposite direction when it is left on a slope with opposite slope.

Despite its simplicity, \emph{Inclined Plane} is a challenging task for a forecaster since the correct generalization requires to reason about:
\begin{enumerate}
    \item The invariance breaking point when the slope changes direction
    \item Abstract properties of the objects, e.g. the fact that some objects can roll and others can not.
    \item Relations among them, e.g. understand that objects can not pass through each other.
\end{enumerate}

\begin{figure}[t]
\includegraphics[width=0.5\linewidth]{ 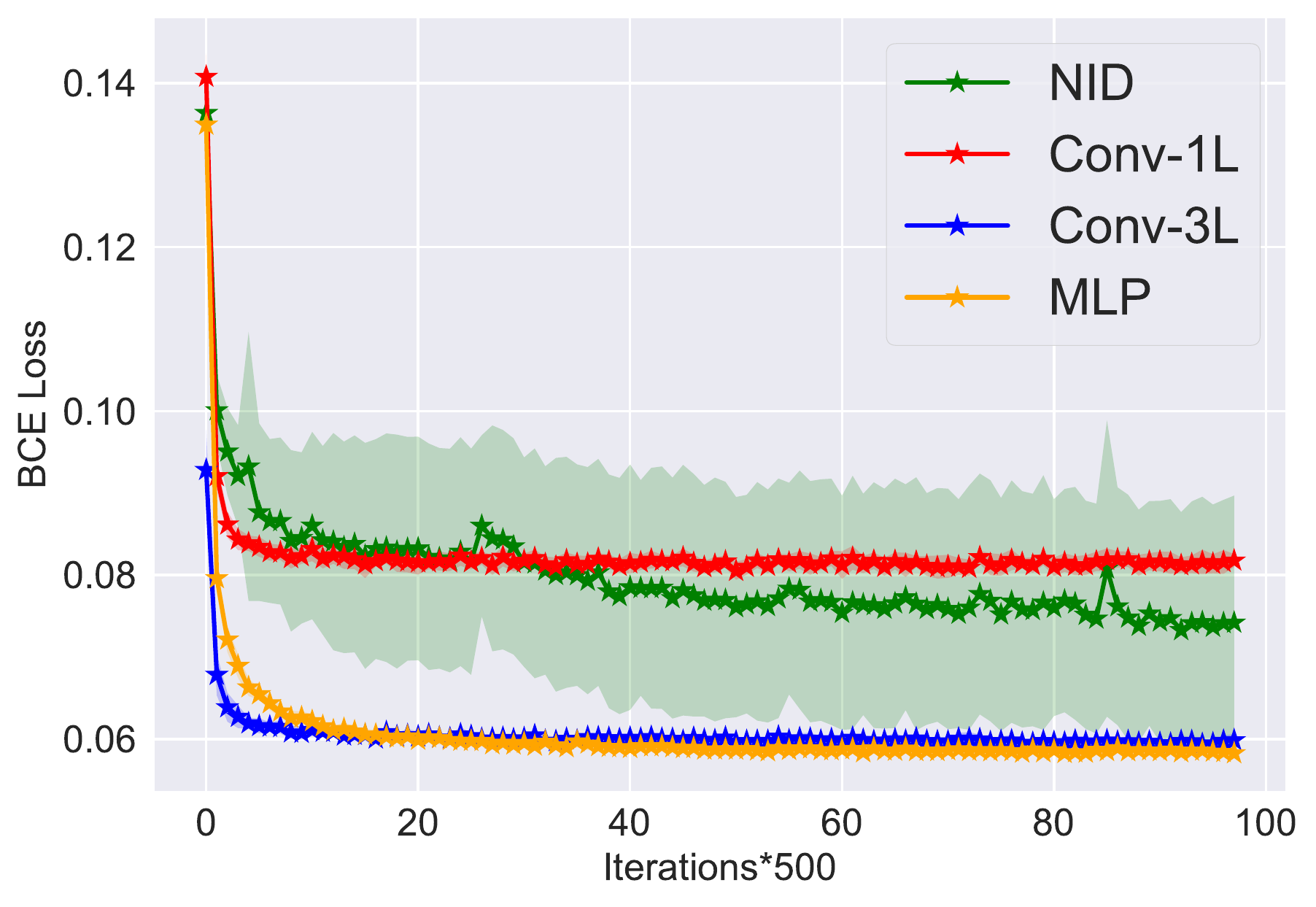}
\caption{ \label{fig:inclined_plane_learning} Learning curves for our NID in green against the three proposed baselines. We can see that our model is difficult to optimize and suffers from high variance. The learning curves are averaged across $10$ seeds and averaged in bins of $500$ steps.}
\label{fig:learning}
\end{figure}

The results in Figure \ref{fig:learning} show that our model suffers from higher variance during training when compared to the three baselines. Also, the mean value of the learning curve converges slower, suggesting that our architecture is harder to optimize. However, looking at Figure \ref{fig:compound_test_inclined_plane} we show the error of the simulated evolution of the system under the test set initial distribution. Notice instead that in Figure \ref{fig:compound_train_inclined_plane} that reports the same experiment executed on the training set, the model performance  is much closer.
It can be seen that despite having learned to simulate the evolution of the training set, CNNs and MLPs always fail in correctly generalizing while Neural NID succeed in this case.

 Figure \ref{fig:rollouts_inclined_plane} allows to visually inspect the performance difference reported in Figure \ref{fig:compound_test_inclined_plane}. In particular, the first row presents the rollout simulated by our Neural NID. It can be seen that it correctly captures the downwards motion of the yellow object that was not seen on that slope during training. The other rows report the evolution predicted by the baselines that are wrong for various reasons. For example, the convolutional network with only one layer assumes full invariance so it is not expressive enough to model the two possible directions of motion. However, if we add some flexibility like $2$ additional convolutional layers with all to all connections across the channels, we get the result in the last row that predict the yellow and green object to move upwards. Finally, the MLP predictions in the second rows move the purple and yellow objects on the left slope where they were located during the training phase.

\section{Inclined Plane with Agent}
We have argued that with Neural NID we aim to propose a new framework for model inference in model-based reinforcement learning. That is, Neural NID should be able to model the consequences that external actions have on the environment. The task \emph{Inclined Plane} allow to certify Neural NID capability to predict how the object relations and the environment physics affect the evolution of the system. However, we did not consider an agent acting in those environments. Therefore, we introduce  an agent and we obtain a new environment that we call \emph{Inclined Plane with Agent}.

The agent is represented exactly as ordinary objects are represented, i.e. assigning it an index $o_{agent}: 1 \leq o_{\mathrm{agent}} \leq |\mathcal{O}| $ and encoding its position in the environment as a slice of the input tensor denoted as $\mathbf{x}_t[o_{\mathrm{agent}}]$. However, an architectural change is needed to take actions into account. We choose to modify the input of the decoding function $f^{\mathrm{dec}}$ as follows:
\begin{equation}
    f^{\mathrm{dec}}[o,p,a] = P( z_t | \mathbf{x}_t, o, p, a) = \mathrm{softmax}_{2}(\mathrm{MLP}(\mathrm{Concat}(\mathbf{v}_t[o,p], \mathbf{e}_a)))
    \label{eq:actions}
\end{equation}
The agent can choose between four discrete actions: \emph{Move left without grabbing}, \emph{Move right without grabbing}, \emph{Move left while grabbing}, \emph{Move right while grabbing}.
The agent can move to a position even if it is occupied by another object. Then if it takes one of the two \emph{grabbing} actions from that position, the other object moves with the agent.

\begin{figure}[h!]
    \centering
\begin{tabular}{cccc}
       \subfloat[NID $t=5$]{%
       \includegraphics[width=0.225\linewidth]{ 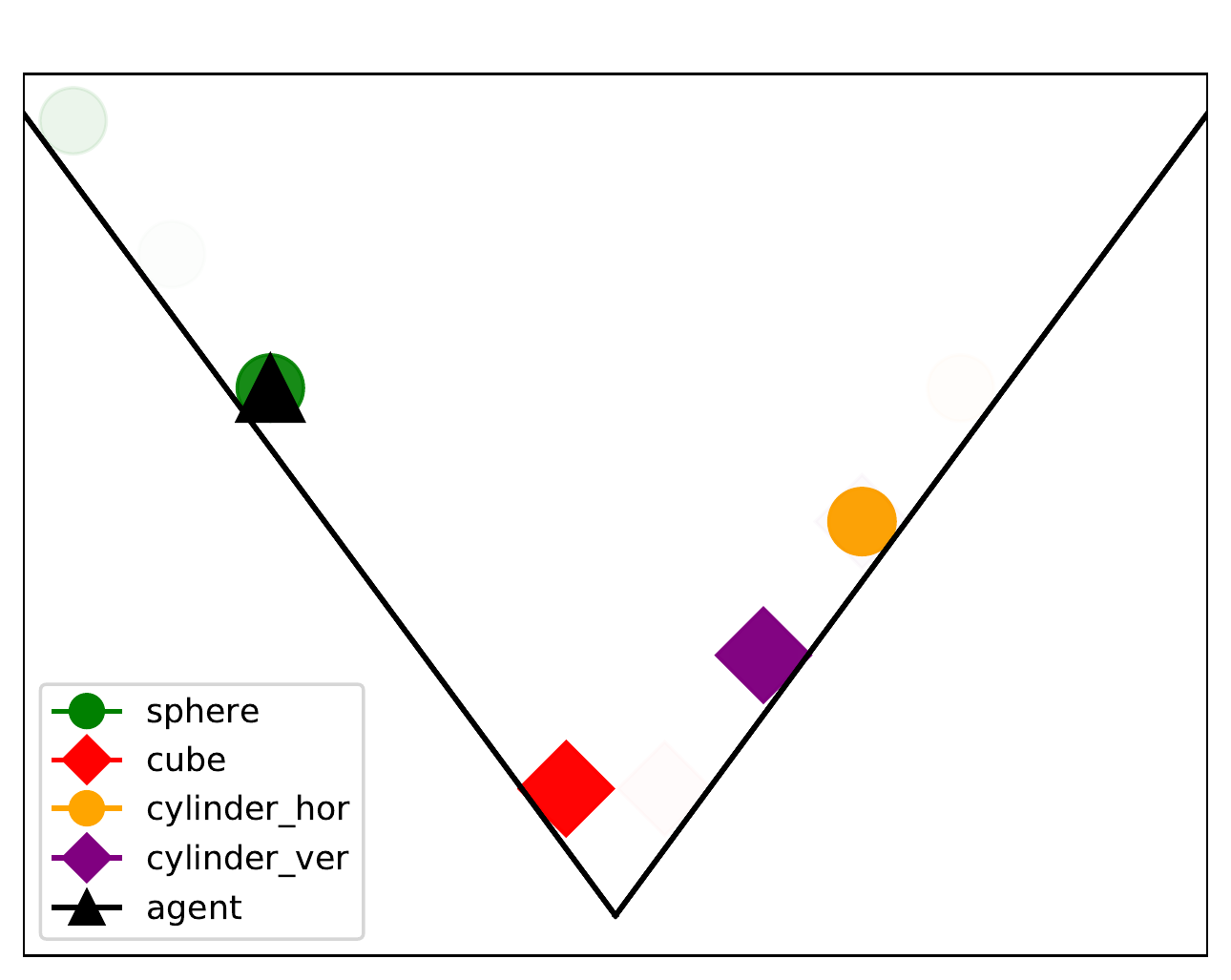}
       } &
       \subfloat[NID $t=6$]{%
       \includegraphics[width=0.225\linewidth]{ 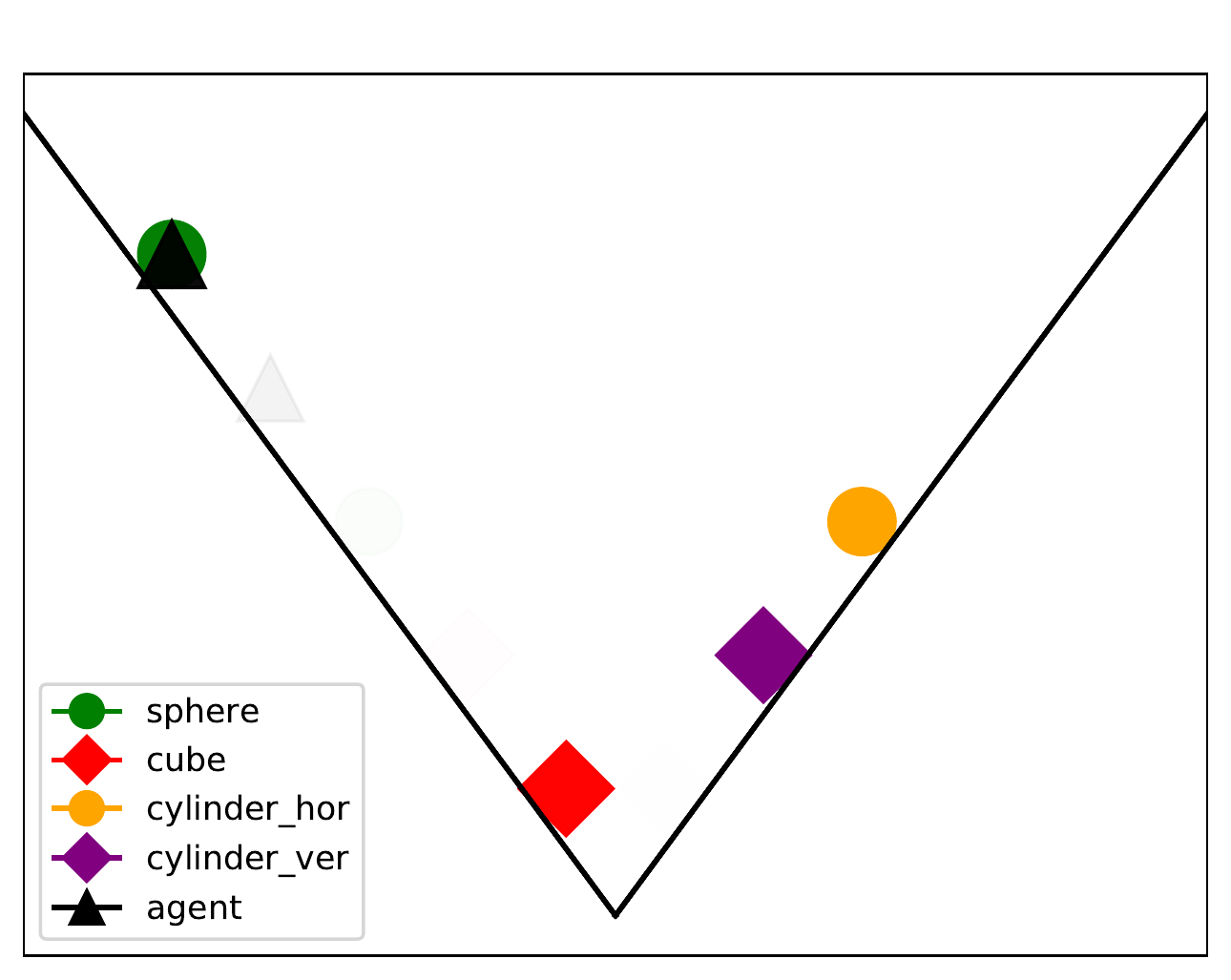}
       } &
       \subfloat[NID $t=7$]{%
       \includegraphics[width=0.225\linewidth]{ 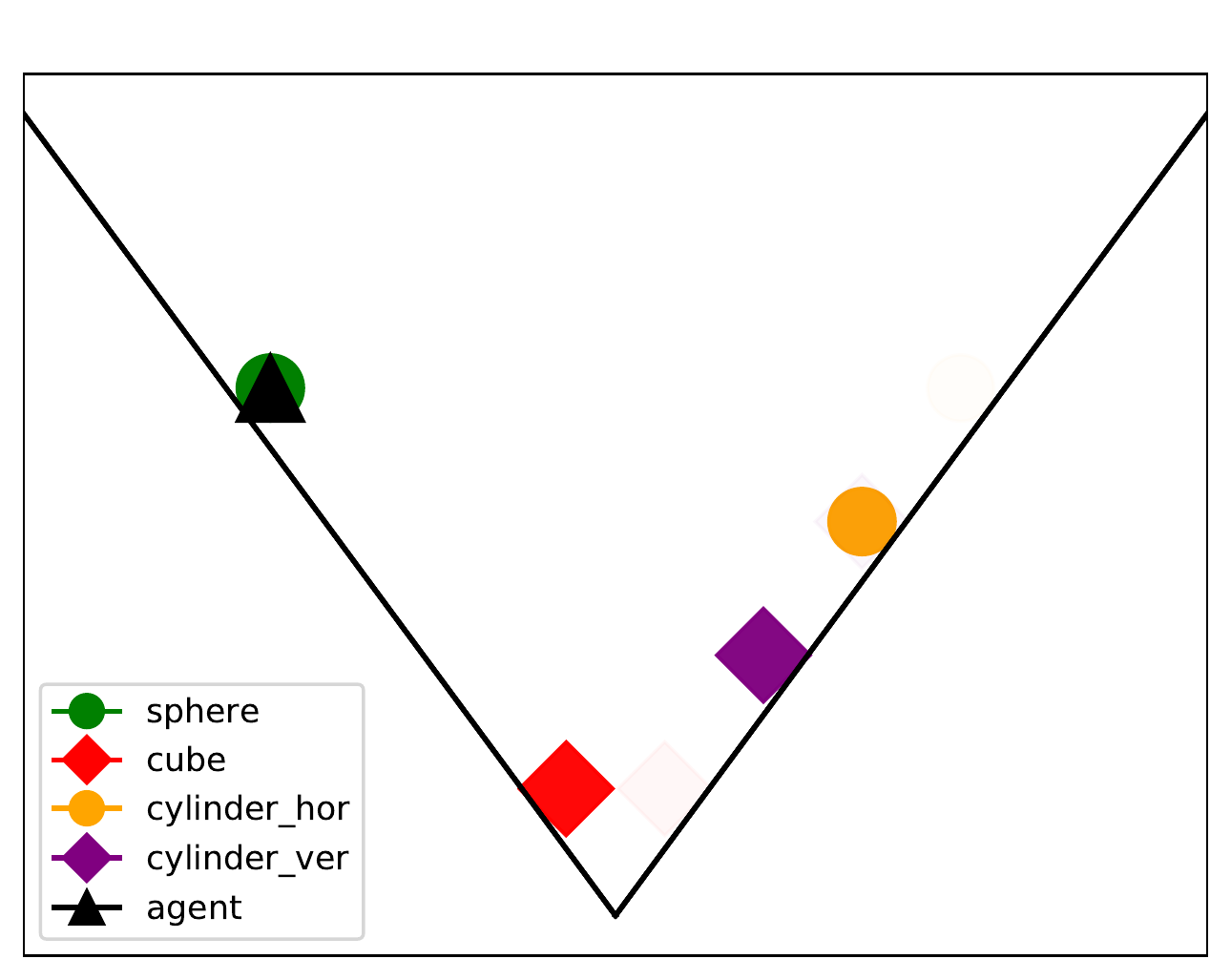}
       } &
       \subfloat[NID $t=8$]{%
       \includegraphics[width=0.225\linewidth]{ 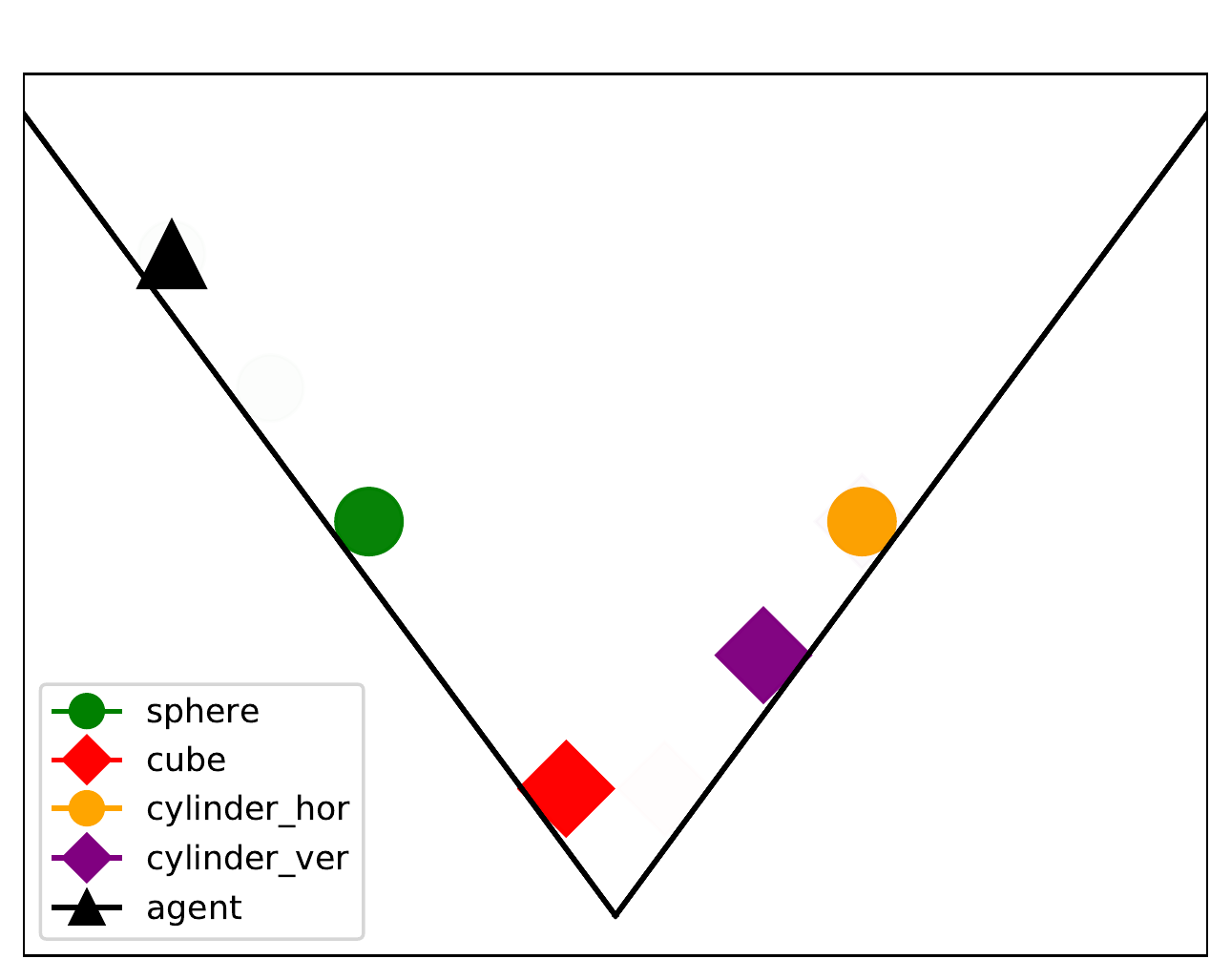}
       } \\
\subfloat[Truth $t=5$]{%
       \includegraphics[width=0.225\linewidth]{ 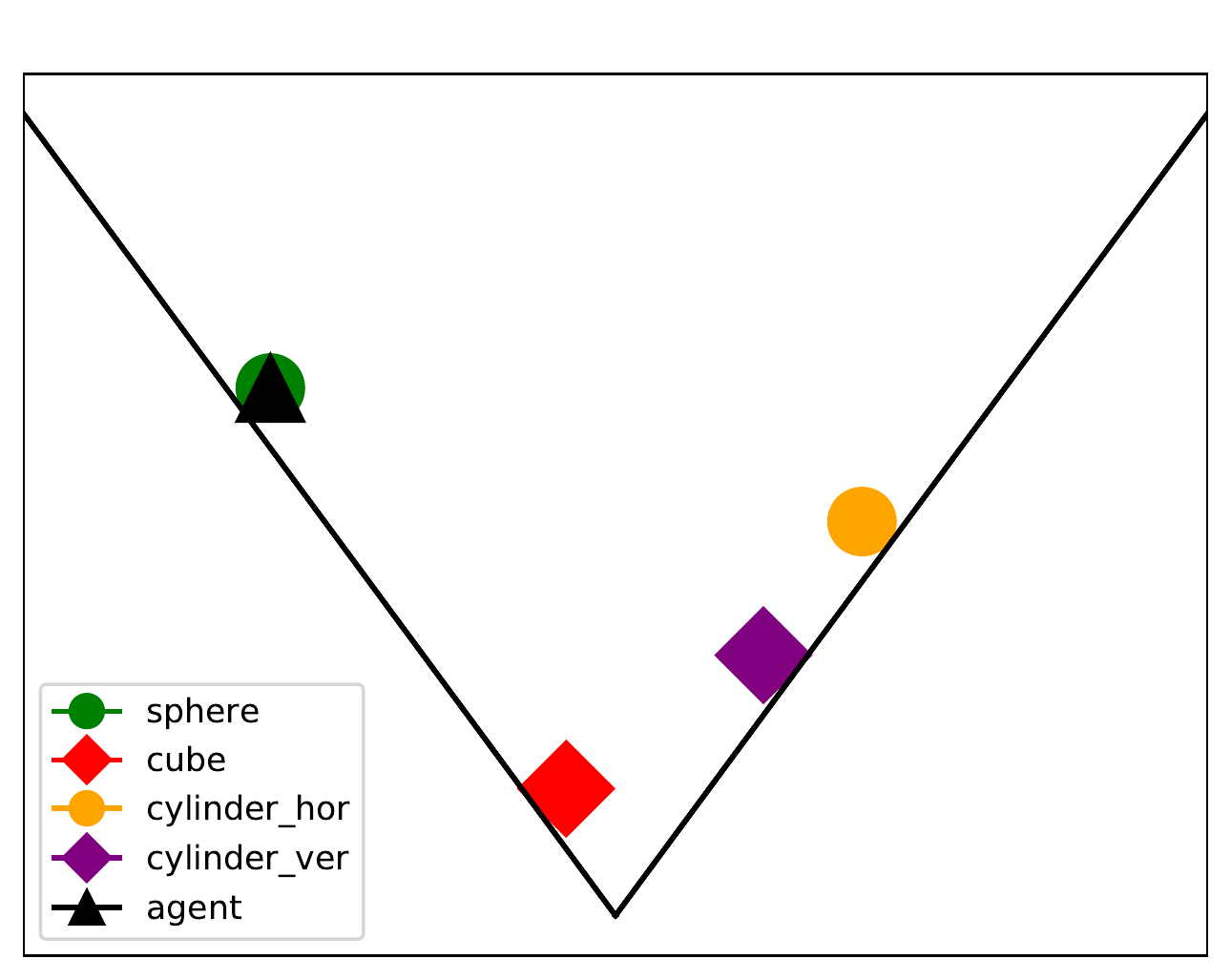}
     } &
\subfloat[Truth $t=6$]{%
       \includegraphics[width=0.225\linewidth]{ 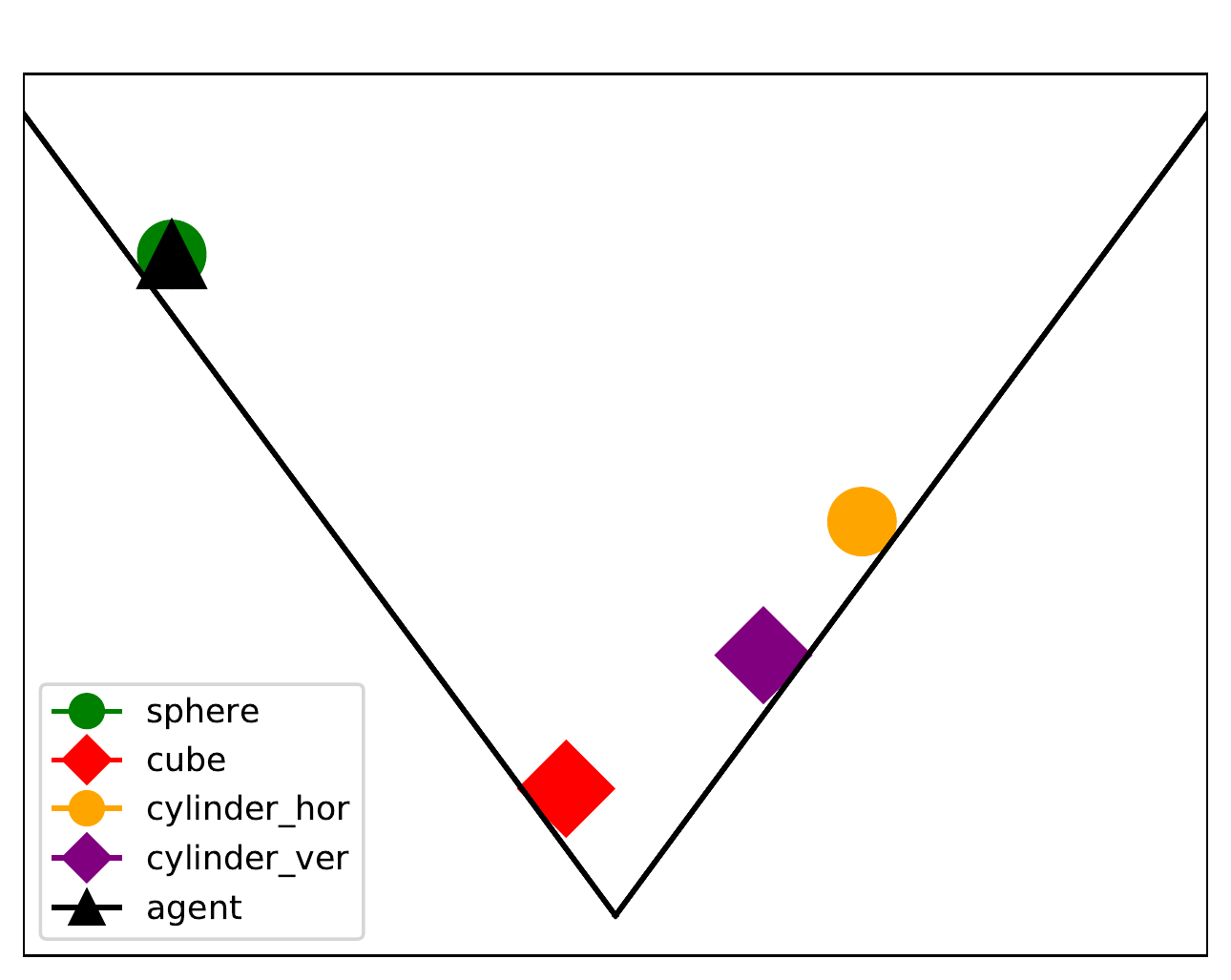}
     } &
\subfloat[Truth $t=7$]{%
       \includegraphics[width=0.225\linewidth]{ 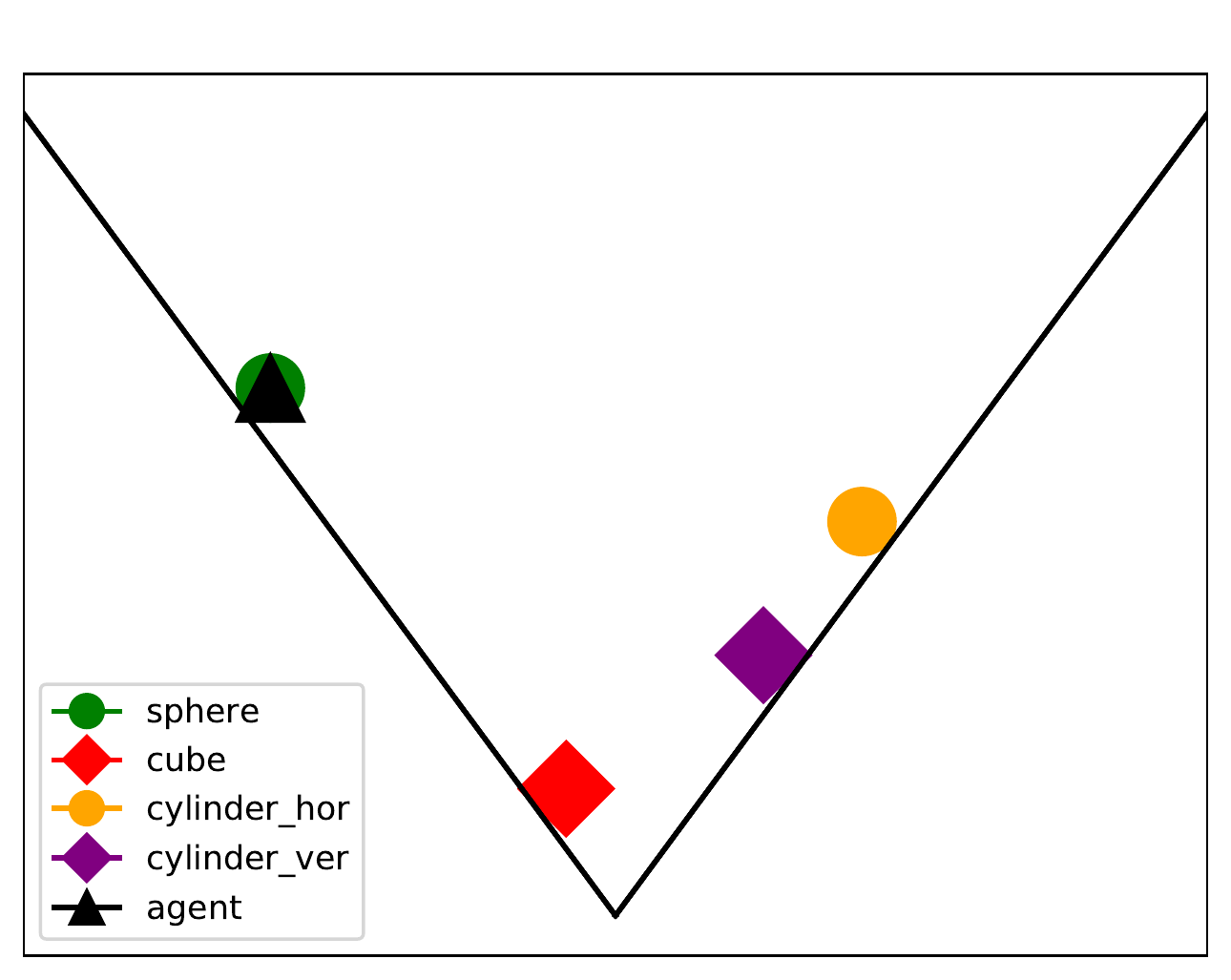}
     } &
\subfloat[Truth $t=8$]{%
       \includegraphics[width=0.225\linewidth]{ 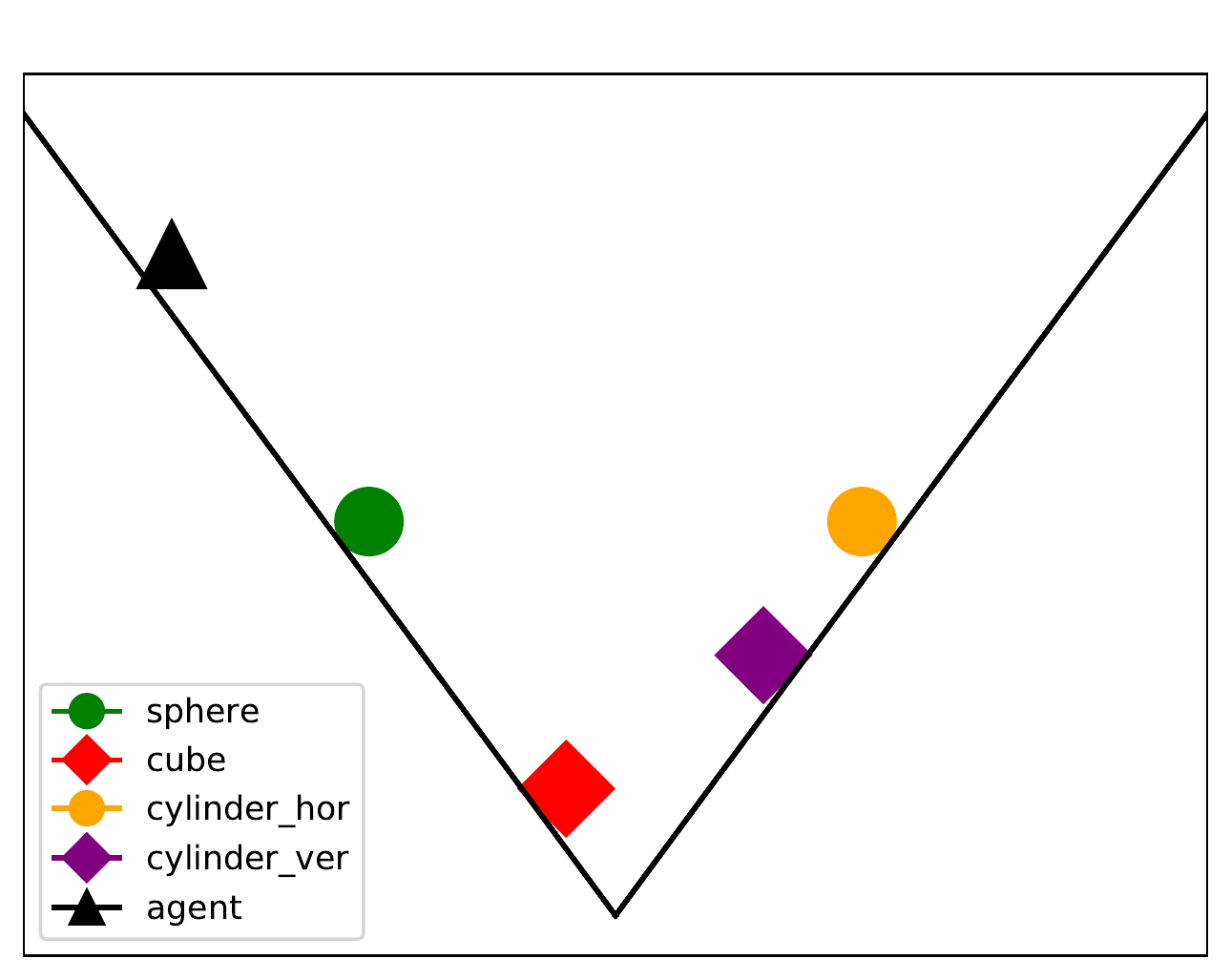}
     } \\
\end{tabular}
\caption{Visual inspection of the models predictions on the test set for \emph{Inclined Plane with Agent}. From this frame sequence, it can be noticed that the learned dynamics model takes into account correctly the effects of both the slope and the agent on the green ball. For example, from $t=5$ to $t=7$, the green ball follows the movements that the agent (the black triangles)  takes while at $t=8$, the agent takes the action \emph{Move left without grabbing}. Consequently, the green ball is at that point subject only to the gravity due to the right slope, so it moves one step downhill. Our model captures also this movement due to the environment properties in addition to the movements due to the interaction with the agent.}
\label{fig:rollouts_valley_action_ball_manipulation}
\end{figure}

\begin{figure}[h!]
    \centering
\begin{tabular}{cccc}
       \subfloat[NID $t=5$]{%
       \includegraphics[width=0.225\linewidth]{ 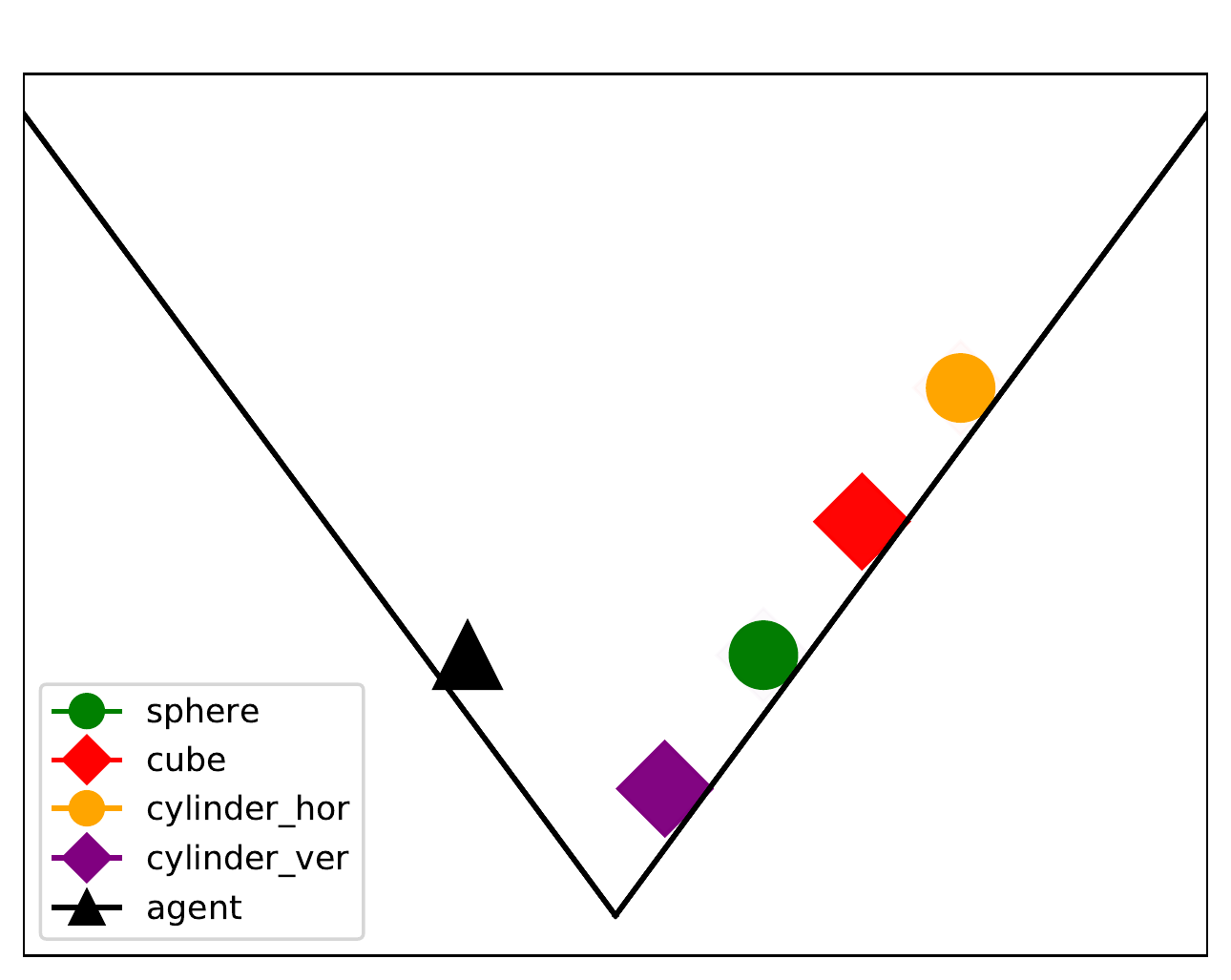}
       } &
       \subfloat[NID $t=6$]{%
       \includegraphics[width=0.225\linewidth]{ 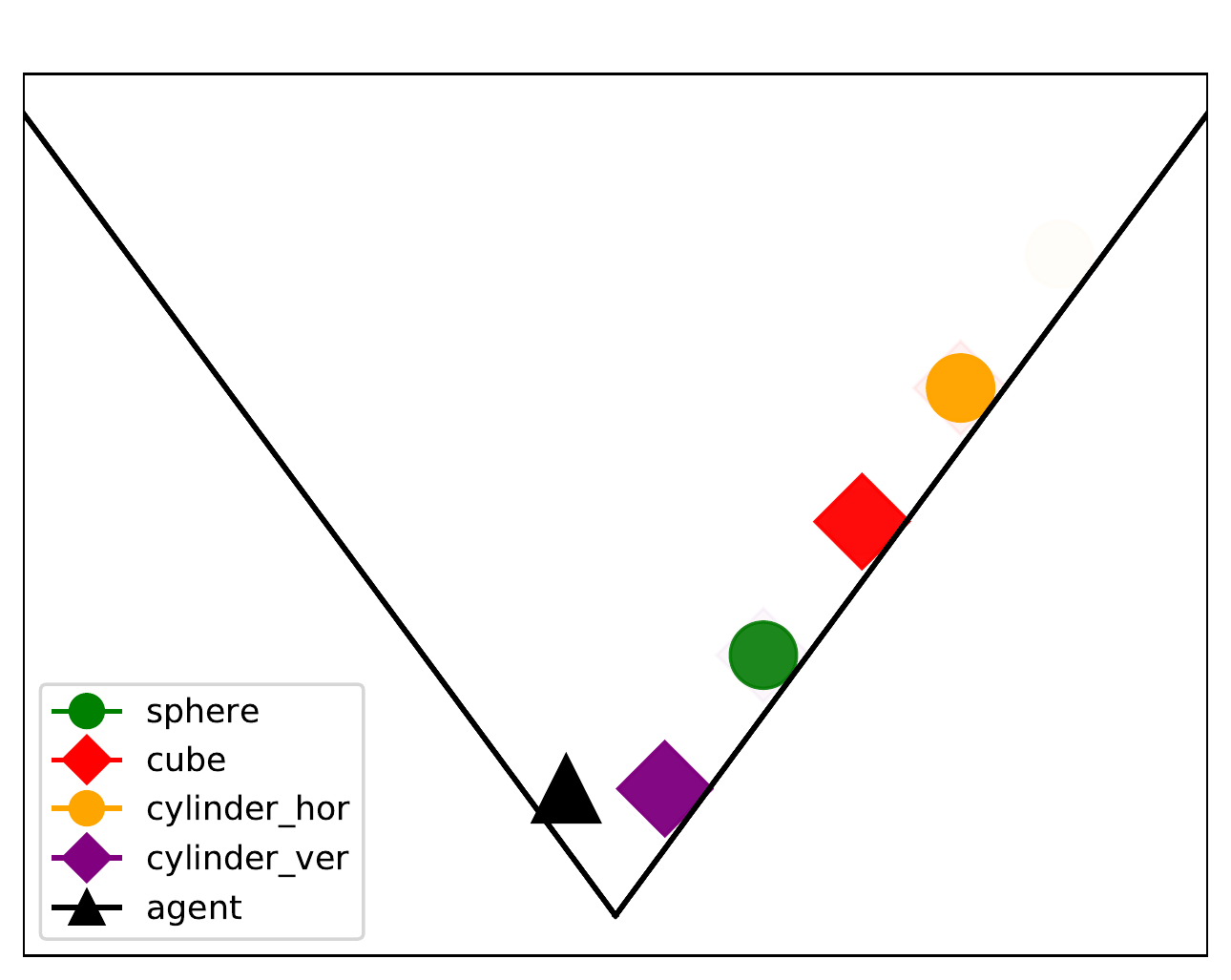}
       } &
       \subfloat[NID $t=7$]{%
       \includegraphics[width=0.225\linewidth]{ 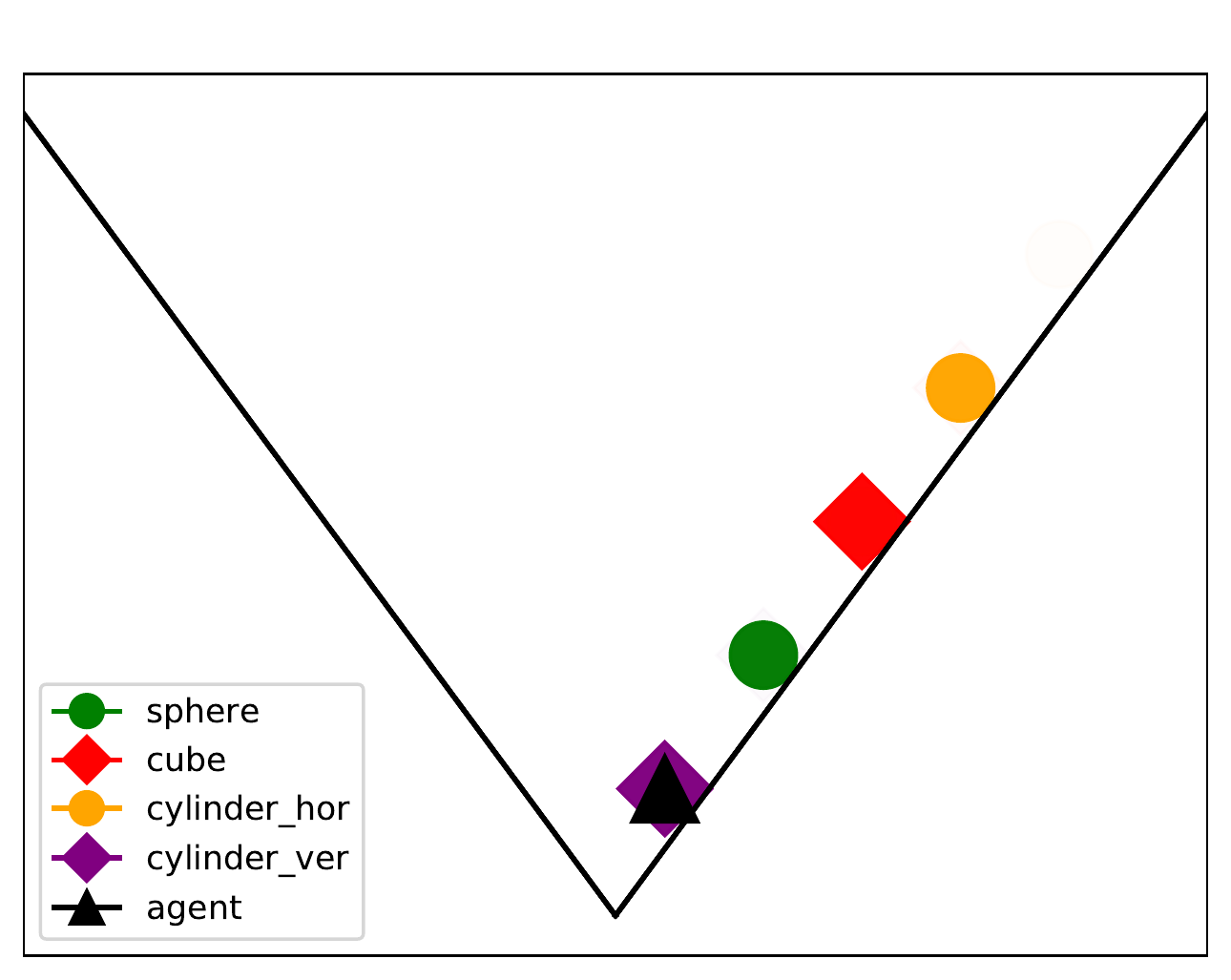}
       } &
       \subfloat[NID $t=8$]{%
       \includegraphics[width=0.225\linewidth]{ 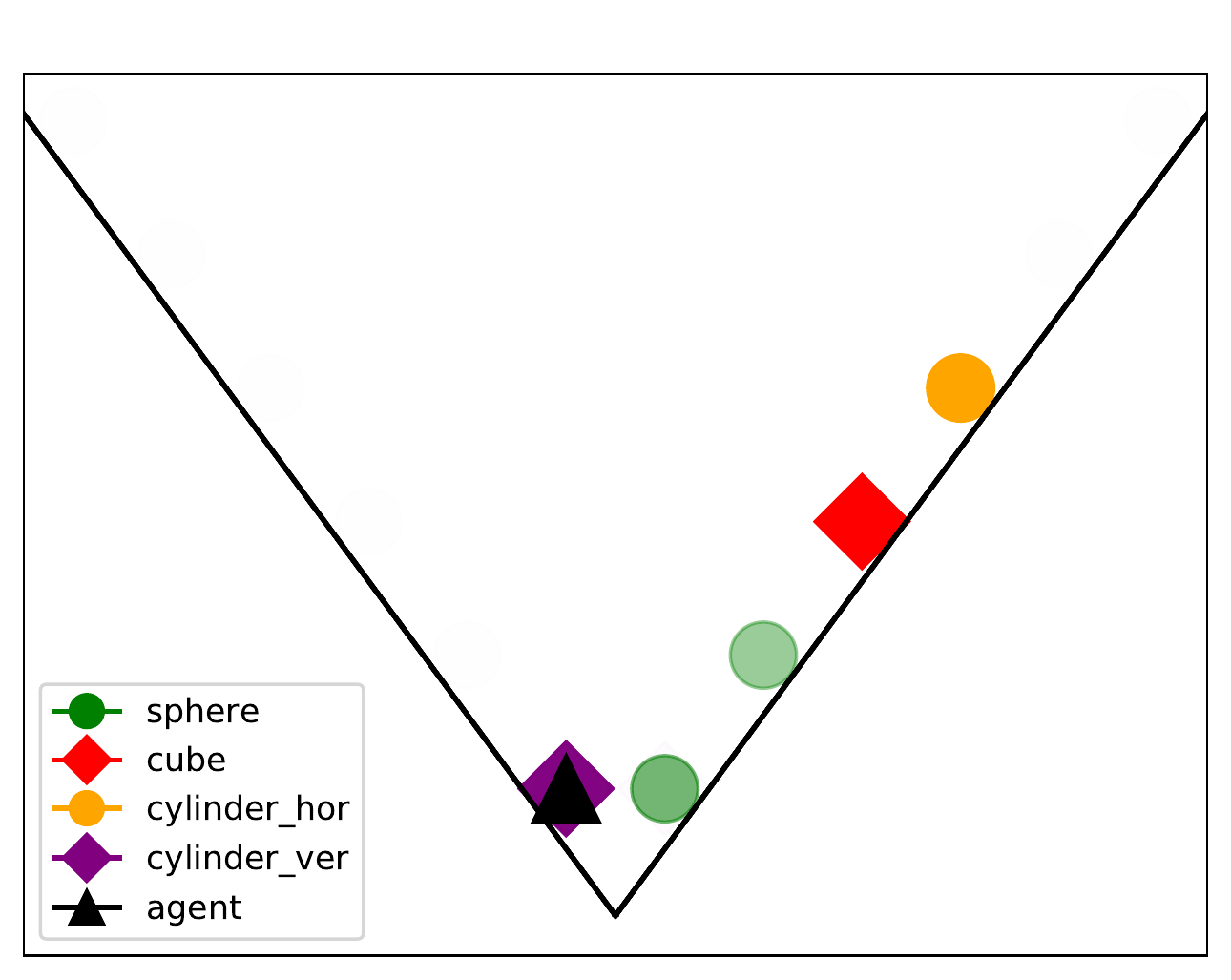}
       } \\
\subfloat[Truth $t=5$]{%
       \includegraphics[width=0.225\linewidth]{ 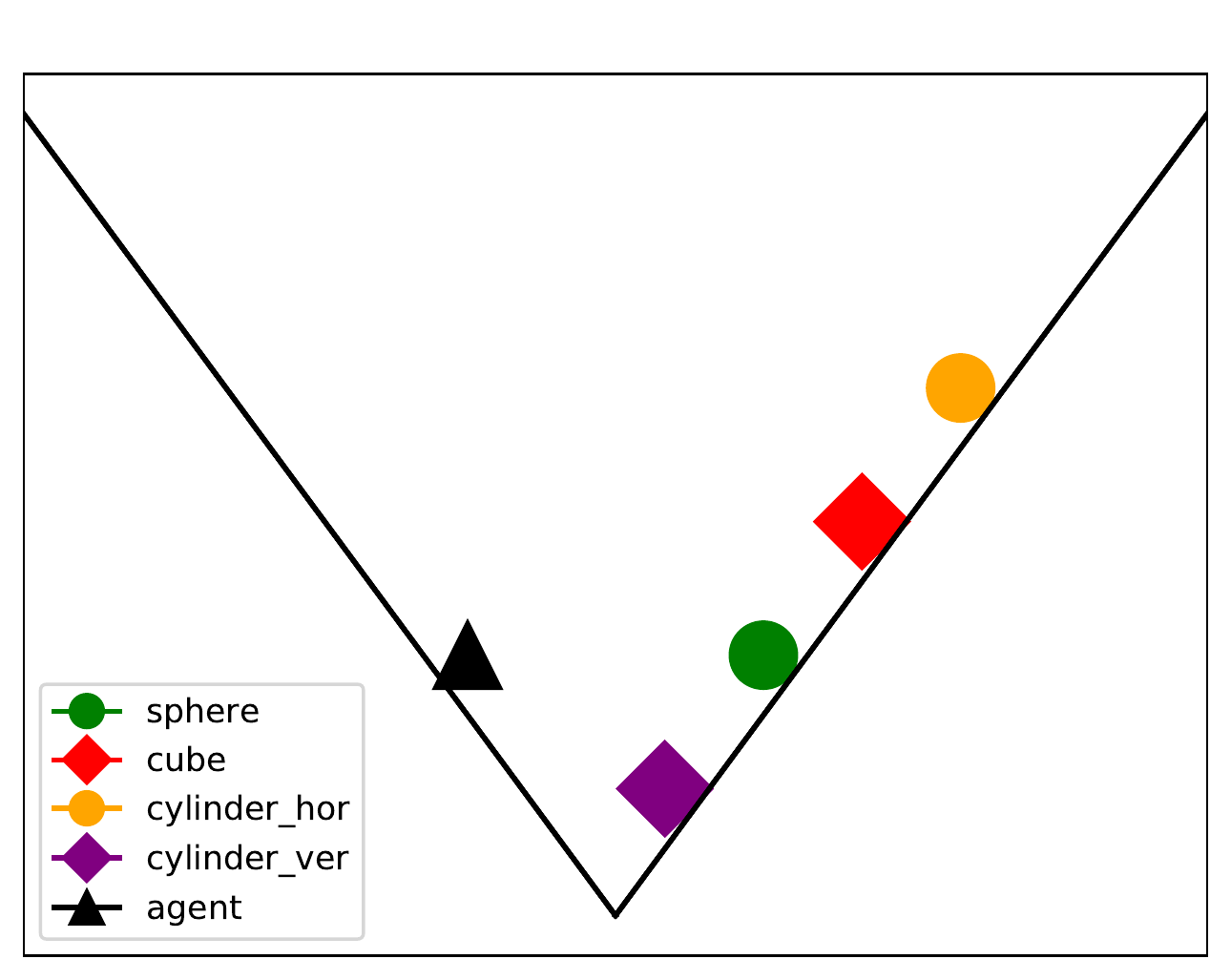}
     } &
\subfloat[Truth $t=6$]{%
       \includegraphics[width=0.225\linewidth]{ 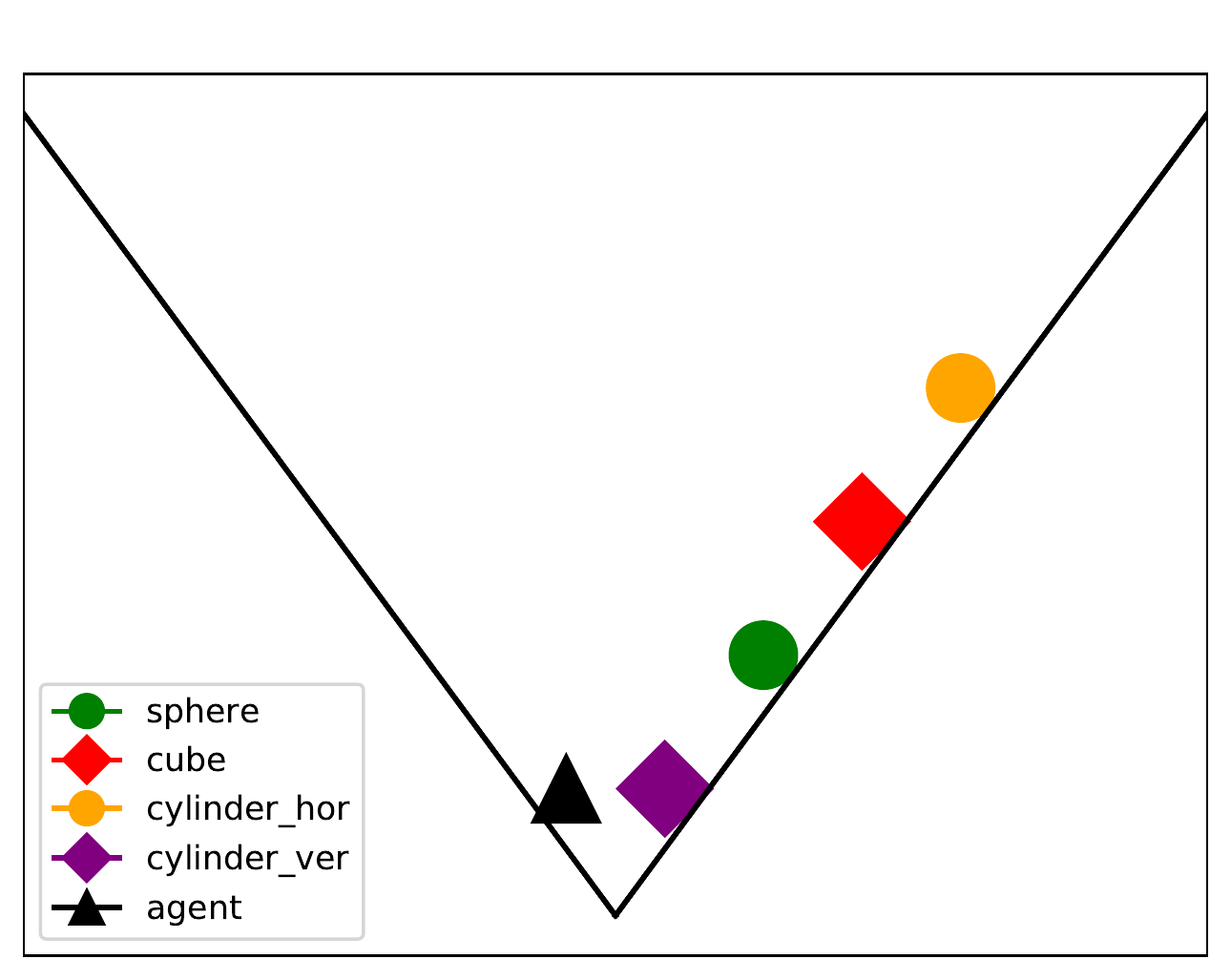}
     } &
\subfloat[Truth $t=7$]{%
       \includegraphics[width=0.225\linewidth]{ 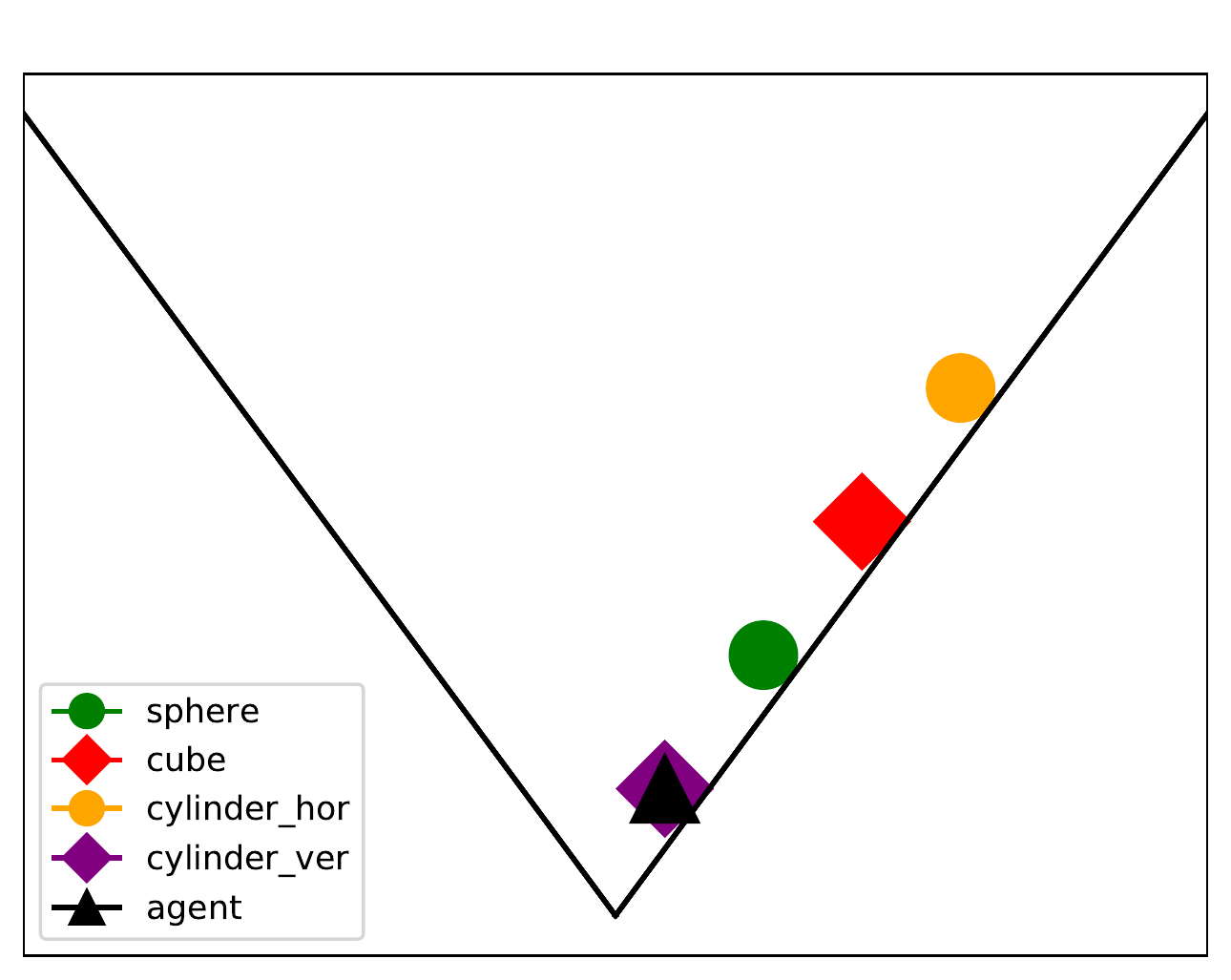}
     } &
\subfloat[Truth $t=8$]{%
       \includegraphics[width=0.225\linewidth]{ 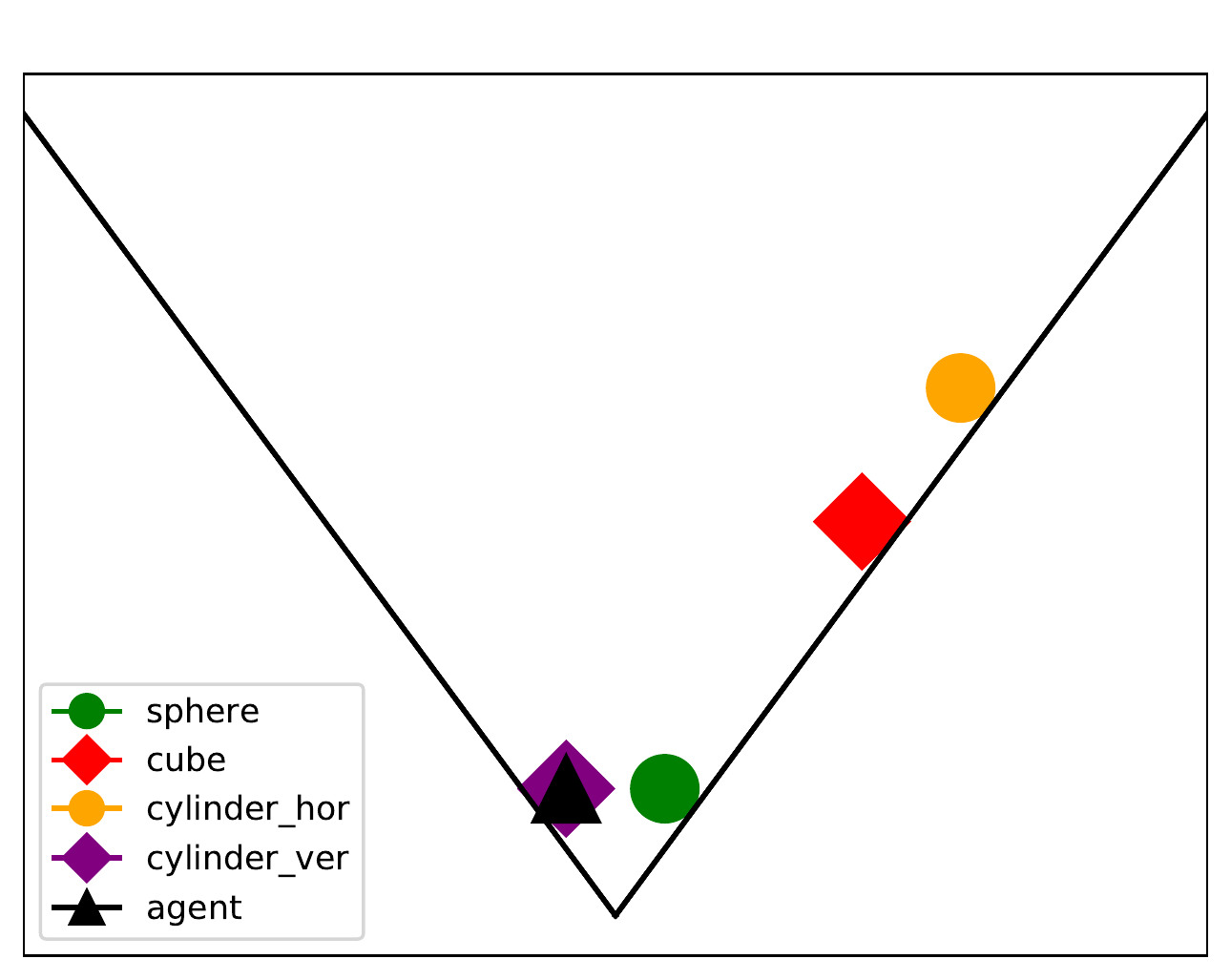}
     } \\
\end{tabular}
\caption{Another visual inspection of the models predictions on the test set for \emph{Inclined Plane with Agent}. From $t=5$ to $t=7$, we see that Neural NID captures well the decision of the agent to move one step on the right. Then at $t=7$, the agent is in the same position of the purple cube and it takes the action \emph{Move left while grabbing}. This action changes the state of the purple cube that can no longer sustain the green ball. Consequently, it rolls down from the right slope. The correct modeling of such kind of "chain reactions" requires to understand the effects that the agent's action on one object can have on the others. In this difficult situation our Neural NID struggles to provide a perfect prediction, however the subplot (d) shows that Neural NID puts the largest probability mass on the correct position of the green ball.}
\label{fig:rollouts_valley_action_cube_manipulation}
\end{figure}

\begin{figure}[t]
    \centering
\begin{tabular}{cccc}
       \subfloat[NID $t=0$]{%
       \includegraphics[width=0.225\linewidth]{ 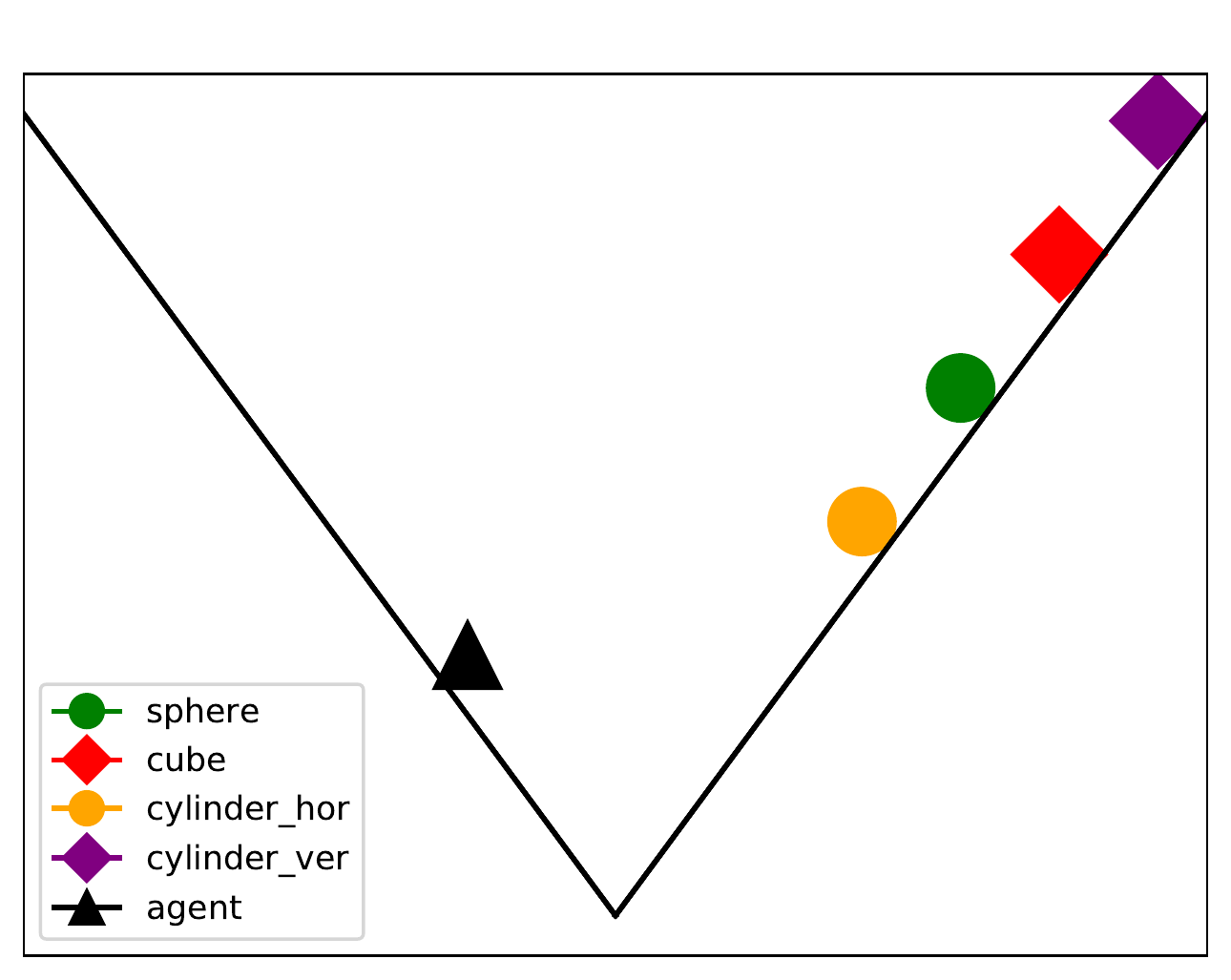}
       } &
       \subfloat[NID $t=1$]{%
       \includegraphics[width=0.225\linewidth]{ 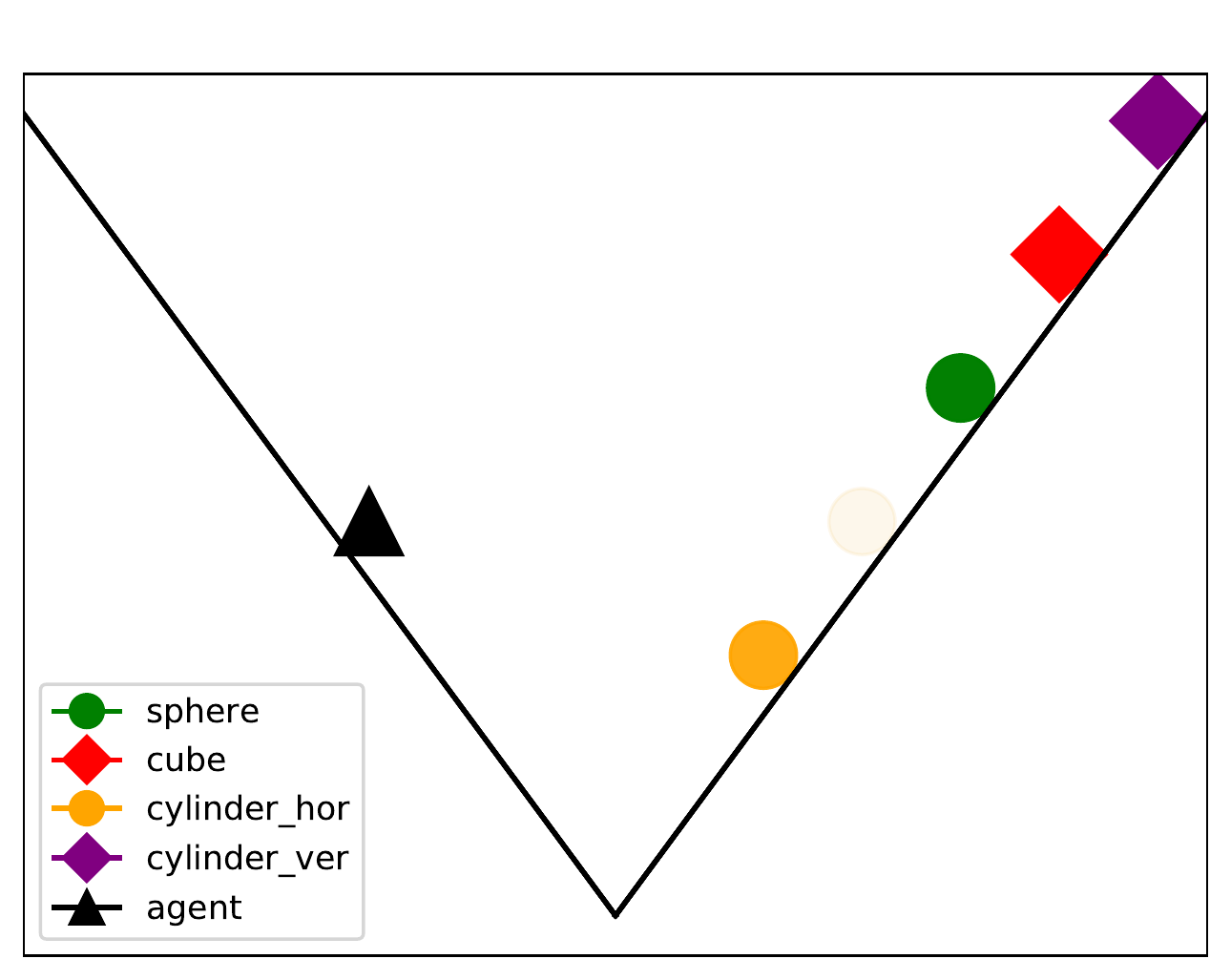}
       } &
       \subfloat[NID $t=2$]{%
       \includegraphics[width=0.225\linewidth]{ 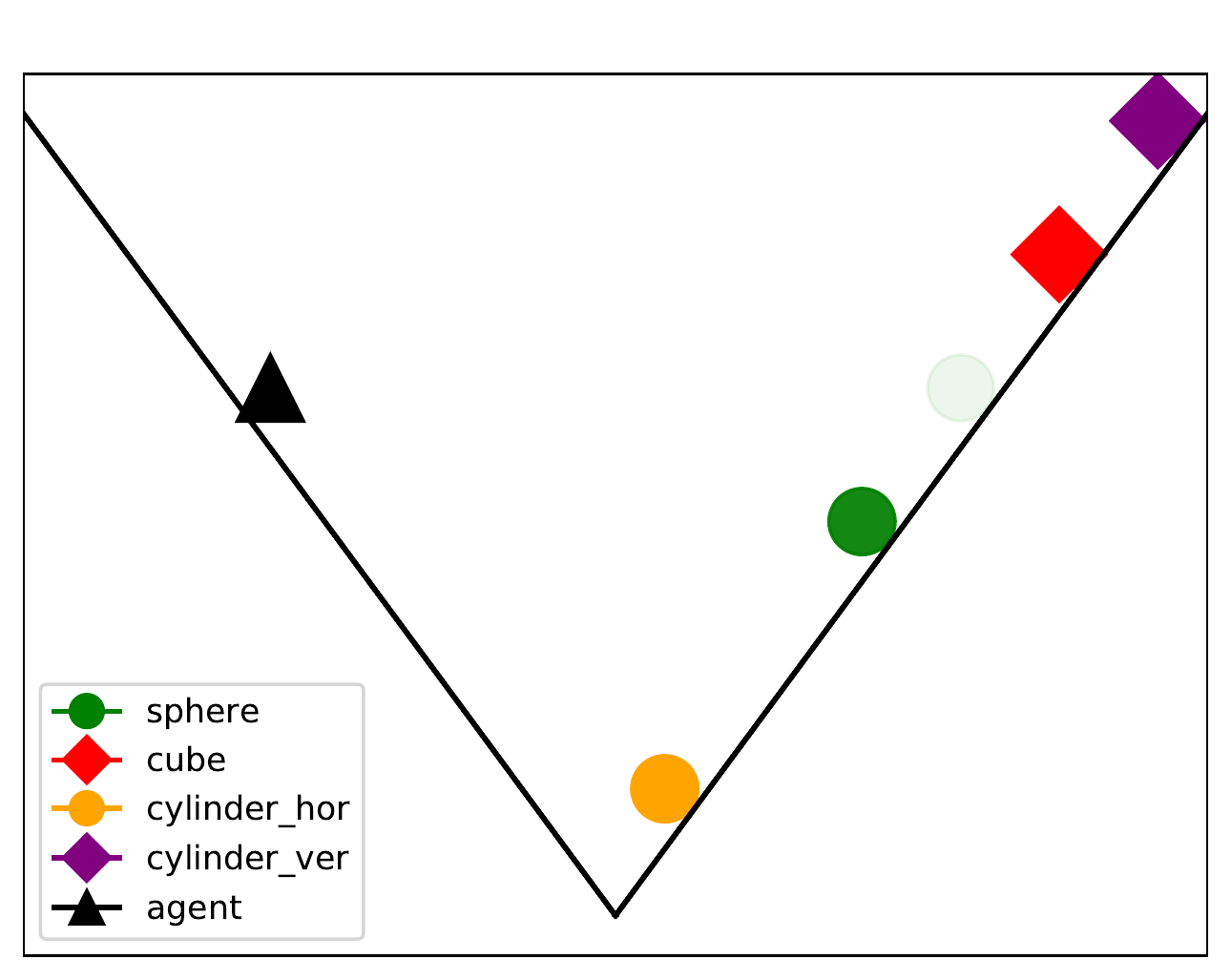}
       } &
       \subfloat[NID $t=3$]{%
       \includegraphics[width=0.225\linewidth]{ 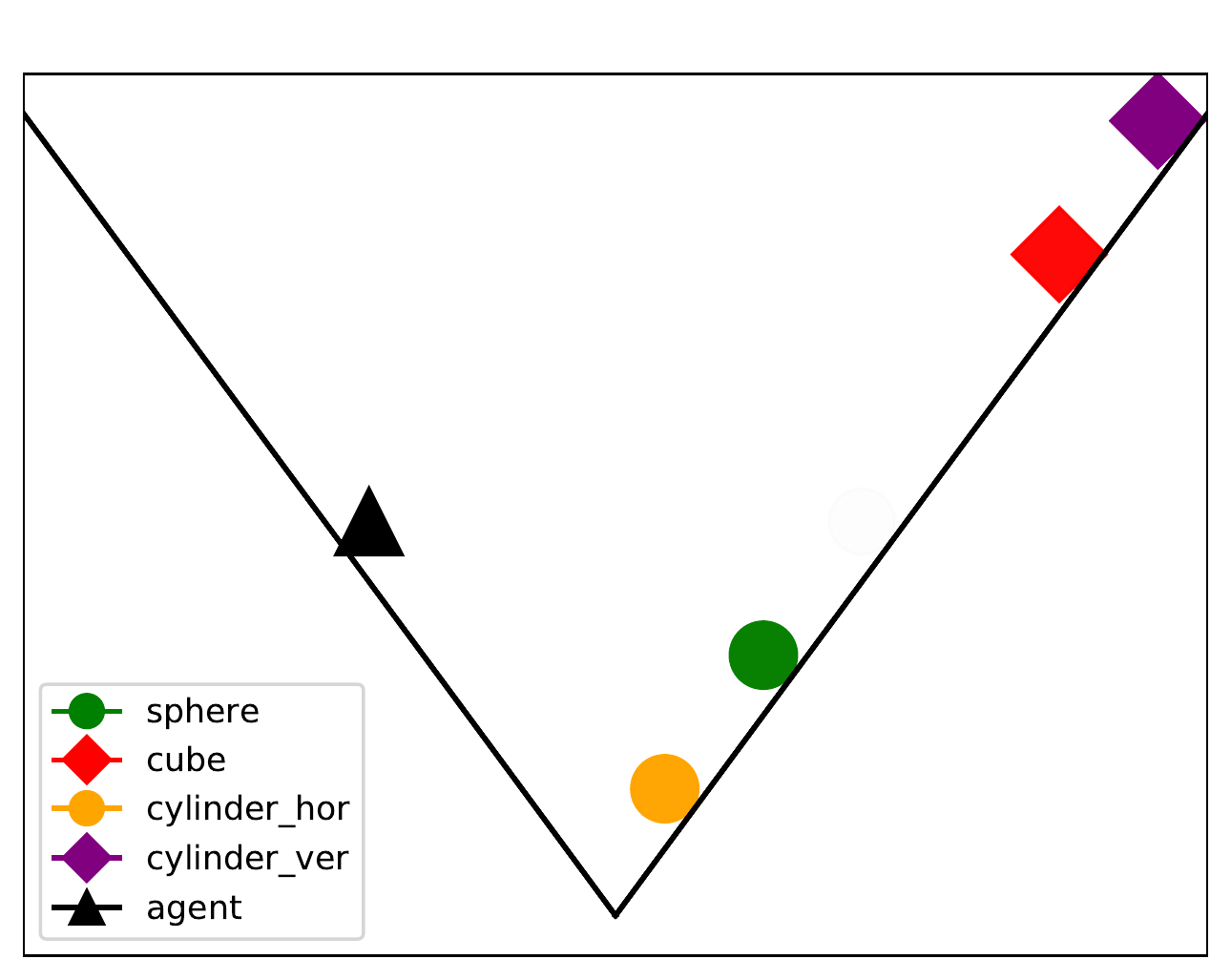}
       } \\
\subfloat[Truth $t=0$]{%
       \includegraphics[width=0.225\linewidth]{ frames/ValleyActions/rollout_rolling_down_targets/start.pdf}
     } &
\subfloat[Truth $t=1$]{%
       \includegraphics[width=0.225\linewidth]{ 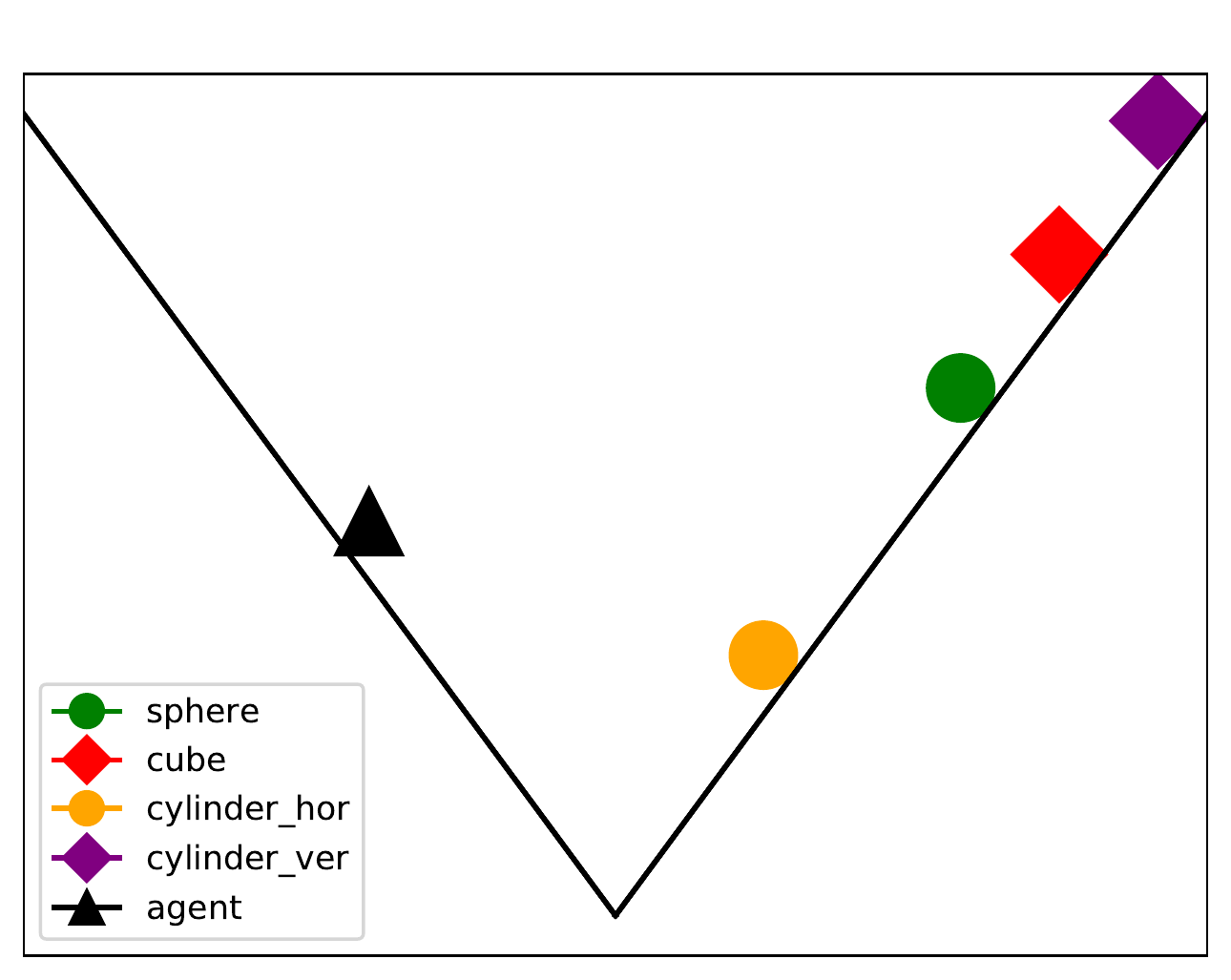}
     } &
\subfloat[Truth $t=2$]{%
       \includegraphics[width=0.225\linewidth]{ 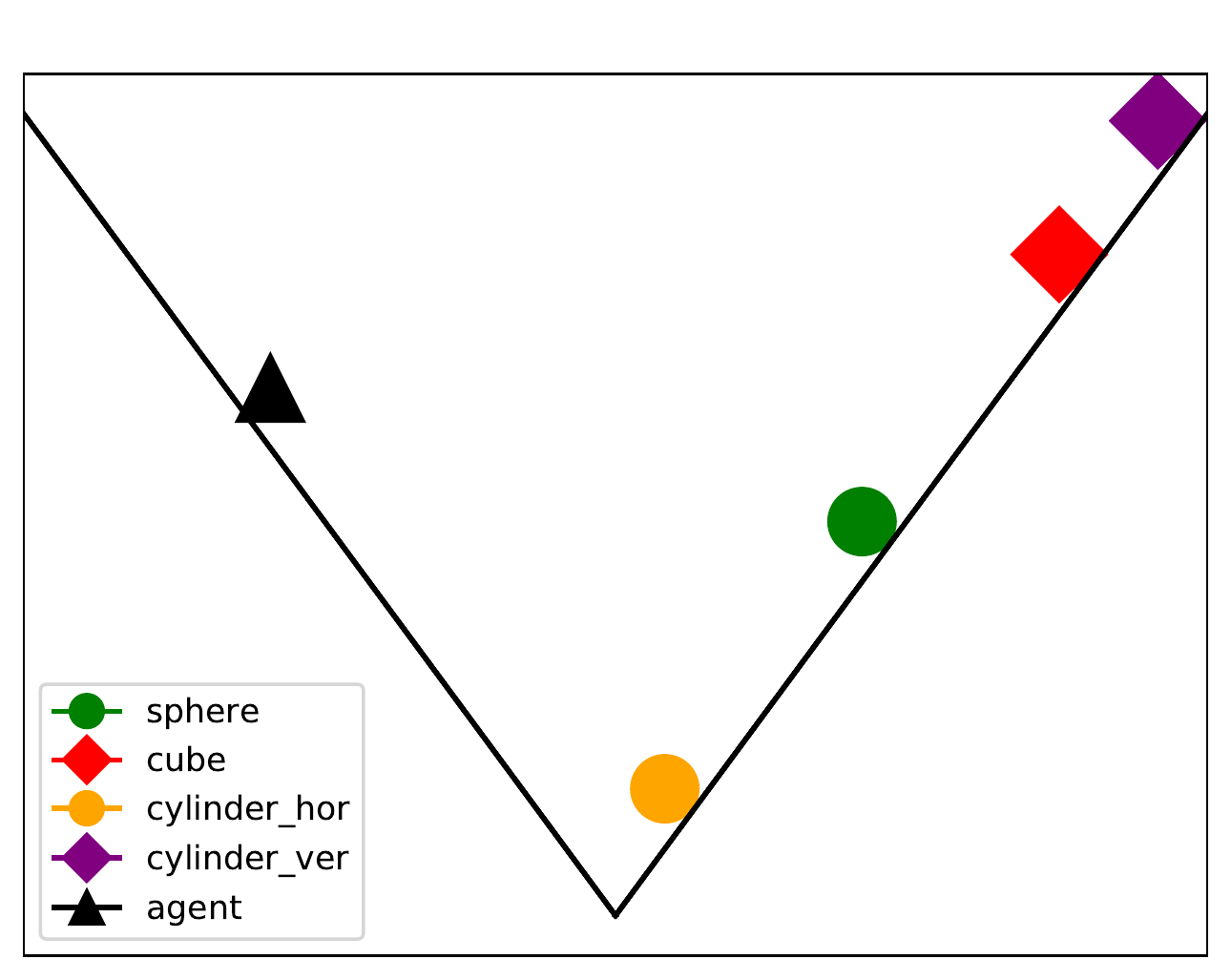}
     } &
\subfloat[Truth $t=3$]{%
       \includegraphics[width=0.225\linewidth]{ 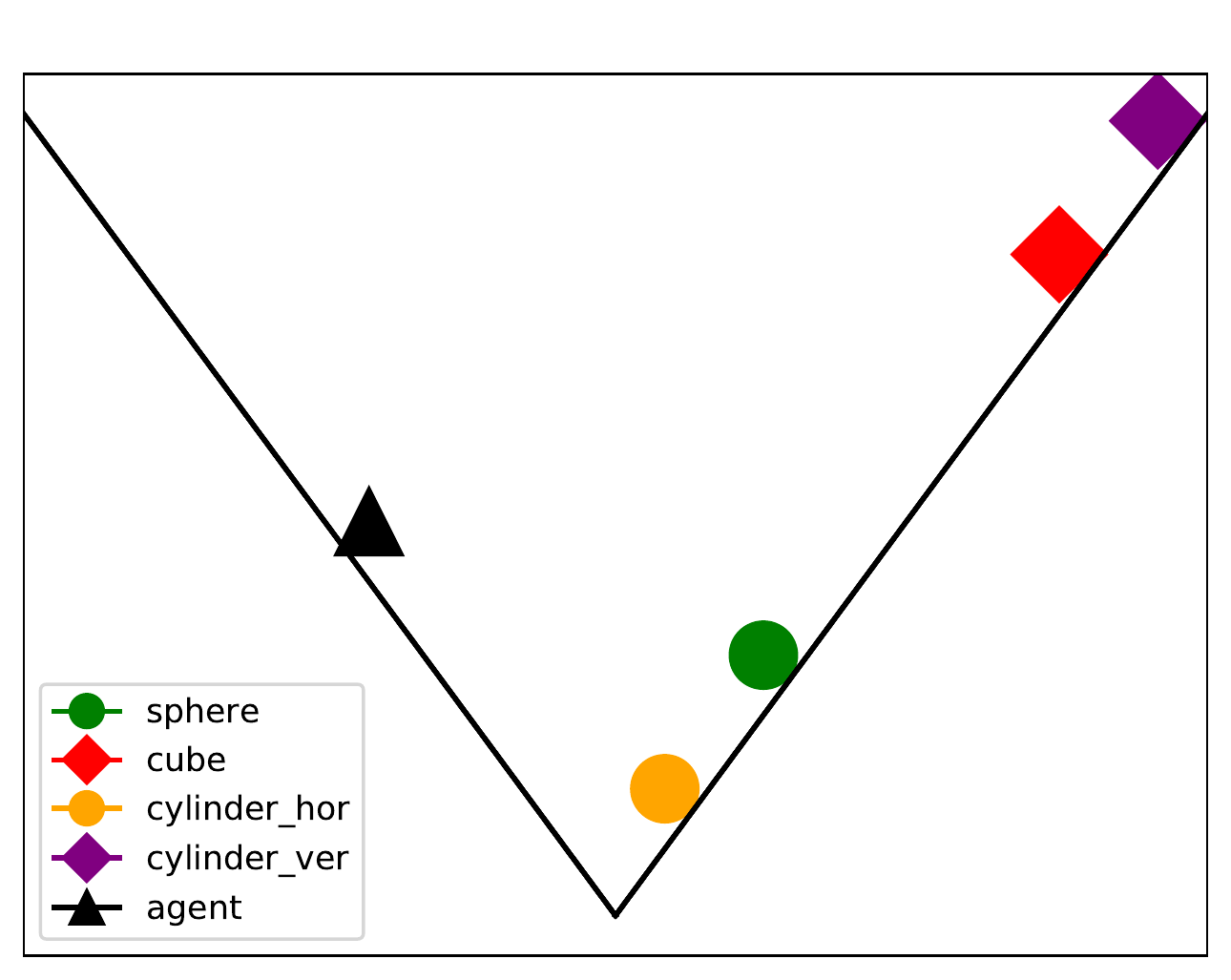}
     } \\
\end{tabular}
\caption{Visual inspection of the out-of-training-distribution generalization in the environment \emph{Inclined Plane with Agent}. Looking at the right slope we notice that the Neural NID predictions for the yellow and green balls match the true evolution of the system. Recall that the yellow ball was never seen on that slope during training so in order to correctly predict its behaviour the model needs to reason about the abstract properties of that object.}
\label{fig:rollouts_valley_action_rolling_down}
\end{figure}

Rollouts of Neural NID predictions are compared against the ground truth in Figures \ref{fig:rollouts_valley_action_ball_manipulation}, \ref{fig:rollouts_valley_action_cube_manipulation}, \ref{fig:rollouts_valley_action_rolling_down}.

\begin{table}[t]
\centering
\begin{tabular}{l|l}
\textbf{Hyperparameter}                                   & \textbf{Value}                      \\ \hline
\hline
$d^R$                                & 4 \\
$K$                                 & 4\\
$d^1$                                 & 2\\
$d^P$                                         & 4                      \\
Relational Convolutional Filter Sizes                                    & $S_1 = 1$                   \\
Outcome Convolutional Filter Sizes                                    & $S_2 = 1$                   \\
Optimizer, lr, batch size & RMSProp, 1e{-}2, 1 (Online Training)      \\
Entropy regularizers                & $\lambda_1 = 5e{-}7, \lambda_2 = 5e{-}6$  \\
$\mathrm{MLP}$ Layers                      & 2 \\
$\mathrm{MLP}$ Activation                      & $\mathrm{tanh}$ \\
\hline
\end{tabular}
\caption{Hyperparameters for Neural NID common to all the experiments 
\label{tab:hyp}}
\end{table}

\section{Ablation study}
We carried out an ablation study to verify which of the Neural NID building blocks are the more critical in achieving a strong out-of-distribution generalization.
The following paragraph investigates ablation regarding the following components:
\begin{enumerate}
    \item The importance of entropy regularization in the loss function (\emph{cf.} \eqref{eq:loss})
    \item The importance of the matrix $\boldsymbol{W}$ rank upper bound $K$.
    \item The attention mechanism. In particular, we compare a sample dependent to a sample independent version of the attention mechanism for the function $f^{\mathrm{enc}}$.
\end{enumerate}
At first, we present the result obtained changing only one hyparameters at time while leaving all the others as in Table \ref{tab:hyp}. After that, we present a wider ablation study.
\begin{figure}[h]
\centering
\begin{tabular}{cc}
    \subfloat[\textbf{Ablation:} Compound error for the test case. The higher accumulated error without entropy regularization shows its importance while the rank upper bound seems to be a less sensitive parameter. The green line overlaps with the blue one.
\label{fig:ablation_learning_curves}]{\includegraphics[width=0.5\textwidth]{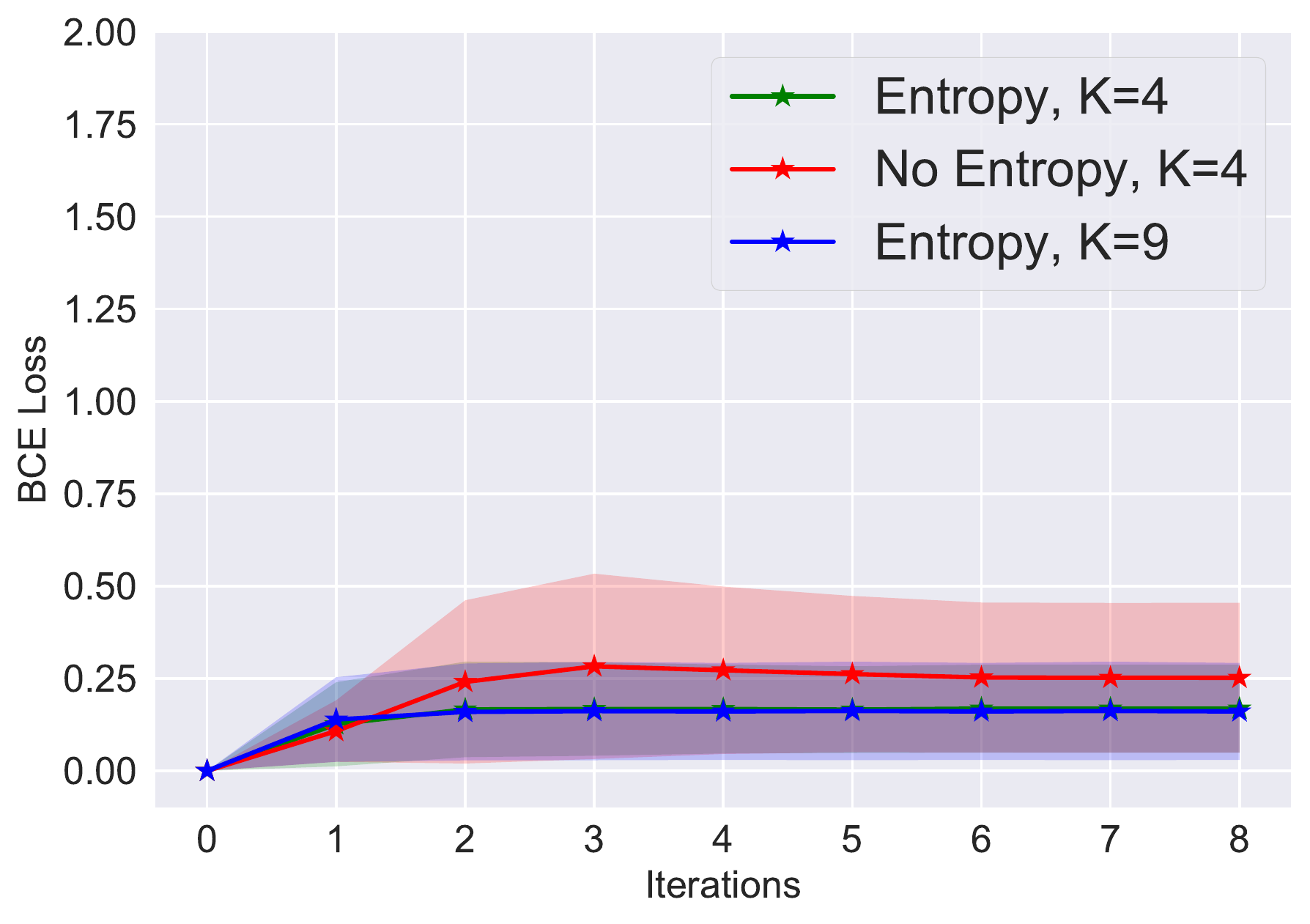}}
     &
     \subfloat[\textbf{Entropy Ablation:} Compound error for Neural NID comparing different entropy regularizers schemes. Both seems to be needed to keep the out-of-distribution error under control.\label{fig:entropyn_ablation_curves}]{\includegraphics[width=0.5\textwidth]{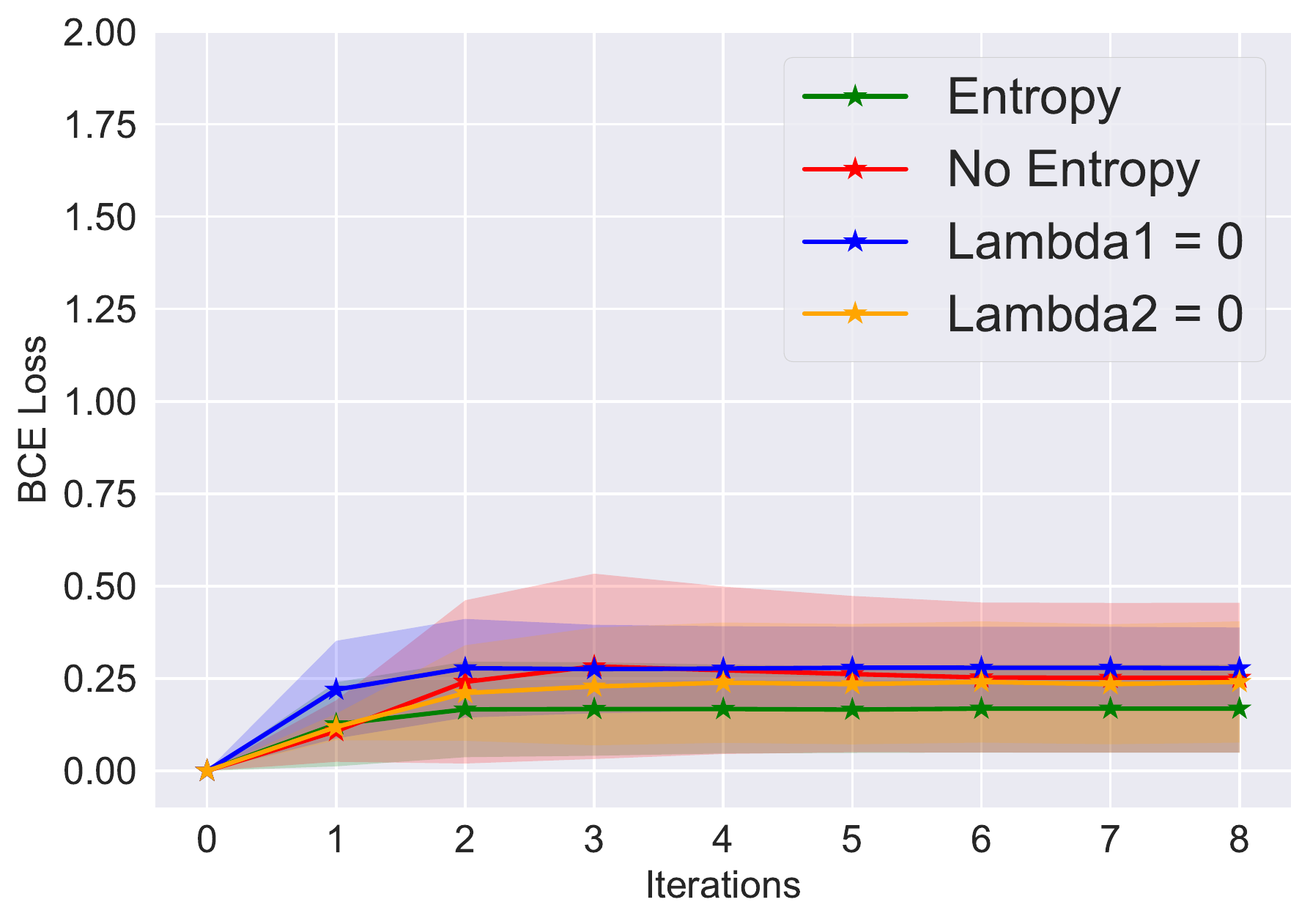}}\\
     \subfloat[\textbf{Attention Ablation:} Compound error for Neural NID with sample dependent and sample independent attention mechanism.\label{fig:attention_ablation_learning_curves}]{\includegraphics[width=0.5\textwidth]{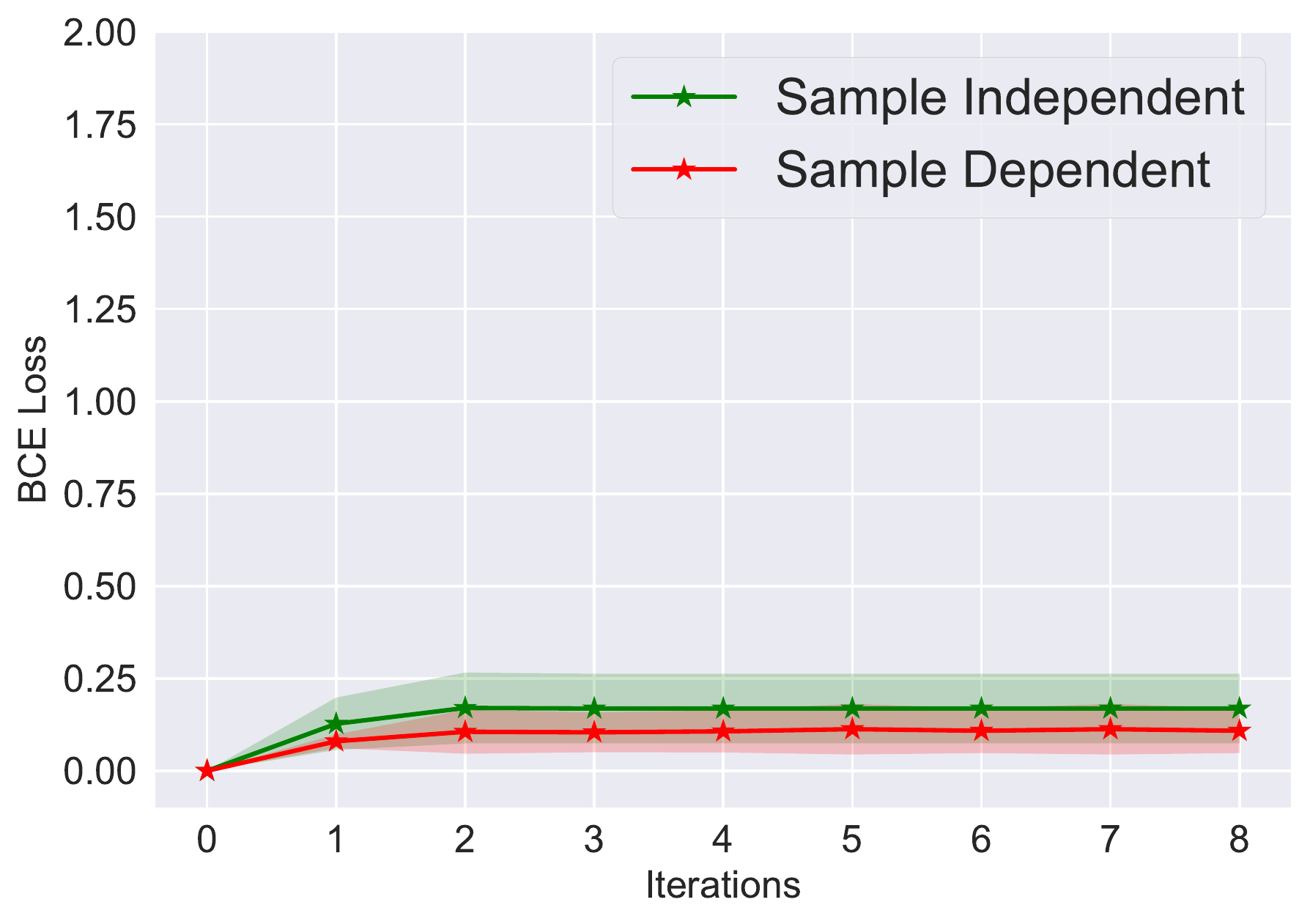}}
     &
     \end{tabular}
\end{figure}
\section{Entropy ablation}
The effect of entropy regularization in the loss \eqref{eq:loss} is clearly visible in the embedding space for the vector $\bar{\mathbf{v}}^P$, that we recall is obtained as follows:
\begin{equation}
    \bar{\mathbf{v}}^P =\sigma (\mathrm{Concat}(\mathbf{e}_o, \mathbf{e}_p) \boldsymbol{W})
\end{equation}
where $\sigma$ is the sigmoid function applied pointwise.
In Figure \ref{fig:ablation_no_entropy} we report the obtained $\bar{\mathbf{v}}^P$ for all the possible combinations of $(o,p) \in \{ \mathcal{O} \times \{ 1, \dots, D \}$. Figure \ref{fig:ablation_no_entropy} refers to the environment \emph{Inclined Plane} where $D=12$ and $\mathcal{O} = \{1,\dots, 5\}$. The coloring in Figure \ref{fig:ablation_no_entropy} are assigned according to the object index and the colours match the ones used in the environment renderings, e.g. Figure \ref{fig:rollouts_inclined_plane}.
Figure \ref{fig:ablation} is the result of the same experiment but using the standard entropy regularization with values for $\lambda_1$ and $\lambda_2$ given in Table \ref{tab:hyp}.

The visual comparison between Figures \ref{fig:ablation_no_entropy} and \ref{fig:ablation} reveals that without entropy regularization the values attained by $\bar{\mathbf{v}}^P$ are highly scattered. On the contrary, those are clustered when we introduce entropy in the loss function.
This fact has consequences on the out-of-training distribution generalization.

Indeed, in Figure \ref{fig:ablation_learning_curves} we show that the error for the case without entropy compounds quicker along rollouts. One can notice also that without entropy regularization the variance across different training initializations is higher. 
\newline
Both those factor can be explained by the absence of entropy in the loss. Indeed, as the former point, the clustered representation obtained in that case makes object with equal properties indistinguishable for the following layers, thus those are forced to find a common representation for the behaviour of both objects. 
\newline
Regarding the latter point, without taking entropy into account, there are no longer embedding space configurations that are favourable with respect to the others. Therefore, the higher variance is due to the fact that some training initializations converge to configurations that are more convenient for out-of-training distribution generalization while others do not.

Beyond that, we verified that both the regularizers, i.e. $\lambda_1 \neq 0$ and $\lambda_2 \neq 0$, are necessary to encourage convenient representations in the embedding space. In particular, Figures \ref{fig:ablation_lambda1_0} and \ref{fig:ablation_lambda2_0} show that in the case of only $\lambda_1$ nonzero a clustered representation is hardly recovered while it emerges when $\lambda_2$ alone is used. All the others hyperparameters are fixed as in Table \ref{tab:hyp}.
\newline
Despite this finding, when looking at the out-of-distribution error in Figure \ref{fig:entropyn_ablation_curves} one can notice that the best performance is attained when both the regularizers are used.

In the Appendix, we report a more fine grained study where we tested different values of the hyperparameters $\lambda_1$ and $\lambda_2$. This more detailed ablation suggests that a complex relation exists between the two regularizers making difficult to extract general conclusions about their reciprocal tuning.

\begin{figure}[h]
    \centering
\begin{tabular}{cc}
      \subfloat[$\lambda_1$ coloring \label{fig:lambda_1_silohuette}]{%
       \includegraphics[width=0.5\linewidth]{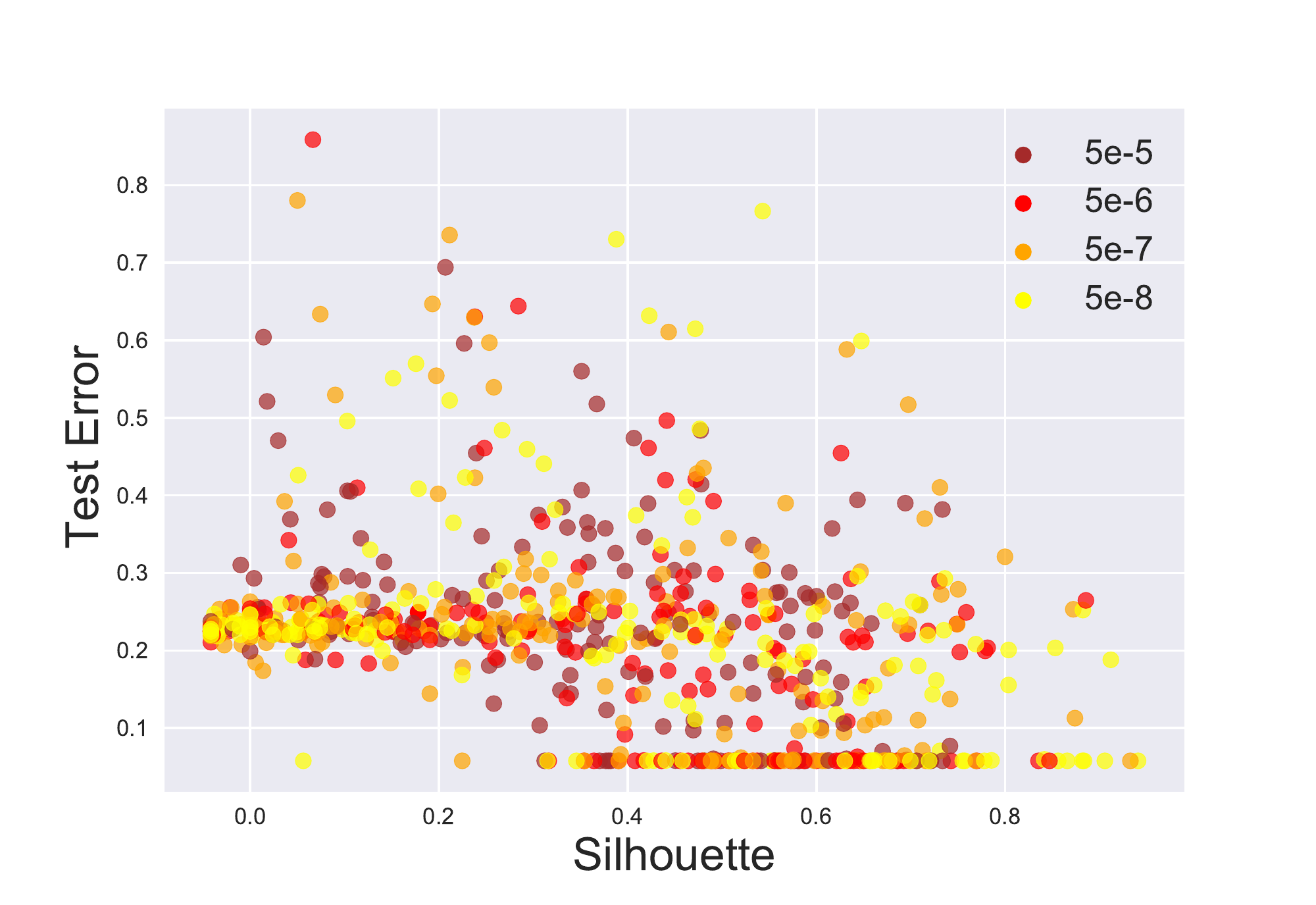}
       } &
       \subfloat[$\lambda_2$ coloring \label{fig:lambda_2_silohuette}]{%
       \includegraphics[width=0.5\linewidth]{Lambda2Ablation_both_rows.pdf}
       } \\
       \subfloat[$K$ coloring \label{fig:K_silohuette}]{%
       \includegraphics[width=0.5\linewidth]{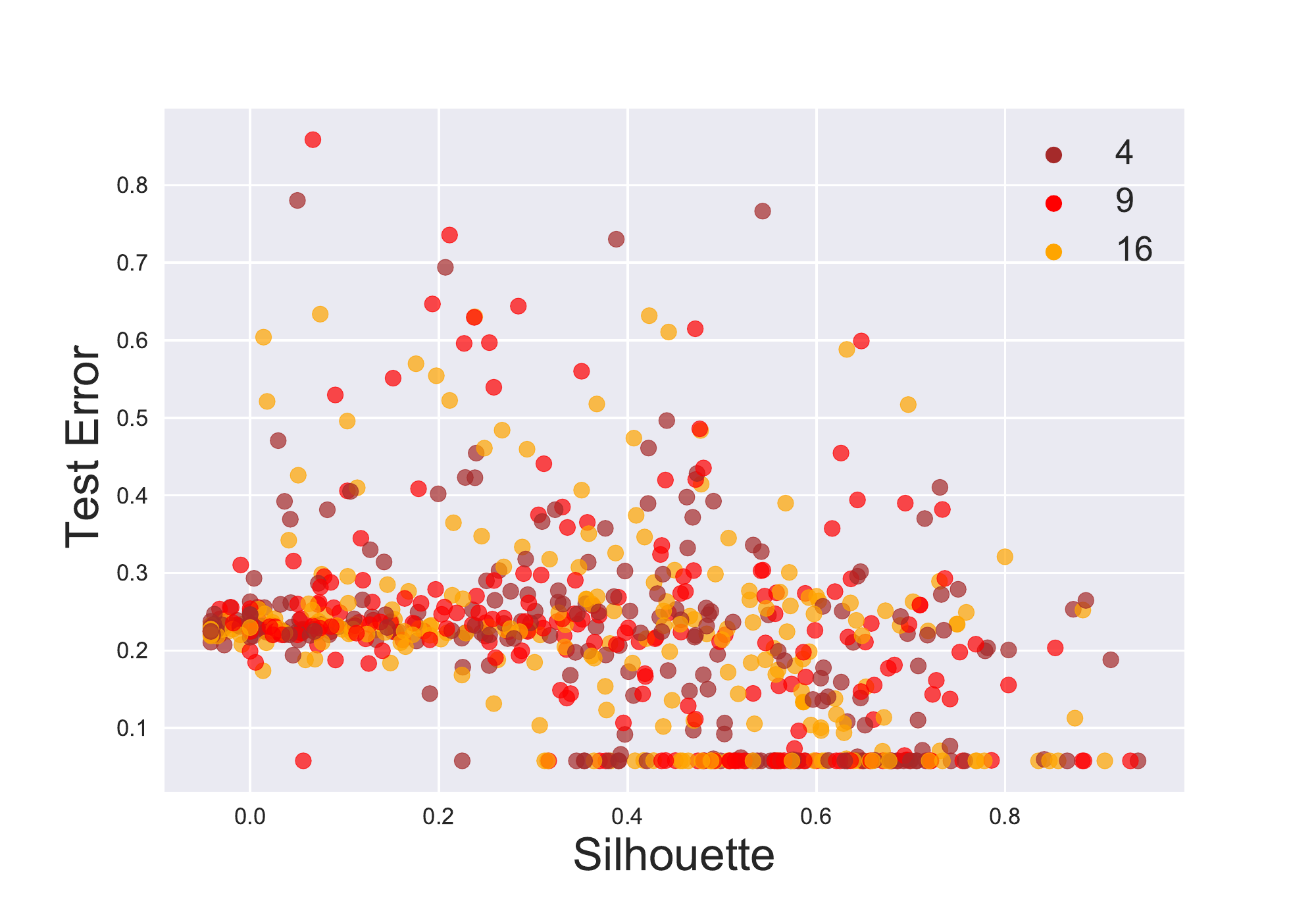}
       } &
       \subfloat[Initialization coloring \label{fig:rows_silohuette}.]{%
       \includegraphics[width=0.5\linewidth]{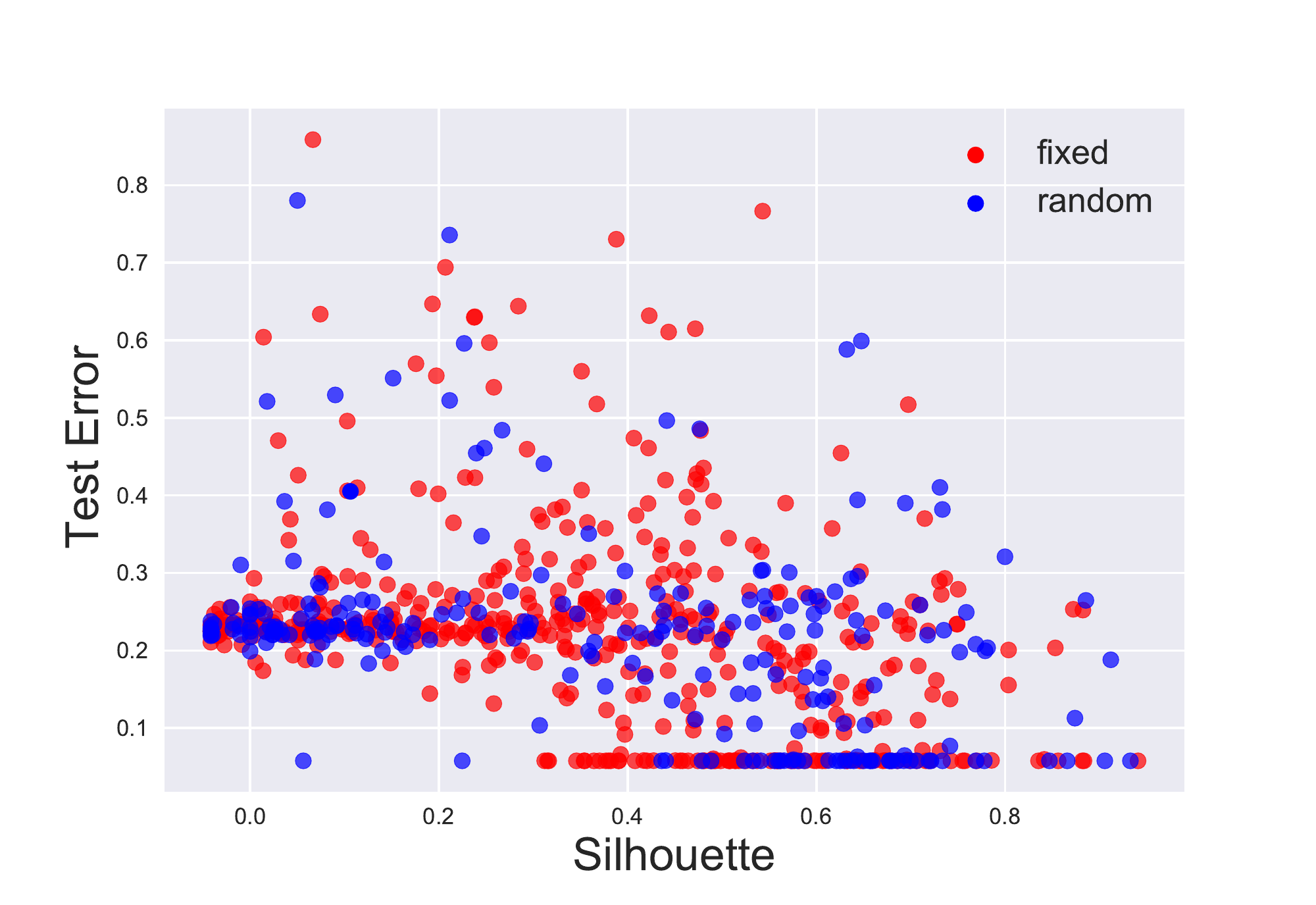}
       } \\
    \end{tabular}
\caption{Relation between Silhouette and out-of-distribution test error. We report the same point cloud colored according to the values of $\lambda_1$, $\lambda_2$, $K$ or the weight initialization scheme.
}
\label{fig:silohuette_vs_error}
\end{figure}
\section{Sample dependent versus sample independent attention}
\label{sec:sample_dep_vs_indep}
We tried to modify the standard attention mechanism as follows:
\begin{align}
    \mathbf{v}^P = & \sigma((\mathrm{Concat}(\mathbf{e}_o, \mathbf{e}_p))^T \boldsymbol{W}_1^P ) \boldsymbol{W}_2^P = \nonumber \\
    & \sigma((\mathrm{Concat}(\mathbf{e}_o, \mathbf{e}_p))^T \mathrm{softmax}(\boldsymbol{Q}) \boldsymbol{V}^P ) \boldsymbol{W}_2^P
    \label{eq:sample_independent}
\end{align}
In the above formulation, the attention matrix $\boldsymbol{W}_1^P$ is the same for each input $\mathrm{Concat}(\mathbf{e}_o, \mathbf{e}_p)$. We refer to this formulation as \emph{Sample independent attention}.
In the main text, we introduced instead an attention matrix that depends on the sample $\mathrm{Concat}(\mathbf{e}_o, \mathbf{e}_p)$. In formulas:
\begin{equation}
    \mathbf{v}^P = \sigma( \mathrm{softmax}(\mathrm{Concat}(\mathbf{e}_o, \mathbf{e}_p)^T\boldsymbol{Q}) \boldsymbol{V}^P ) \boldsymbol{W}_2^P
\end{equation}
Figure \ref{fig:attention_ablation_learning_curves} reports the cumulative error along a rollout in \emph{Inclined Plane} for the sample dependent or sample independent attention while keeping the others hyperparameters fixed as in Table \ref{tab:hyp}. For this choice, it emerges that the sample dependent version generalizes better and with lower variance than the sample independent version.

\section{Systematic ablation study}

We performed a grid search over all the possible configuration arising from $\lambda_1 \in \{ 5e{-}8, 5e{-}7, 5e{-}6, 5e{-}5 \}$,
$\lambda_2 \in \{ 5e{-}8, 5e{-}7, 5e{-}6, 5e{-}5 \}$,
random or fixed weight initialization,
$K \in \{4, 8, 16\}$ for fixed initialization and $K \in \{4, 9, 14\}$ for the random initialization. In addition, we used $10$ different seeds for the fixed initialization scheme and $5$ for the random one. That means that, in total,
$720$ Neural NID models have been compared.

In Figure \ref{fig:silohuette_vs_error}, we plot the cumulative error at the end of an episode on the test set versus the Silhouette score of the clustering scheme introduced in Section \ref{sec:evaluating_compactness}. The coloring scheme reflects the value of the ablation parameters ( $\lambda_1$, $\lambda_2$, $K$ and the initialization scheme).
We can notice that in the most favourable region for generalization ( low test error, high Silhouette) we fing models with constant values of $\lambda_1$ and $\lambda_2$, see Figures \ref{fig:lambda_1_silohuette} and \ref{fig:lambda_2_silohuette}. In particular, the choice $\lambda = 5e{-}8$, $\lambda_2 = 5e{-}6$ seems the best for \emph{Inclined Plane}. On the contrary, the value of $K$ and the initialization schemes have less impact. Notice for example the variety of colours in the region with low test error and high Silhouette in Figures \ref{fig:K_silohuette} and \ref{fig:rows_silohuette}.

\begin{figure}[t]
    \centering
\begin{tabular}{cc}
       \subfloat[$\lambda_1 = 0$ Seed 1]{%
       \includegraphics[width=0.5\linewidth]{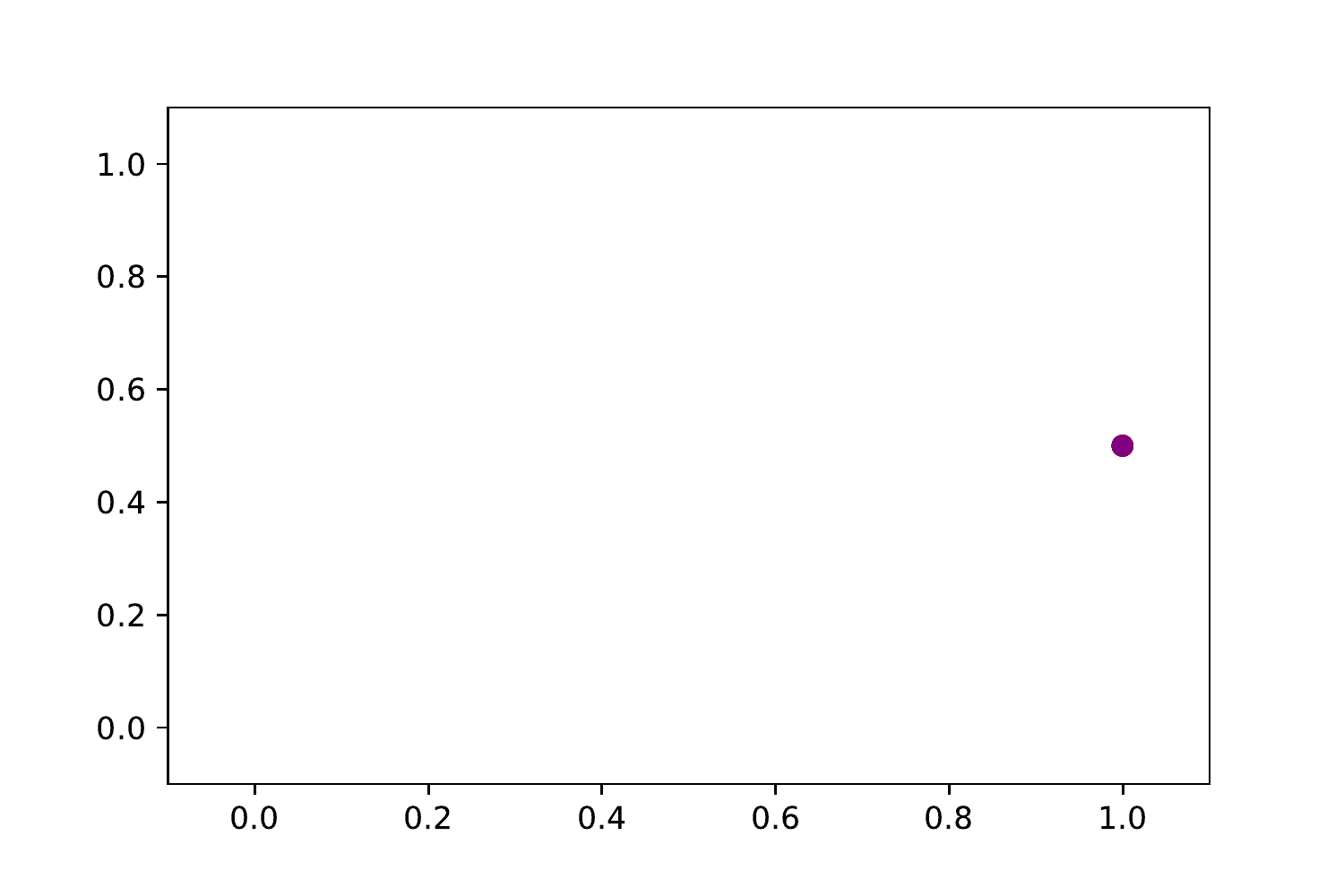}
       } &
       \subfloat[$\lambda_1 = 0$ Seed 2]{%
       \includegraphics[width=0.5\linewidth]{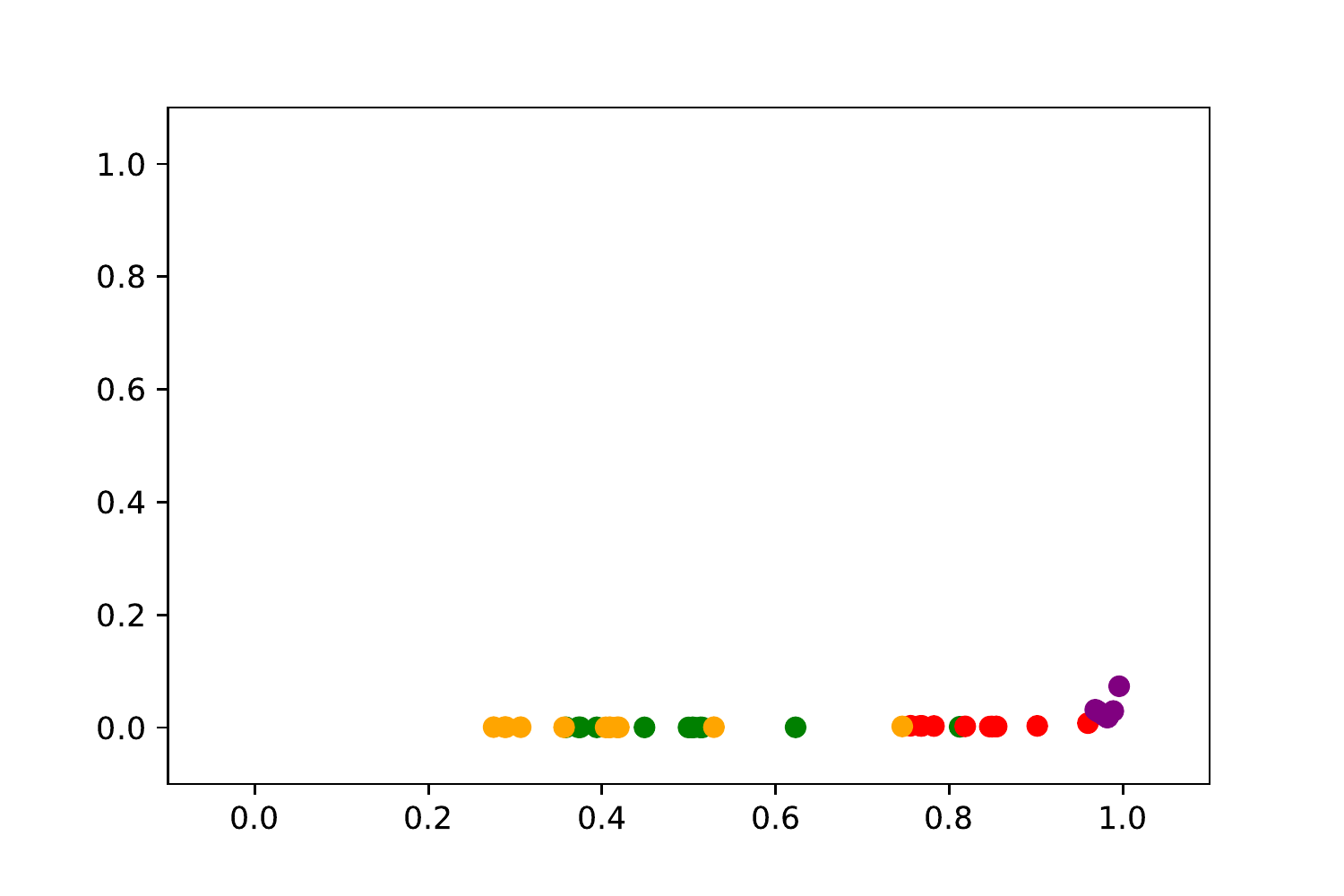}
       } \\
       \subfloat[$\lambda_1 = 0$ Seed 3]{%
       \includegraphics[width=0.5\linewidth]{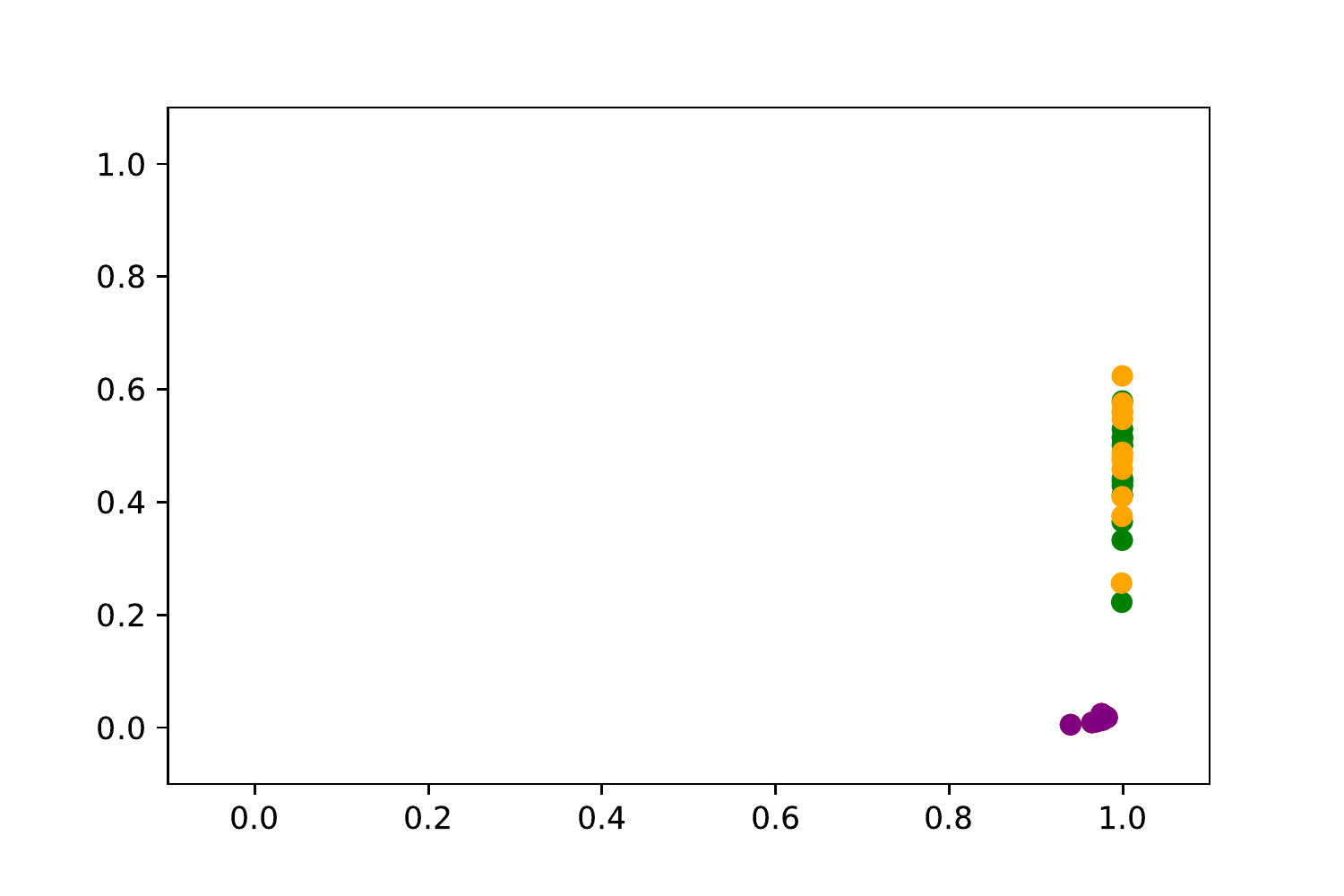}
       } &
       \subfloat[$\lambda_1 = 0$ Seed 4]{%
       \includegraphics[width=0.5\linewidth]{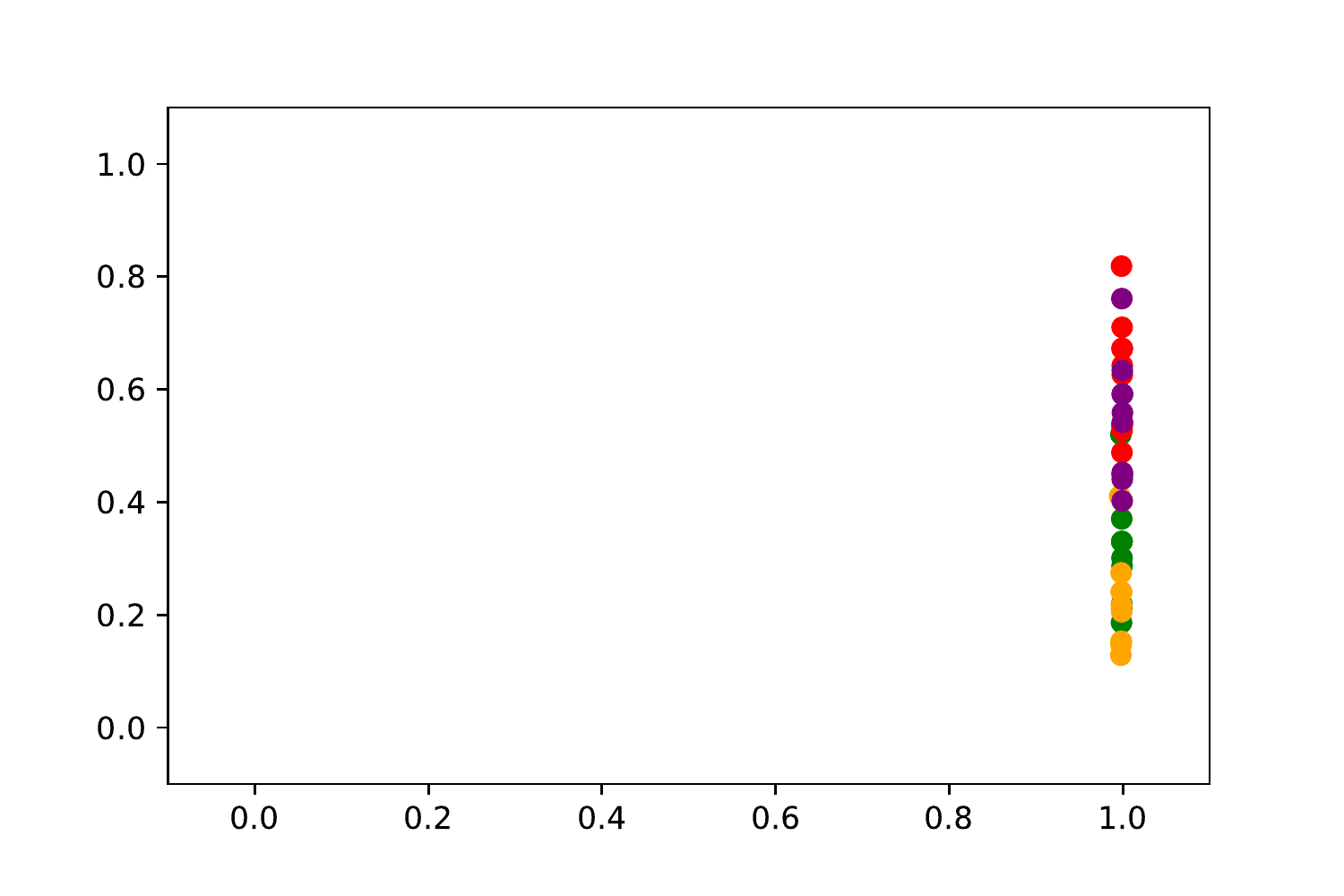}
       }
    \end{tabular}
\caption{Ablation Study of the Embedded Space for the environment \emph{Inclined Plane}. Only $\lambda_2 \neq 0$.
}
\label{fig:ablation_lambda1_0}
\end{figure}

\begin{figure}[t]
    \centering
\begin{tabular}{cc}
       \subfloat[$\lambda_2 = 0$ Seed 1]{%
       \includegraphics[width=0.5\linewidth]{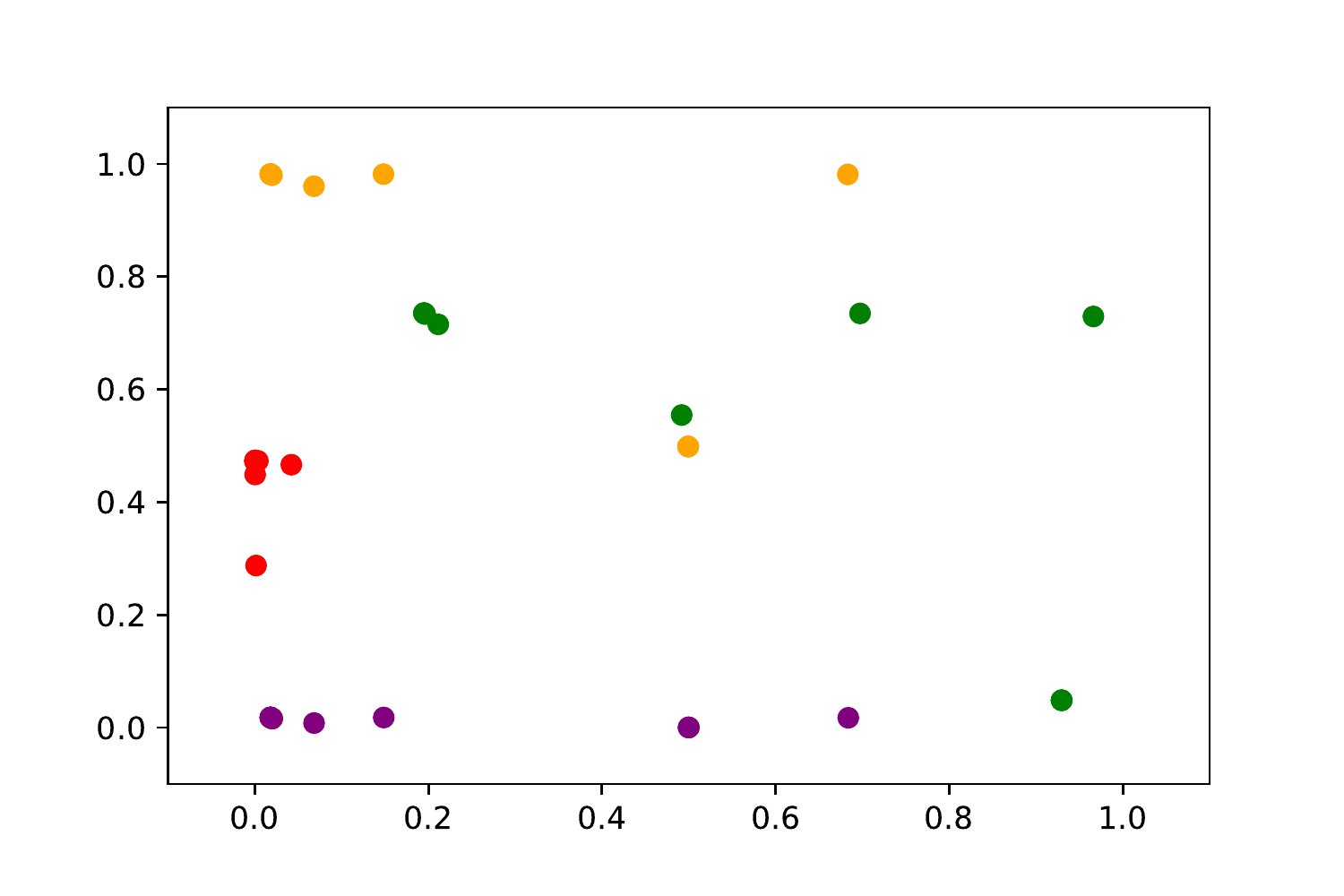}
       } &
       \subfloat[$\lambda_2 = 0$ Seed 2]{%
       \includegraphics[width=0.5\linewidth]{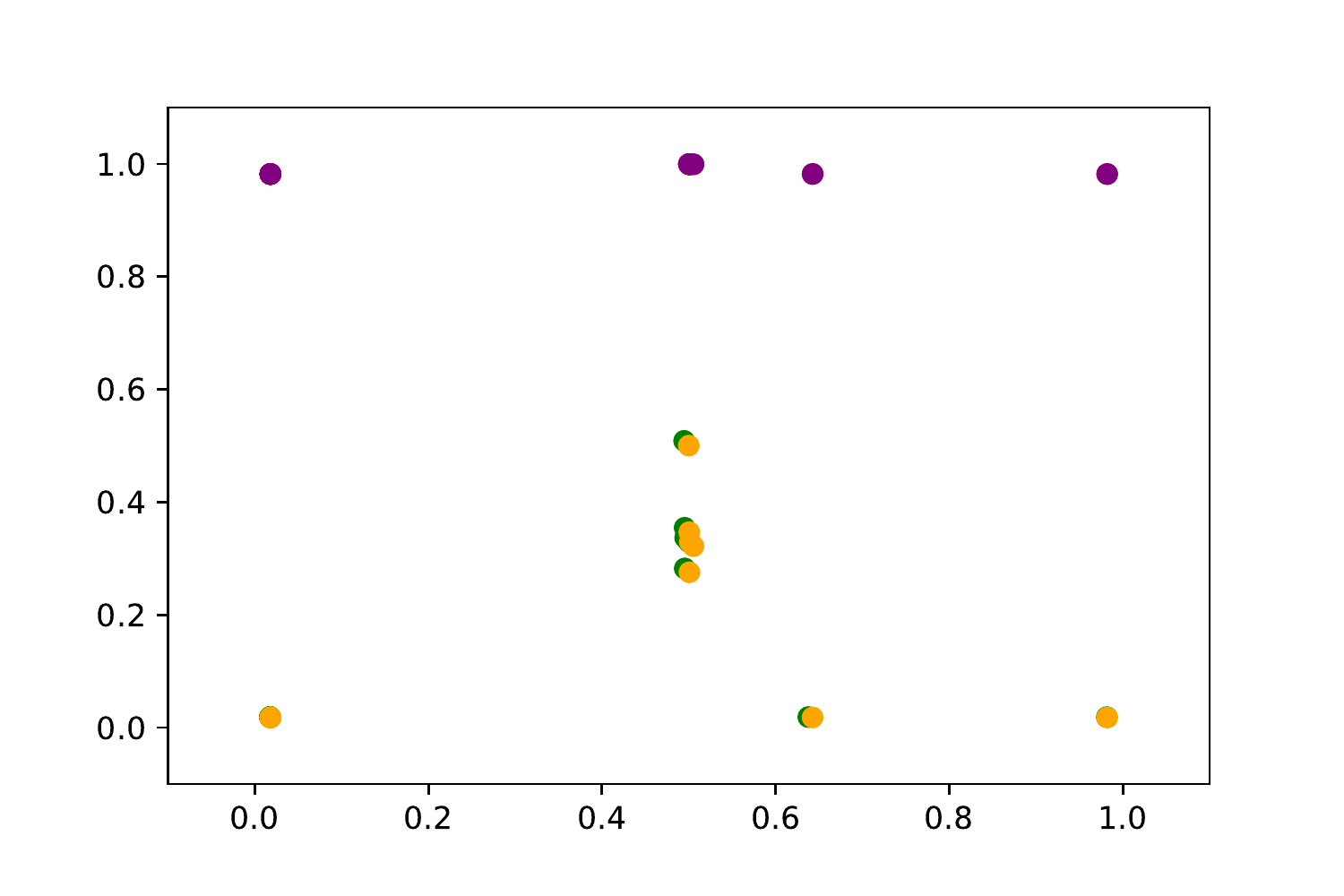}
       } \\
       \subfloat[$\lambda_2 = 0$ Seed 3]{%
       \includegraphics[width=0.5\linewidth]{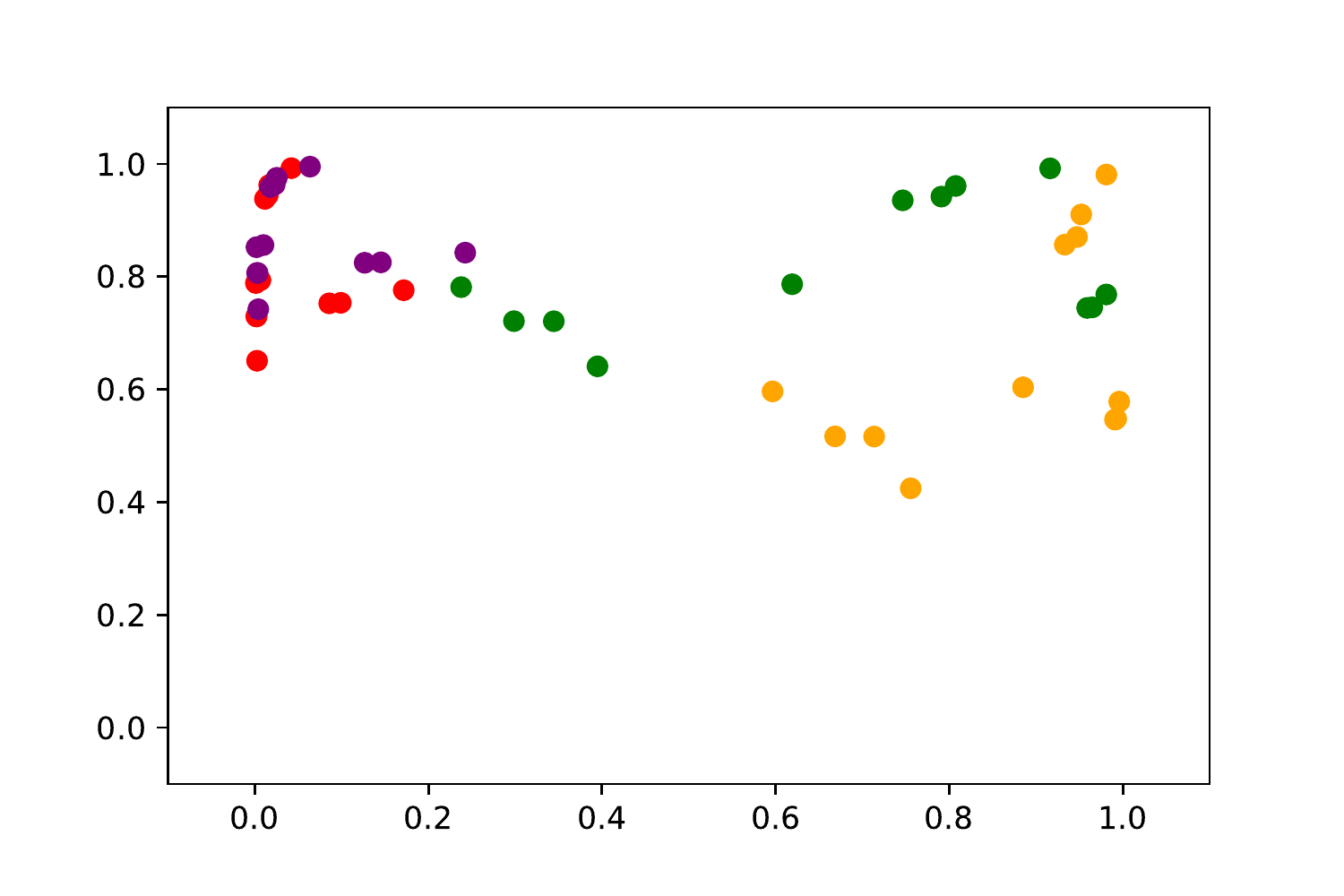}
       } &
       \subfloat[$\lambda_2 = 0$ Seed 4]{%
       \includegraphics[width=0.5\linewidth]{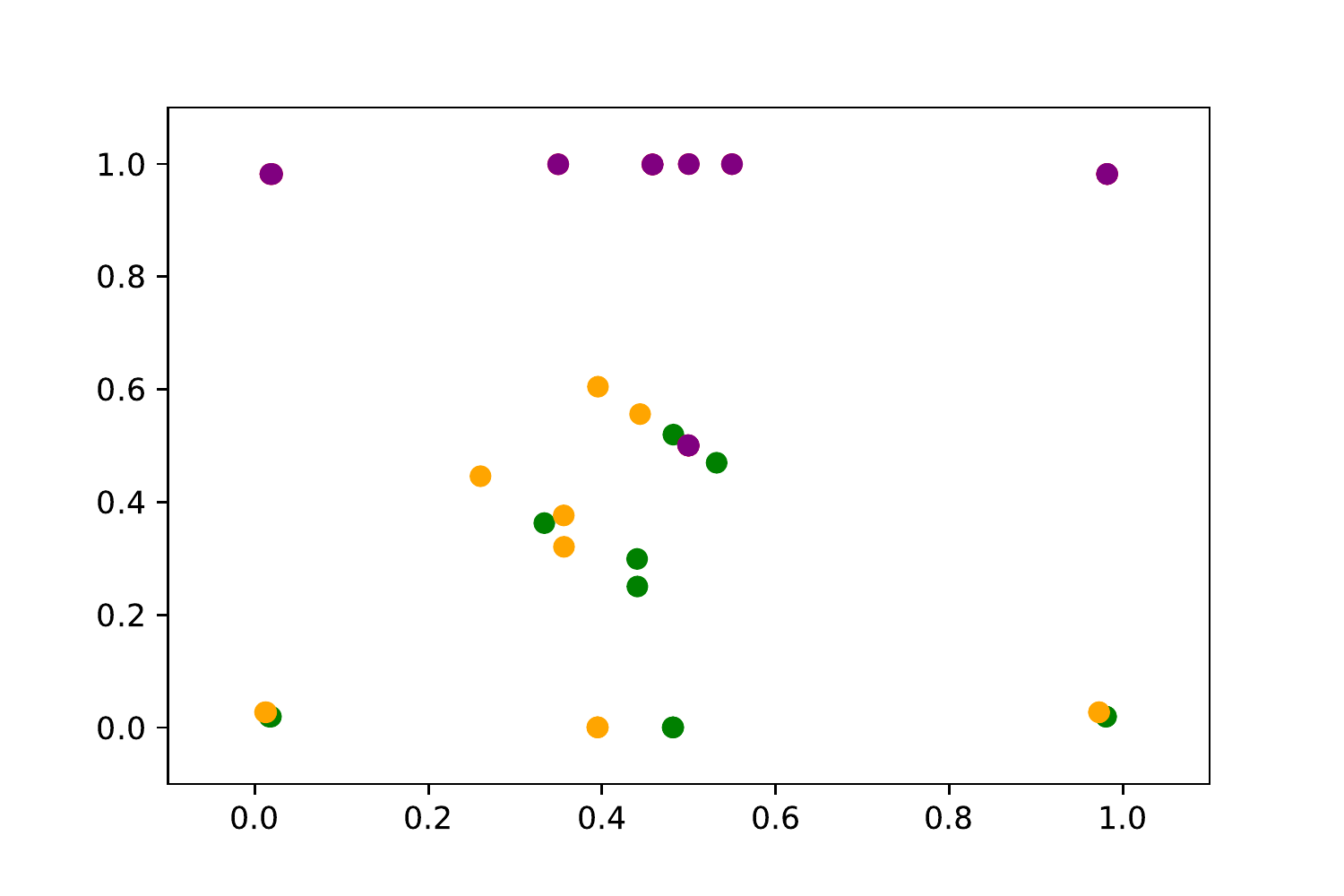}
       }
    \end{tabular}
\caption{Ablation Study of the Embedded Space for the environment \emph{Inclined Plane}. Only $\lambda_1 \neq 0$.
}
\label{fig:ablation_lambda2_0}
\end{figure}

\begin{figure}[t]
    \centering
\begin{tabular}{cc}
       \subfloat[$\lambda_1 = \lambda_2 = 0$ Seed 1]{%
       \includegraphics[width=0.5\linewidth]{embedding_space21.pdf}
       } &
       \subfloat[$\lambda_1 = \lambda_2 = 0$ Seed 2]{%
       \includegraphics[width=0.5\linewidth]{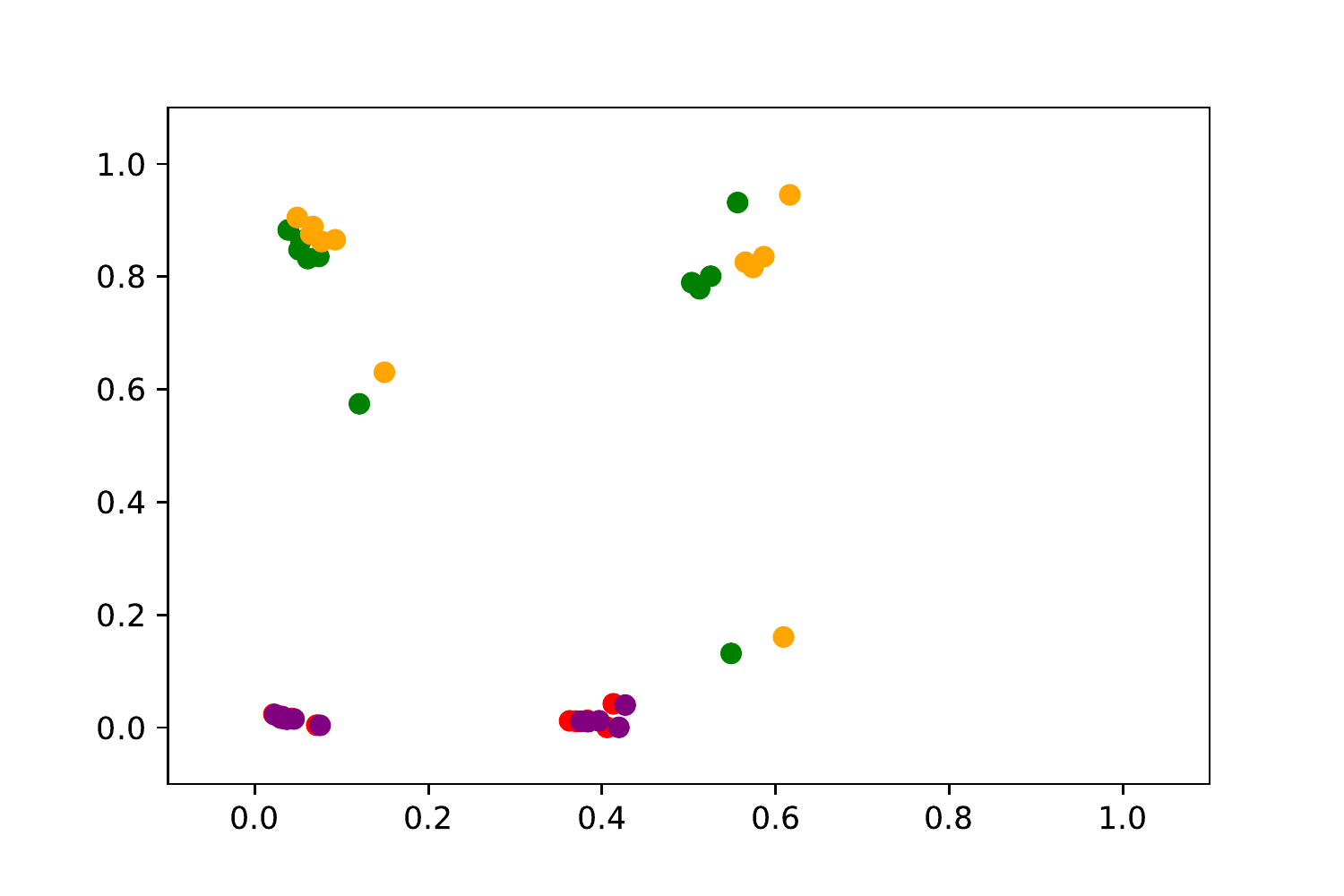}
       } \\
       \subfloat[$\lambda_1 = \lambda_2 = 0$ Seed 3]{%
       \includegraphics[width=0.5\linewidth]{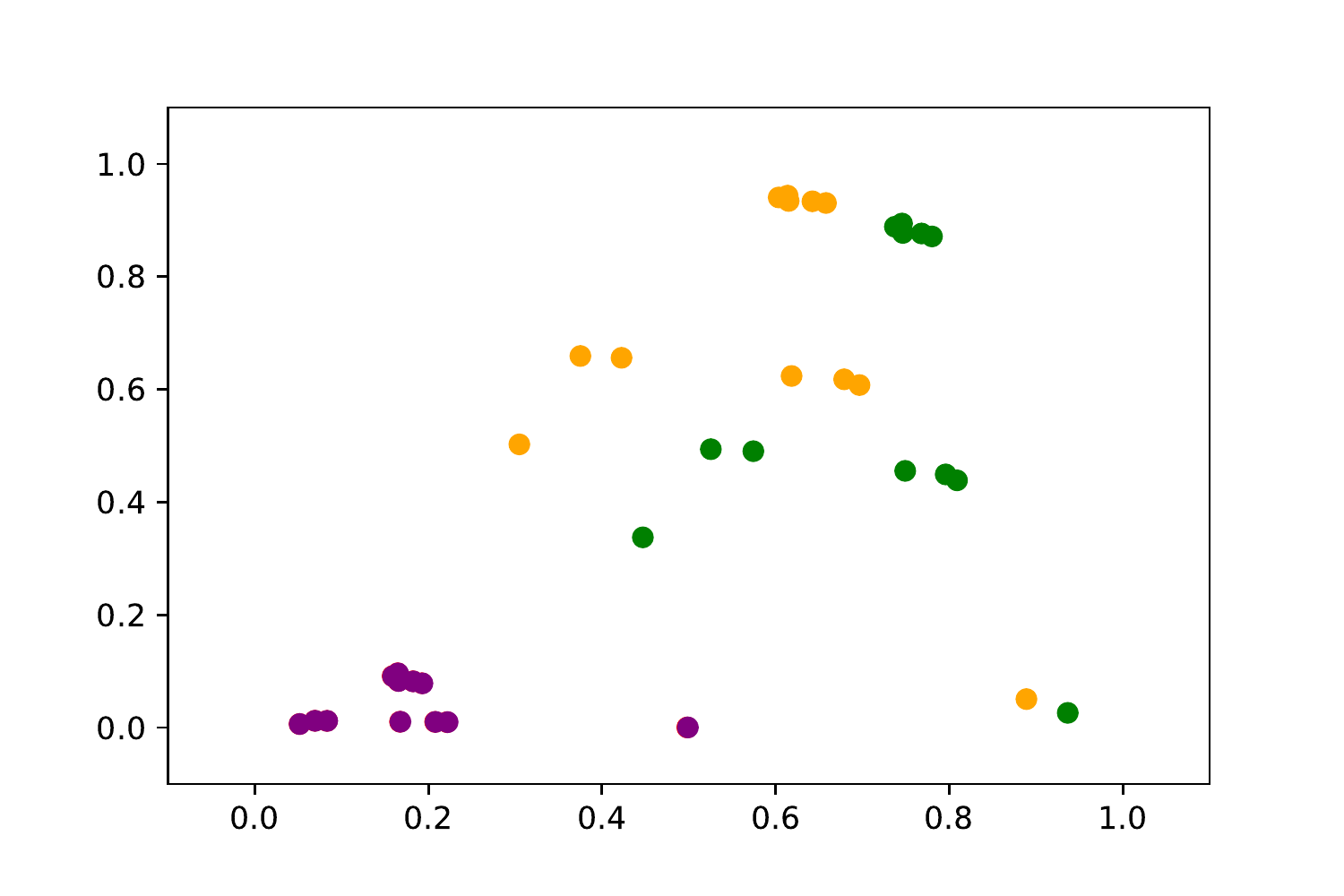}
       } &
       \subfloat[$\lambda_1 = \lambda_2 = 0$ Seed 4]{%
       \includegraphics[width=0.5\linewidth]{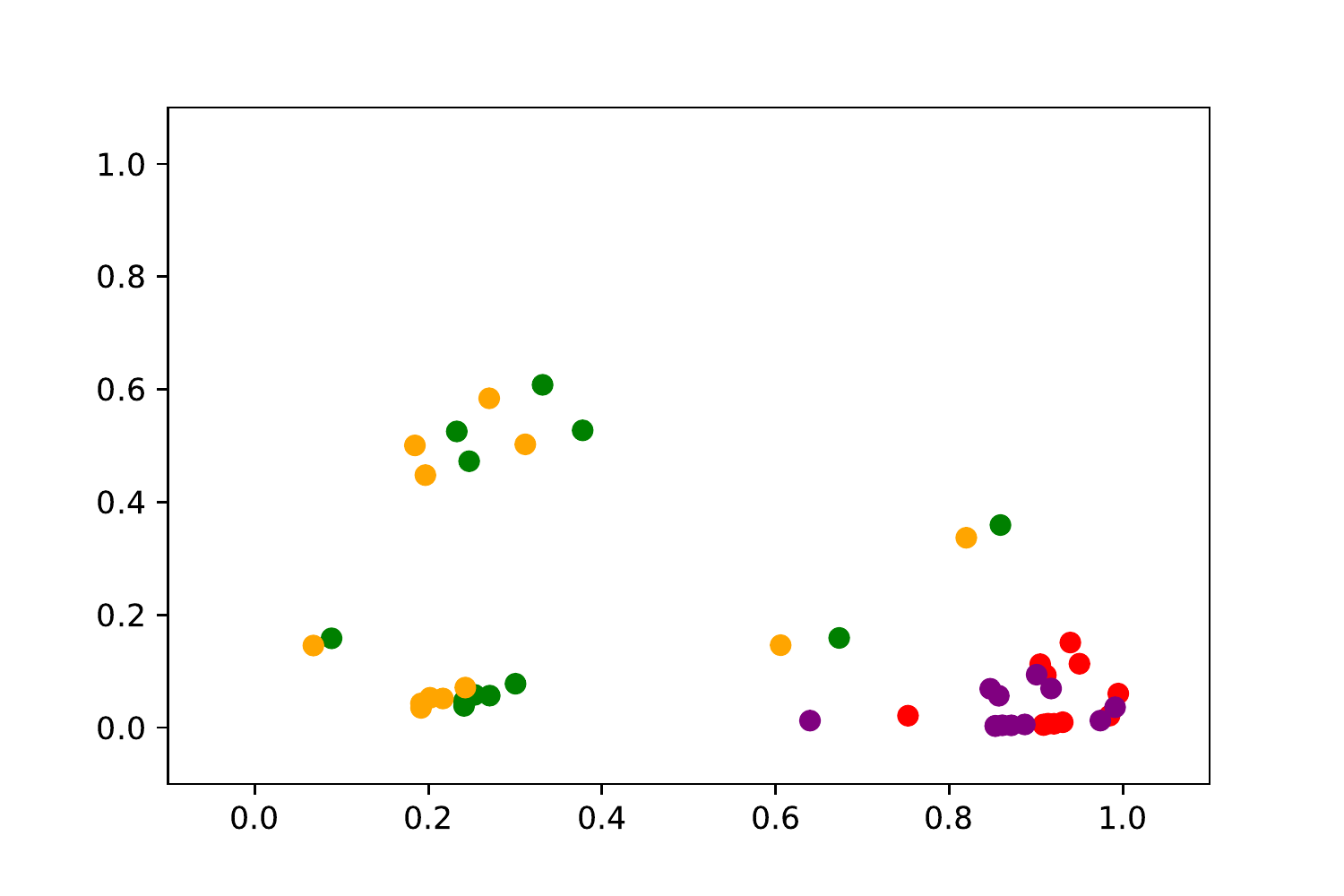}
       }
    \end{tabular}
\caption{Ablation Study of the Embedded Space for the environment \emph{Inclined Plane}. It can be clearly seen that the embedding space present a spread encoding of object type and its position. As a consequences the layers following this one do not use a compact shared representation. This fact makes impossible to achieve the desired out-of-training distribution generalization.
}
\label{fig:ablation_no_entropy}
\end{figure}

\begin{figure}[t]
    \centering
\begin{tabular}{cc}
       \subfloat[$\lambda_1 = 5e{-}7\quad \lambda_2 = 5e{-}6, K = 4$ Seed 1]{%
       \includegraphics[width=0.5\linewidth]{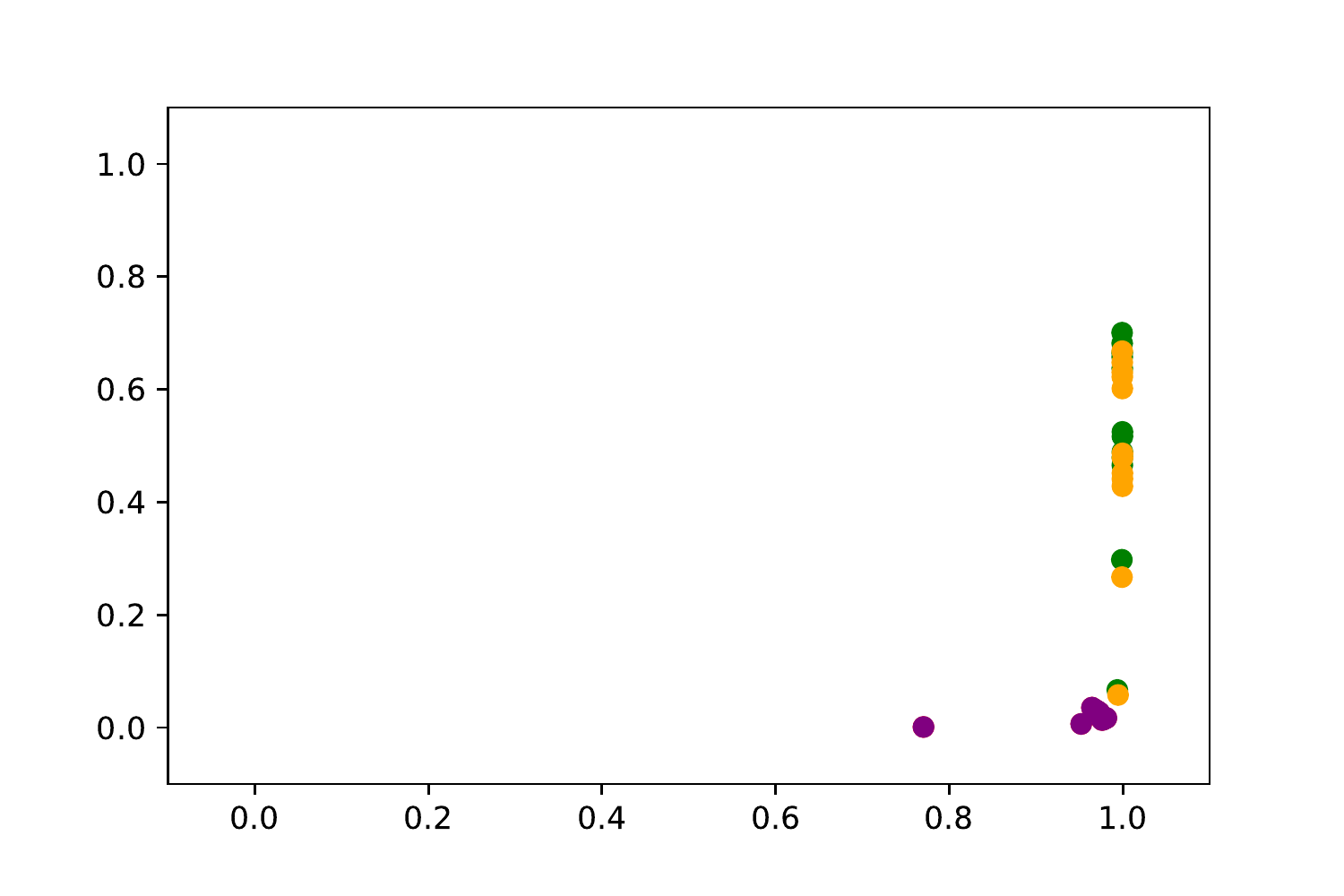}
       } &
       \subfloat[$\lambda_1 = 5e{-}7\quad \lambda_2 = 5e{-}6, K = 4$ Seed 2]{%
       \includegraphics[width=0.5\linewidth]{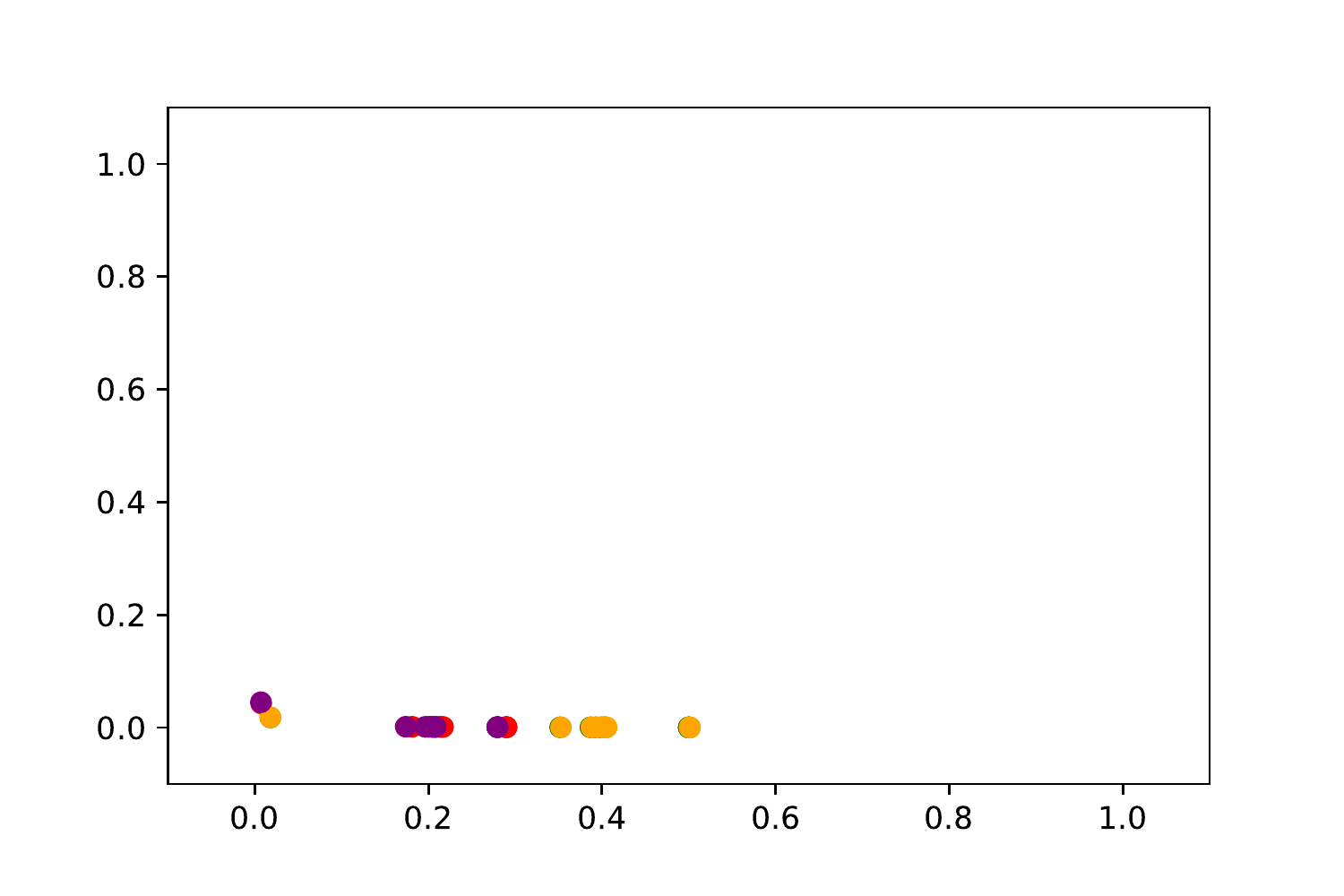}
       } \\
       \subfloat[$\lambda_1 = 5e{-}7\quad \lambda_2 = 5e{-}6, K = 4$ Seed 3]{%
       \includegraphics[width=0.5\linewidth]{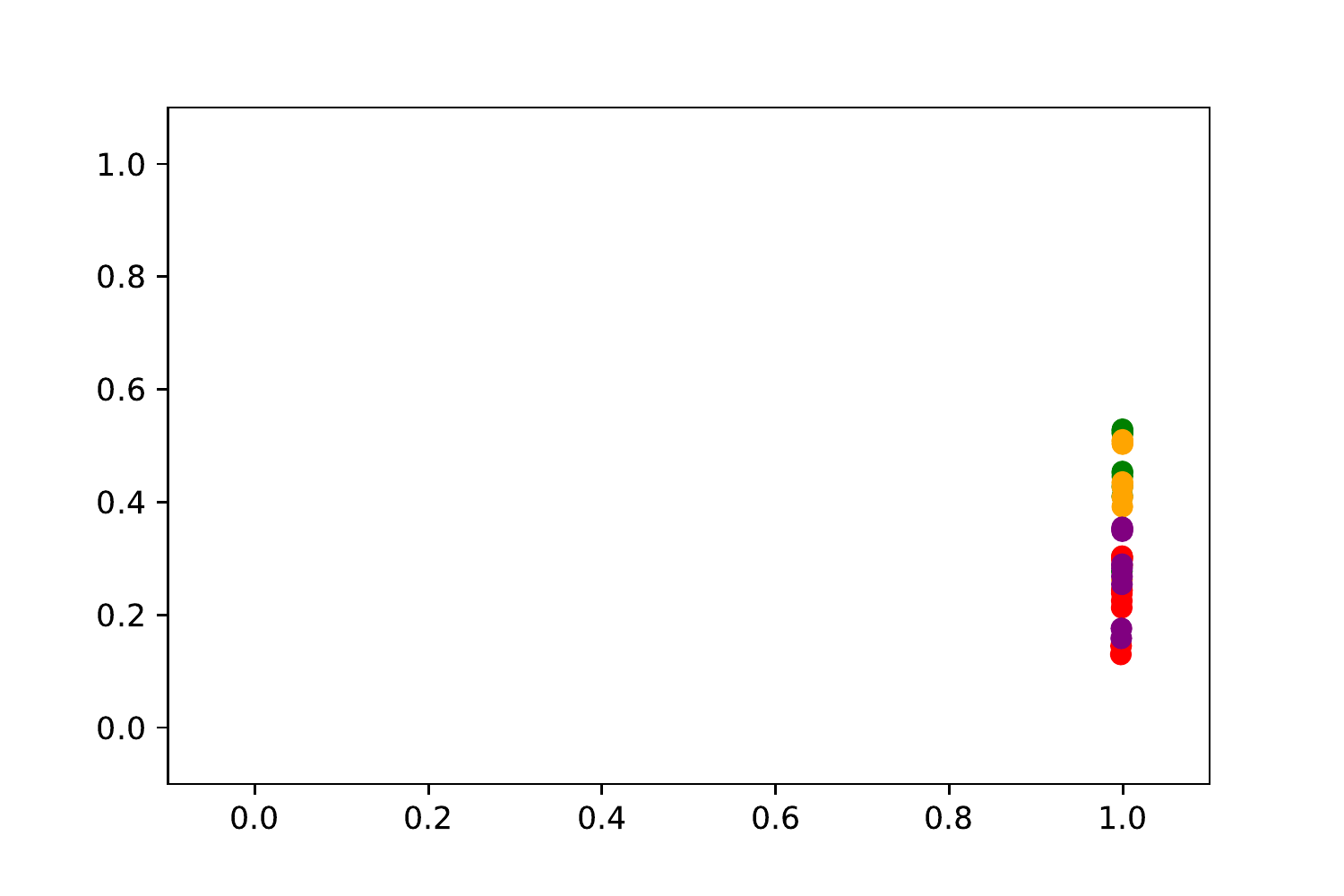}
       } &
       \subfloat[$\lambda_1 = 5e{-}7\quad \lambda_2 = 5e{-}6, K = 4$ Seed 4]{%
       \includegraphics[width=0.5\linewidth]{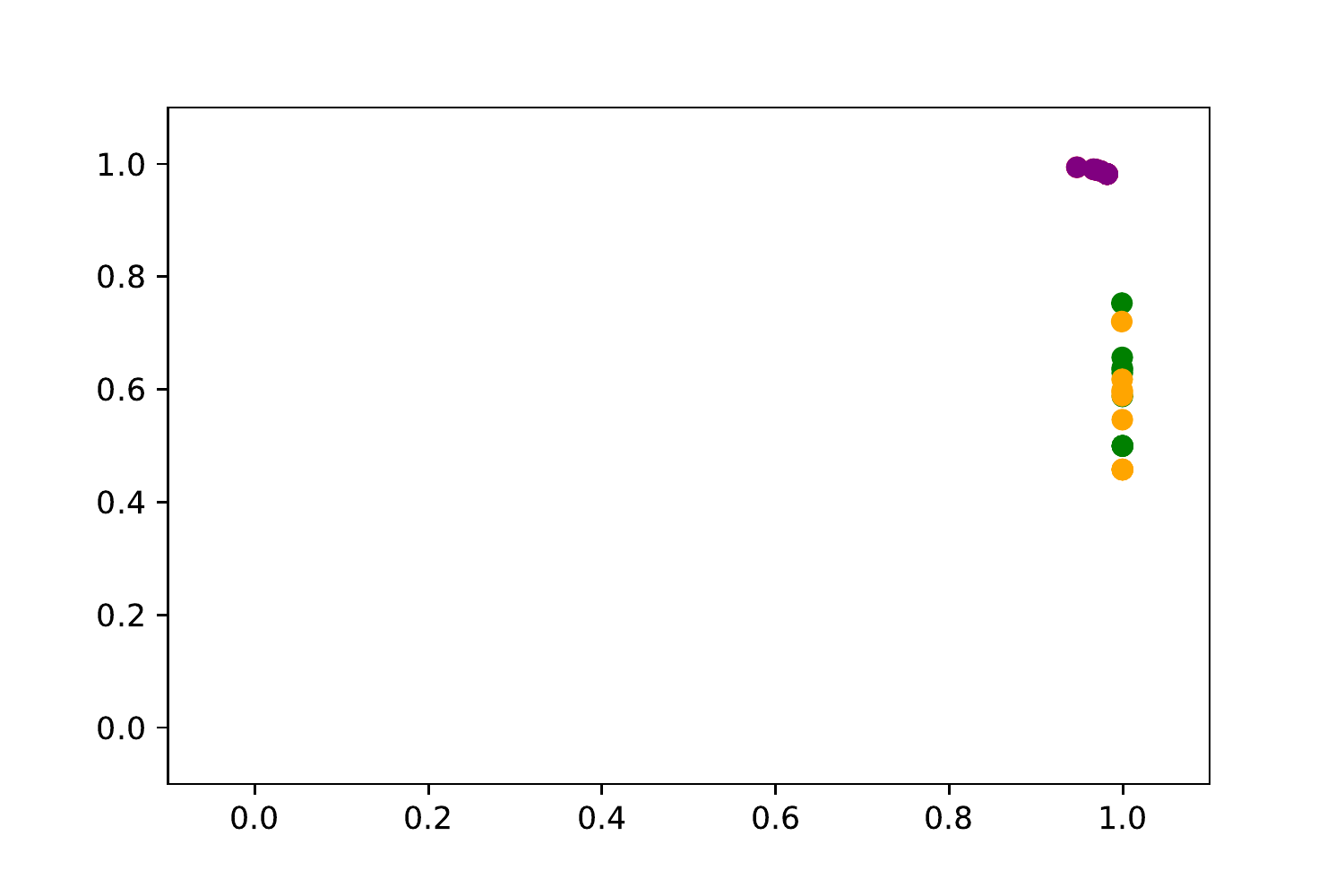}
       }
\end{tabular}
\caption{Ablation Study of the Embedded Space for the environment \emph{Inclined Plane}. The embedding space here present evident clusters containing either the embedding of the rolling objects (green and yellow) or of the non-rolling ones (red and purple).
}
\label{fig:ablation}
\end{figure}

\end{document}